\documentclass[twocolumn]{article}

\usepackage[margin=0.75in, top=1in, bottom=1in]{geometry}
\usepackage[T1]{fontenc}
\usepackage{times}
\usepackage{natbib}
\usepackage{amsmath,amssymb}
\usepackage{booktabs,tabularx}
\usepackage{xcolor}
\usepackage{hyperref}
\usepackage{graphicx}
\usepackage{subcaption}
\usepackage{float}
\usepackage{comment}
\usepackage{microtype}
\usepackage{fancyhdr}
\usepackage{xurl}
\usepackage{tikz}
\usetikzlibrary{arrows.meta, positioning, fit, backgrounds}
\hypersetup{colorlinks=true, linkcolor=blue!60!black, citecolor=blue!60!black, urlcolor=blue!60!black}

\usepackage{amsmath,amsfonts,bm}

\def\eqref#1{equation~\ref{#1}}

\def\1{\bm{1}}

\DeclareMathAlphabet{\mathsfit}{\encodingdefault}{\sfdefault}{m}{sl}
\SetMathAlphabet{\mathsfit}{bold}{\encodingdefault}{\sfdefault}{bx}{n}

\newcommand{\R}{\mathbb{R}}

\title{\textbf{Shaped SAEs: Expanding the Toolkit for Mechanistic Interpretability}}

\author{
  Oveys Delafrooz Noroozi\textsuperscript{1} \quad
  Bhagyashree Puranik\textsuperscript{1} \quad
  Upamanyu Madhow\textsuperscript{1}\\[4pt]
  \textsuperscript{1}University of California, Santa Barbara\\
  \texttt{\{oveys, bpuranik, madhow\}@ucsb.edu}
}

\date{}

\begin{document}

\maketitle
\thispagestyle{fancy}

\begin{abstract}
Sparse autoencoders (SAEs) seek to decompose the dense activations of large language models (LLMs) into sparse, potentially interpretable features using a learned overcomplete dictionary. In this work, we ask whether it is possible to enhance the semantic selectivity of the learned dictionary atoms by linearly \emph{shaping} the dense activations prior to learning an SAE. We explore rank-reducing transformations based on the recently introduced J-space corresponding to the averaged Jacobian from the current layer to the LLM output, projecting activations onto the ``live'' subspace spanned by eigenmodes of the Jacobian with larger singular values (which we expect to be more semantically relevant). We also introduce a new approach for training SAEs in which (a)~an affine shaping transformation is learned along with the dictionary atoms, (b)~the atom selection mechanism is based on the TopK projections along the one-dimensional subspace spanned by the atoms, and (c)~reconstruction is carried out using the same atoms. We call this approach a Matched Filter SAE (MF-SAE) based on analogies with communication theory. The promise of shaping is demonstrated via evaluations against dense activations and state-of-the-art SAEs on benchmarks including SAEBench and SAE Probes.

\end{abstract}

\section{Introduction}
\label{sec:intro}
Mechanistic interpretability seeks to develop insight into the internal representations and computations of large language models (LLMs). A key challenge is that model activations are high-dimensional dense representations encoding many concepts at once (i.e., they are ``polysemantic''), making it difficult to isolate individual concepts and computational variables, and to relate them to model behavior. Sparse autoencoders (SAEs) have been proposed as a practical approach for replacing dense neural activations with potentially more interpretable feature vectors corresponding to their sparse representation with respect to an overcomplete dictionary
\citep{cunningham2023sparse,bricken2023monosemanticity,templeton2024scaling}. Significant effort has been devoted to developing a variety of SAE formulations
\cite{gao2025scaling,bussmann2024batchtopk,rajamanoharan2024gated,rajamanoharan2024jumprelu,bussmann2025matryoshka}, along with benchmarks assessing semantic selectivity \cite{karvonen2025saebench,kantamneni2025saeprobe}.

Our broad objective is to investigate whether the toolkit for mechanistic interpretability using SAEs can be expanded by asking a simple question: can we enhance the semantic selectivity of the learned dictionary atoms by allowing ourselves the freedom to \emph{shape} activations prior to learning an SAE? We make two key contributions that answer this question in the affirmative:

\noindent
(1) \emph{J-space Shaping:} Using the recently introduced Jacobian method \cite{gurnee2026workspace}, we shape activations accounting for their impact on \emph{downstream transmission}. Locally, the Jacobian of a downstream mapping describes how perturbations to a residual-stream activation propagate through the model. Its leading right singular vectors therefore identify source-space directions with high sensitivity under that mapping. We use the leading singular vectors of a corpus-averaged Jacobian to divide the residual stream into a high-transmission \emph{live} subspace and its orthogonal \emph{quiet} complement. The terminology is deliberately relative: quiet directions are less transmitted by the selected average Jacobian, but they are not necessarily an exact null space, devoid of semantic information, or irrelevant in every context. However, we do expect the live directions to be more semantically relevant, and therefore investigate the impact of shaping activations by projecting them onto the live subspace.

\noindent
(2) \emph{Matched Filter SAE (MF-SAE):} We introduce a new SAE formulation in which we apply a learnable affine transformation prior to encoding and decoding with an overcomplete dictionary, with the affine transformation and dictionary atoms learned jointly based on reconstruction loss. The encoder outputs are TopK rectified projections along the unit-norm atoms, while the decoder uses the same atoms and projections for reconstruction. We are inspired by the concept of matched filters from communication theory to learn atoms corresponding to dominant directions in the data, but noting that matched filters are only optimal when the noise is white, we insert a linear transformation up front as a learnable ``whitening'' transformation. The bias term is motivated as a learnable centering mechanism that minimizes the energy of the atoms needed to represent the data, similar to the centering routinely used in signal constellations in communication systems.

We show that our proposed methods lead to improved performance on standard benchmark tasks on semantic feature extraction under a constrained feature budget on SAEBench and SAE Probes. We make the following high-level observations regarding performance on SAEBench:
\begin{itemize}
    \item Semantically shaped dense activations (prior to learning an SAE) perform better than original dense activations.
    \item The MF-SAE applied to standard dense activations performs better than state-of-the-art SAEs as well as dense activations on SAEBench.
    \item The MF-SAE applied to semantically shaped activations performs slightly better than the MF-SAE at very tight feature budgets, but does not appear to improve performance at higher budgets. Thus, the problem of how best to combine semantic shaping with sparse coding remains open.
\end{itemize}
Performance trends on the SAE Probes benchmark are more variable, as we discuss in more detail in our results section (Section~\ref{sec:results}).
We evaluate on Gemma-2-2B at layers 12 and 20, and on GPT-2 Small at layers 7 and 8 (Appendix~\ref{app:gpt2}). On SAEBench, the MF-SAEs outperform the available baseline SAEs for both models, while on SAE Probes the gains are marginal. Probing accuracy is higher on Gemma-2-2B than on GPT-2 Small on both benchmarks, which we attribute to model scale: the tasks of SAE Probes in particular are more complex and benefit from a larger model.

\section{Related Work}
\label{sec:related}
\noindent
{\bf SAEs for language models:} SAEs use dictionary learning as a means of decomposing the dense, superposed representations of language models into sparse features that are potentially more interpretable \citep{cunningham2023sparse,bricken2023monosemanticity}. A variety of SAE formulations have been proposed, including $L_1$-regularized SAEs, TopK and BatchTopK SAEs \cite{gao2025scaling,bussmann2024batchtopk}, Gated SAEs \cite{rajamanoharan2024gated}, JumpReLU SAEs \cite{rajamanoharan2024jumprelu}, and Matryoshka SAEs \cite{bussmann2025matryoshka}. These methods differ in how they impose sparsity and trade it against reconstruction fidelity. With respect to this state of the art, our contribution is two-fold: (a)~we propose shaping dense activations prior to learning sparse representations, focusing on J-space \cite{gurnee2026workspace} based transformations that account for downstream information transmission; (b)~we propose a new SAE formulation including a learned geometric shaping and an atom selection mechanism combining ideas from TopK and matching pursuit.

\noindent
{\bf Jacobian-based representations and computation:} Our Jacobian-based shaping builds on the work in \cite{gurnee2026workspace}, which averages the linearized map from an intermediate residual stream to later residuals and uses the resulting token-associated directions to define and interrogate a sparse ``J-space.'' While we build on the insight that activation variance and downstream information transfer are not directly related, we use the Jacobian in \cite{gurnee2026workspace} in a different way, employing its singular value decomposition to identify important directions for downstream information transfer to shape the activations prior to learning an SAE-based sparse representation. Unrelated but worth mentioning are Jacobian Sparse Autoencoders \cite{farnik2025jacobian}, which jointly seek sparsity in the input features, output features, and the connections between them for an MLP layer.

\noindent
{\bf Cautionary notes:} Recent work has raised important questions about the extent to which SAE features improve mechanistic analysis.
Good reconstruction does not by itself guarantee a faithful or useful feature decomposition. Feature splitting and absorption can cause apparently coherent latents to omit expected instances \cite{chanin2024absorption}. It was originally shown in \cite{kantamneni2025saeprobe} (the SAE Probes benchmark) that SAE representations do not consistently outperform dense activations or simple baselines for semantic probing, including under data scarcity, label noise, class imbalance, and distribution shift; however, improved SAEs (including our own) have addressed these objections. It is shown in \cite{mueller2025mib} (the MIB benchmark) that SAE features do not uniformly improve localization of causal variables relative to alternative representations. The AxBench benchmark provides complementary evidence in the context of concept detection and steering, where simple non-SAE baselines can be highly competitive \citep{wu2025axbench}. These results caution against treating sparse, interpretable-looking features or reconstruction fidelity as ends in themselves, and motivate exploring alternatives to the current SAE paradigm. As a counterpoint, it is argued in \cite{peng2026_SAE_position_paper} that the primary utility of SAEs is as a tool for concept discovery, despite their limitations in tasks such as steering and circuit analysis. In this paper, we do not seek to resolve broader questions regarding the utility of SAEs; our focus is to expand the universe of available sparse representations with the goal of enhancing semantic selectivity under stringent feature budgets.

\noindent
{\bf Benchmarks for mechanistic interpretability:} Several recent benchmarks seek to evaluate semantic selectivity of proposed feature representations.
\textbf{SAEBench} is the most comprehensive SAE-oriented evaluation, comparing architectures across reconstruction, sparsity, feature quality, probing, and functional metrics \citep{karvonen2025saebench}. \textbf{SAE Probes} directly tests whether SAE coordinates provide a useful sparse basis for recovering semantic information, including comparisons with dense representations \citep{kantamneni2025saeprobe}. Our experimental evaluations compare our proposed shaped SAEs to both dense activations and state-of-the-art SAEs using SAEBench and SAE Probes.

\textbf{MIB} shifts the emphasis toward mechanistic validity, evaluating localization of causal variables and circuits rather than feature interpretability alone \citep{mueller2025mib}. \textbf{AxBench} evaluates representation-based concept detection and steering, explicitly comparing SAEs with alternative approaches to representation intervention \citep{wu2025axbench}. For the steering, concept detection, and causal variable localization tasks considered in these benchmarks, a variety of supervised techniques are shown to outperform SAEs. In this paper, we do not seek to address this gap. Our focus is on the complementary question of improving the unsupervised, semantically selective, sparse representations provided by SAEs.

\section{Jacobian-based Shaping}
\label{sec:shaping_activations}
Let $d$ denote the model dimension in a decoder-only language model, and let $h_{\ell,t} \in \R^{d}$ denote a residual-stream activation at a specified layer $\ell$ and token position $t$. Let $h_{\text{final}, t'} \in\R^{d}$ denote the final-layer residual stream at position $t'$. The averaged Jacobian matrix $J$ defined in \cite{gurnee2026workspace} is a first-order approximation for how perturbations in $h_{\ell,t}$ propagate downstream to the final layer and affect $h_{\text{final}, t'}$ for $t' \geq t$, averaged over a corpus $\mathcal{D}$ of prompts sampled from a pretraining-like distribution:
\begin{equation}
    J_{\ell} \;=\; \mathbb{E}_{\,t,\ t' \geq t,\ \text{prompt} \sim \mathcal{D}}
    \left[ \frac{\partial h_{\text{final},t'}}{\partial h_{\ell,t}} \right]
    \;\in\; \R^{d \times d}.
    \label{eq:jlens-matrix}
\end{equation}

The J-lens defined in \cite{gurnee2026workspace} applies the averaged Jacobian $J_{\ell}$ to activations at layer $\ell$ and then unembeds them (after normalization) to obtain an output token distribution corresponding to that token. Sorting the output distribution corresponding to an activation at layer $\ell$ provides a human-readable description of what tokens the activation might tend to produce at the output, averaged across contexts.

We utilize the averaged Jacobian in an entirely different way in this paper. Prior to developing sparse representations for activations at layer $\ell$, we seek to shape them using the Jacobian $J = J_{\ell}$ to emphasize downstream sensitivity, in order to better use the representation capacity of the SAE. We suppress $\ell$ from the notation henceforth for simplicity.

\noindent
{\bf Live and quiet J-spaces:} We quantify downstream sensitivity via the SVD of the Jacobian, $J = U \Sigma V^T$, where $U$ and $V$ are $d \times d$ orthonormal matrices containing the left and right singular vectors, and $\Sigma = \mathrm{diag}(\sigma_1,\ldots,\sigma_d)$ is the diagonal matrix of singular values sorted in decreasing order ($\sigma_1 \geq \sigma_2 \geq \cdots \geq \sigma_d$). We define the ``live'' space as directions corresponding to the largest $r$ singular values $\sigma_i$, $i \leq r$, collected in the diagonal matrix $\Sigma_{\mathrm{live}} = \mathrm{diag}(\sigma_1,\ldots,\sigma_r)$. The corresponding left and right singular vectors are collected into $d \times r$ matrices $U_{\mathrm{live}}$ and $V_{\mathrm{live}}$, respectively.
The value of $r$ used to partition into live and quiet spaces is a hyperparameter that we experiment with.

\noindent
{\bf Shaping via the live space:} Under the intuition that the live modes are more semantically informative given their higher downstream gain, we shape the activations by projecting them onto the live space. That is, for a residual-stream activation $h$, we obtain a shaped activation $h_{\mathrm{live}}$ by projecting $h$ onto the live space as follows:
\begin{equation}
h_{\mathrm{live}} = V_{\mathrm{live}} V_{\mathrm{live}}^T h.
\label{j_shaping}
\end{equation}
We may view discarding the projection in the quiet space as a denoising operation. As we have mentioned, quiet directions are not necessarily devoid of semantic information, but given their lower average downstream gain, we hope to gain in semantic selectivity when operating on a tight feature budget.

\begin{figure}[t]
    \centering
    \includegraphics[width=\linewidth, height=0.6\linewidth]{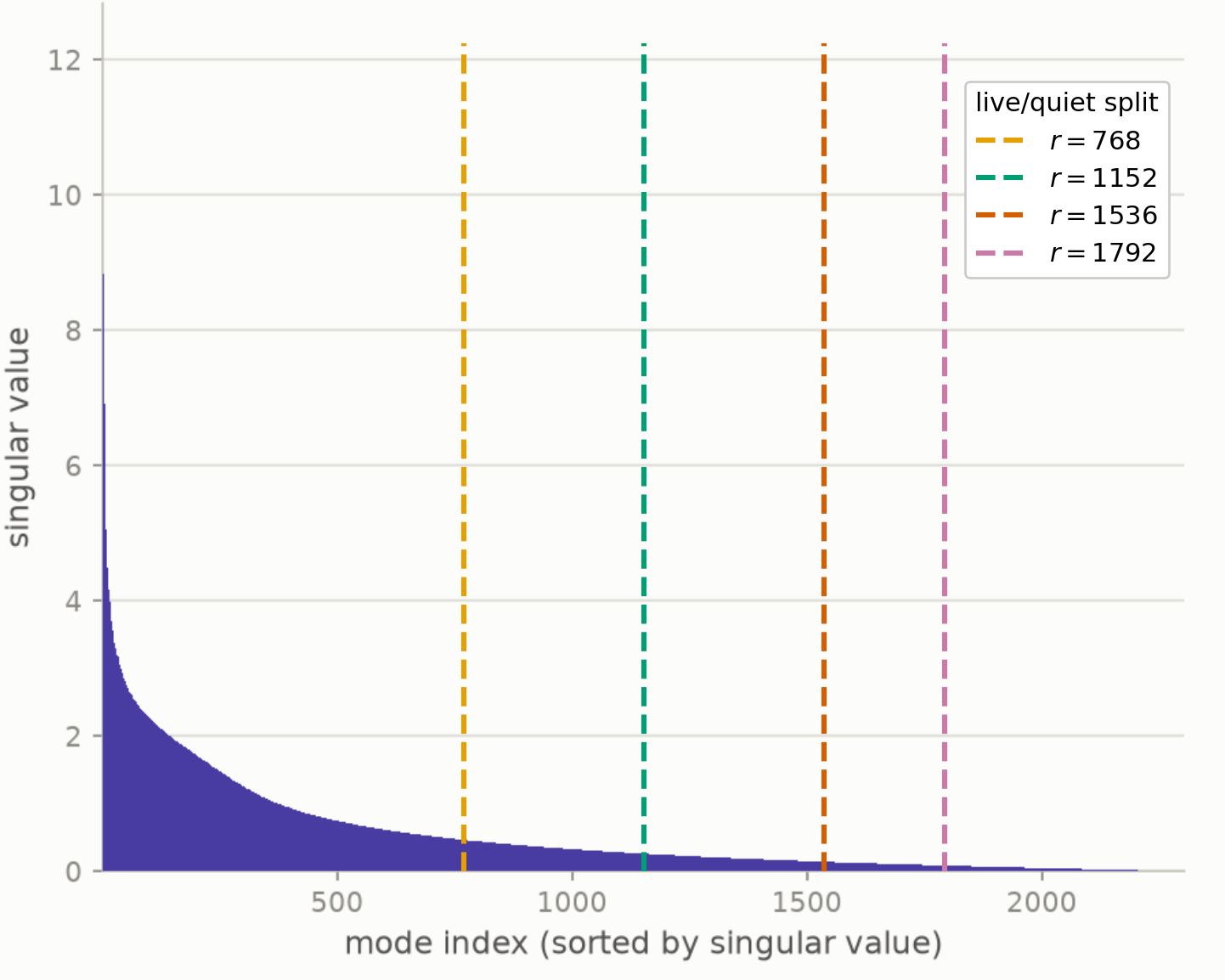}
    \caption{\scriptsize{\textbf{Singular values of the averaged Jacobian $J$ (Gemma-2-2B, layer 12).} Singular values sorted in decreasing order. The spectrum decays rapidly, so a small number of directions carry most of the downstream gain and the remaining directions have little effect on the output. Dashed vertical lines mark the live/quiet splits $r\in\{768,1152,1536,1792\}$ studied in Section~\ref{sec:res-shaping}; the right singular vectors to the left of each line span the live space. We use $r=1536$ throughout.}}
    \label{fig:jacobian-spectrum}
\end{figure}

In our benchmarking with SAEBench and SAE Probes, we compare dense activations $h$ and shaped dense activations $h_{\mathrm{live}}$ against the SAEs that we train, as well as against state-of-the-art SAEs.

\section{The Matched Filter SAE (MF-SAE)}
\label{sec:mfsae}
Let $x \in \R^d$ denote the input to the SAE. In our setting, we may have $x=h$, the original dense activations, or $x = h_{\mathrm{live}}$, the projection onto the live J-space.

\begin{figure*}[t]
\centering
\begin{tikzpicture}[
    font=\footnotesize,
    >={Stealth[length=5pt]},
    node distance=0.5cm,
    op/.style={draw, rounded corners=3pt, align=center, minimum height=1.25cm,
               minimum width=2.0cm, inner sep=3pt},
    affine/.style={op, fill=blue!8},
    sparse/.style={op, fill=green!10},
    var/.style={font=\small, inner sep=1pt},
    flow/.style={->, thick}
]
\node[var]                    (x)    {$x$};
\node[affine, right=0.4cm of x] (enc)  {\textbf{Outer encoder}\\[2pt] $z = W_e x + b_e$};
\node[sparse, right=of enc]   (mf)   {\textbf{Matched filters}\\[2pt] $w_j = \mathrm{ReLU}(\hat d_j^{\top} z)$};
\node[sparse, right=of mf]    (topk) {\textbf{TopK selection}\\[2pt] $\mathcal{S} = \operatorname{arg\,top\text{-}k}_j\, w_j$};
\node[sparse, right=of topk]  (syn)  {\textbf{Synthesis}\\[2pt] $\hat z = \sum_{j\in\mathcal{S}} w_j \hat d_j$};
\node[affine, right=of syn]   (dec)  {\textbf{Outer decoder}\\[2pt] $\hat x = W_d \hat z + b_d$};
\node[var, right=0.4cm of dec] (xh)   {$\hat x$};

\draw[flow] (x)    -- (enc);
\draw[flow] (enc)  -- (mf);
\draw[flow] (mf)   -- (topk);
\draw[flow] (topk) -- (syn);
\draw[flow] (syn)  -- (dec);
\draw[flow] (dec)  -- (xh);

\node[draw, rounded corners=3pt, fill=orange!15, align=center, inner sep=4pt,
      above=0.9cm of topk] (dict)
      {\textbf{Shared dictionary} $D\in\mathbb{R}^{n\times d}$, \quad unit-norm atoms $\hat d_j = d_j/\|d_j\|_2$};
\draw[flow, dashed, orange!70!black] (dict.south -| mf.north)  -- node[left, font=\scriptsize]{analysis} (mf.north);
\draw[flow, dashed, orange!70!black] (dict.south -| syn.north) -- node[right, font=\scriptsize]{synthesis} (syn.north);

\begin{scope}[on background layer]
  \node[draw=green!50!black, dashed, rounded corners=5pt, inner sep=6pt,
        fit=(mf)(topk)(syn), label={[font=\scriptsize, text=green!40!black]below:Sparse coding with the same atoms for selection and reconstruction}] {};
\end{scope}
\end{tikzpicture}
\caption{\textbf{The MF-SAE architecture.} The input $x$ is mapped to a latent $z$ by the learned outer affine encoder $(W_e, b_e)$. The latent is correlated with the unit-norm atoms $\hat d_j$ (matched filters), and the $k$ largest rectified projections are kept, giving the sparse code $c$. The same atoms reconstruct the latent $\hat z$ from $c$, which the learned outer affine decoder $(W_d, b_d)$ maps back to the reconstruction $\hat x$.}
\label{fig:mfsae}
\end{figure*}
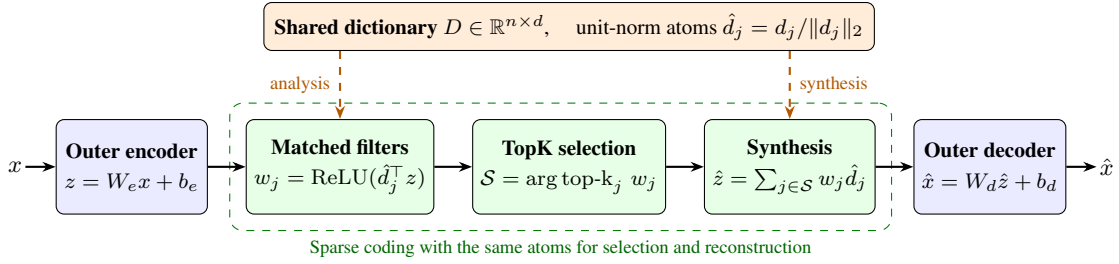

The MF-SAE consists of an outer affine encoder, a dictionary with $n$ elements, and an outer affine decoder (Figure~\ref{fig:mfsae}). The norms of the dictionary elements are not constrained (thus simplifying training), but the atoms used as matched filters for encoding and decoding are unit-norm versions of the dictionary elements.

\paragraph{Outer encoder.} The input is first mapped to a latent
\begin{equation}
    z=W_e x+b_e,
    \label{eq:mf-enc}
\end{equation}
with $W_e\in\R^{d\times d}$ and $b_e\in\R^{d}$ learned.

\paragraph{Dictionary elements and normalized atoms.} The dictionary is a matrix $D\in\R^{n\times d}$ with rows $d_j$, and the atoms are their normalized versions $\hat d_j=d_j/\|d_j\|_2$. The atoms have no biases, and the same unit vector $\hat d_j$ is used both to detect atom $j$ in the latent and to reconstruct it. The encoder ($W_e$, $b_e$) therefore shapes the space, and hence the metric, in which matching is performed.

\paragraph{Sparse coding via TopK matched filter selection.}
We select the top-$k$ atoms with the largest rectified matched filter outputs. Specifically, let
\begin{equation} \label{eq:mf-outputs}
w_j = \max( \hat d_j^\top z , 0 ) = \mathrm{ReLU}(\hat d_j^\top z), \quad j=1,\ldots,n
\end{equation}
denote the rectified projections along the atoms.

We keep the $k$ atoms with the largest projections:
\begin{equation}
    \mathcal S=\operatorname*{arg\,top\text{-}k}_{j}\ w_j.
    \label{eq:mf-select}
\end{equation}
(If fewer than $k$ rectified projections are positive, the number of selected atoms falls below $k$.)

The sparse code $c \in \R^n$ output by the MF-SAE is therefore $c_j = w_j$ for $j \in \mathcal S$ and $c_j = 0$ otherwise.

\paragraph{Decoding.} The latent is reconstructed using the selected atoms, and then mapped back to a reconstruction of the input through the outer affine decoder:
\begin{equation}
    \hat z= \sum_{j=1}^n c_j \hat d_j =  \sum_{j\in\mathcal S}w_j\hat d_j,\qquad
    \hat x=W_d\hat z+b_d,
    \label{eq:mf-dec}
\end{equation}
with $W_d\in\R^{d\times d}$ and $b_d\in\R^{d}$ learned.

\paragraph{Gradients and cost.} Selection uses detached scores, and the coefficients of the selected atoms are recomputed differentiably, so $W_e$, $W_d$, and the atoms receive gradients only through $\mathcal S$. Scoring all atoms is a single product $Z\hat D^\top$ and decoding is a single dense product, so the MF-SAE has the same computational cost as a TopK SAE of the same width.

\paragraph{Relation to TopK SAEs.} Consider a TopK SAE \cite{gao2025scaling} wrapped in the same linear projections, $z=W_{\text{in}}h+b_{\text{in}}$ and $\hat h=W_{\text{out}}\hat z$. Atom $j$ has pre-activation
\begin{equation}
    p_j=e_j^\top(z-b_{\text{pre}})+b_j=\|e_j\|\,\|z-b_{\text{pre}}\|\cos\angle(z-b_{\text{pre}},e_j)+b_j,
    \label{eq:topk-score}
\end{equation}
where $e_j$ is a free encoder row and $b_j$ a free bias. The $k$ largest pre-activations are kept with coefficients $\mathrm{ReLU}(p_j)$ and decoded with free columns $W_{\text{dec}}[:,j]$. The MF-SAE differs in three ways.
\begin{enumerate}
\item[(i)] \emph{Scale.} In \eqref{eq:topk-score} the norm $\|e_j\|$ and the bias $b_j$ can make an atom win the selection independently of how well it matches, and the coefficient is the pre-activation itself. In the MF-SAE the score depends only on the alignment and the coefficient is the least-squares fit of a unit atom.
\item[(ii)] \emph{Bias.} The only per-atom offset in the MF-SAE is $\hat d_j^\top b_e$, which is fixed by the direction of the atom and one shared vector. TopK \cite{gao2025scaling}, BatchTopK \cite{bussmann2024batchtopk}, Matryoshka \cite{bussmann2025matryoshka}, and JumpReLU \cite{rajamanoharan2024jumprelu} SAEs all learn a free bias per latent, and JumpReLU additionally learns a per-latent threshold.
\item[(iii)] \emph{Tying.} The effective encoder direction of atom $j$ in the activation space is $W_e^\top\hat d_j$ and its decoder atom is $W_d\hat d_j$. When $W_d$ is invertible,
\begin{equation}
    W_e^\top\hat d_j=G\,(W_d\hat d_j),\qquad G=W_e^\top W_d^{-1},
    \label{eq:mf-tie}
\end{equation}
with $G$ shared by all atoms, so encoder and decoder are tied up to one learned linear map. In a TopK SAE, $e_j$ and $W_{\text{dec}}[:,j]$ are independent.
Consequently the linear projections are redundant for a TopK SAE, since $W_{\text{in}}$ and $W_{\text{out}}$ can be absorbed into $e_j$ and $W_{\text{dec}}$, but not for the MF-SAE, where they are the only mechanism that lets the encoder differ from the decoder.
\end{enumerate}

\noindent
{\bf Relation to matching pursuit SAEs:} Subject to the non-negativity constraint in \eqref{eq:mf-outputs}, $w_j \hat d_j^\top$ is the least-squares estimate of $z$ when we use only the $j$-th atom. Thus, we may view this as the first step of non-negative matching pursuit \cite{mallat1993matching}, but applied to all atoms at once: every atom is scored against the same latent, with no residual update between selections. This is unlike MP-SAE \cite{costa2026flat}, which selects atoms sequentially against a residual. Sequential selection improves reconstruction on LLM activations but has not improved sparse probing or feature absorption \cite{chanin2026synthsaebench}, and it requires $k$ serial encoder passes.

\subsection{Training objective}
\label{sec:objective}

For a batch of $B$ tokens with MF-SAE inputs $x_b$ and reconstructions $\hat x_b$, the training loss is
\begin{equation}
    \mathcal L
    =\frac{1}{B\,m}\sum_{b=1}^{B}\big\|\hat x_b-x_b\big\|_2^2
    +\lambda\,\bar c,
    \qquad
    \bar c=\frac{1}{B\,k}\sum_{b=1}^{B}\sum_{j\in\mathcal S_b}c_{bj},
    \label{eq:loss}
\end{equation}
where $\mathcal S_b$ and $c_{bj}$ are the selected atoms and coefficients of token $b$ from \eqref{eq:mf-select} and \eqref{eq:mf-dec}, and $m$ is the dimension of the space in which the input lives. On raw activations, $x_b=h_b$ and $m=d$. On the live space, $x_b=h_{\text{live},b}$, the reconstruction is projected back onto the live space, and $m=r$, so the error is measured only on the live directions and normalized by their number. Since the coefficients are non-negative, $\bar c$ is the mean coefficient magnitude of the active atoms. Sparsity is enforced by the top-$k$ selection, so the auxiliary term only regularizes the scale of the coefficients and must be kept small: we use $\lambda=0.01$ in all experiments, as larger values collapse the dictionary to a small number of alive atoms.

\section{Experimental Setup}
\label{sec:setup}
\subsection{Models, Data, and Training}
\label{sec:setup-training}

\paragraph{Models.} We study Gemma-2-2B \cite{gemma2} ($d=2304$, 26 layers) at the residual stream after blocks 12 and 20. Results on GPT-2 Small \cite{radford2019gpt2} are reported in Appendix~\ref{app:gpt2}.

\paragraph{Jacobian.} For each layer, the averaged Jacobian \eqref{eq:jlens-matrix} is estimated once by reverse-mode automatic differentiation over $1000$ prompts truncated to $64$ tokens, with the final block as the target, and is shared by every SAE trained at that layer. Unless stated otherwise, the live space uses $r=1536$ directions ($768$ quiet).

\paragraph{Training.} SAEs are trained on text from the Pile \cite{gao2020pile} (uncopyrighted subset) in sequences of $128$ tokens. Activations from $2048$ sequences are pooled and shuffled, and each step uses a minibatch of $4096$ tokens drawn from the pool, so that a minibatch mixes tokens from many contexts. We train for $7000$ steps (about $28.7$M tokens) with AdamW \cite{loshchilov2019adamw}, learning rate $2\times10^{-4}$, no weight decay, and $\lambda=0.01$. Evaluation uses a held-out partition of the same corpus.

\paragraph{Dictionary and sparsity.} At each layer we train both the raw and the live MF-SAE with $n\in\{16384, 32768\}$ atoms (16K and 32K) and $k\in\{40,80,160,320,640\}$ active atoms per token.

\subsection{Baselines}
\label{sec:setup-baselines}

\paragraph{Dense activations.} As SAE-free references, we probe the coordinates of the raw activation $h$ and of the shaped activation $h_{\text{live}}$ directly. The latter isolates the effect of shaping from that of sparse coding.

\paragraph{SAEBench TopK SAEs.} SAEBench \cite{karvonen2025saebench} provides TopK SAEs \cite{gao2025scaling} on Gemma-2-2B with $16$K latents at $k\in\{40,80,160,320,640\}$, matching our sparsity grid. We use two checkpoints of each: one after about $30$M training tokens, which matches the $28.7$M tokens seen by our SAEs, and the final one after $300$M tokens.

\paragraph{Gemma Scope.} We compare against the residual-stream JumpReLU SAEs \cite{rajamanoharan2024jumprelu} of Gemma Scope \cite{lieberum2024gemmascope} at widths 16K and 32K. Their sparsity is set by a learned threshold rather than a fixed $k$, so we pair each with the MF-SAE of nearest average L0 (Table~\ref{tab:baselines-gemma}).

\paragraph{Matryoshka SAE.} We include the Matryoshka BatchTopK SAE \cite{bussmann2025matryoshka} with $32$K latents and L0~$=40$, the only configuration available for Gemma-2-2B.

\subsection{Evaluation}
\label{sec:setup-eval}

\paragraph{SAEBench sparse probing.} Following SAEBench \cite{karvonen2025saebench}, we evaluate on 36 binary classification tasks from 8 datasets (Bias in Bios, Amazon reviews and sentiment, GitHub code, AG News, and Europarl), with 4000 training and 1000 test examples per task and a context length of 128 tokens. SAE features are mean-pooled over tokens, the $k_{\text{probe}}$ features with the largest difference in class means are selected, and a logistic-regression probe is trained on them. We report test accuracy averaged over tasks for $k_{\text{probe}}\in\{1,2,4,8,16,32\}$, and refer to it as SB-Acc.

\paragraph{SAE Probes.} To test probing on a broader set of tasks, we use the benchmark of \cite{kantamneni2025saeprobe}, with 113 binary datasets in its standard data setting, and report test accuracy averaged over datasets for probes on the top-$k_{\text{probe}}$ SAE features.

\paragraph{Reconstruction and downstream loss.} We report the fraction of variance explained (PVE) of the MF-SAE input $x$, the fraction of alive atoms (atoms that are active on at least one evaluation token), and the cross-entropy loss recovered when the activation is replaced by its reconstruction,
\begin{equation}
    \text{CE recovered}=\frac{\mathrm{CE}_{\text{zero}}-\mathrm{CE}_{\text{rec}}}{\mathrm{CE}_{\text{zero}}-\mathrm{CE}_{\text{clean}}},
    \label{eq:ce-rec}
\end{equation}
where $\mathrm{CE}_{\text{zero}}$ is obtained by zero-ablating the activation and $\mathrm{CE}_{\text{clean}}$ is the loss of the unmodified model. For the live MF-SAE we patch the model in three ways: \emph{live only}, replacing $h$ with $\hat x$; \emph{live+quiet}, replacing $h$ with $\hat x+(h-h_{\text{live}})$, which restores the exact quiet component; and \emph{projection only}, replacing $h$ with $h_{\text{live}}$ without any SAE, which measures the cost of dropping the quiet component and upper-bounds the live-only value. Since the live PVE is computed on $h_{\text{live}}$ rather than $h$, it is not directly comparable to the raw PVE.

\begin{figure*}[h]
    \centering
    \begin{subfigure}[b]{0.48\textwidth}
        \centering
        \includegraphics[width=\linewidth]{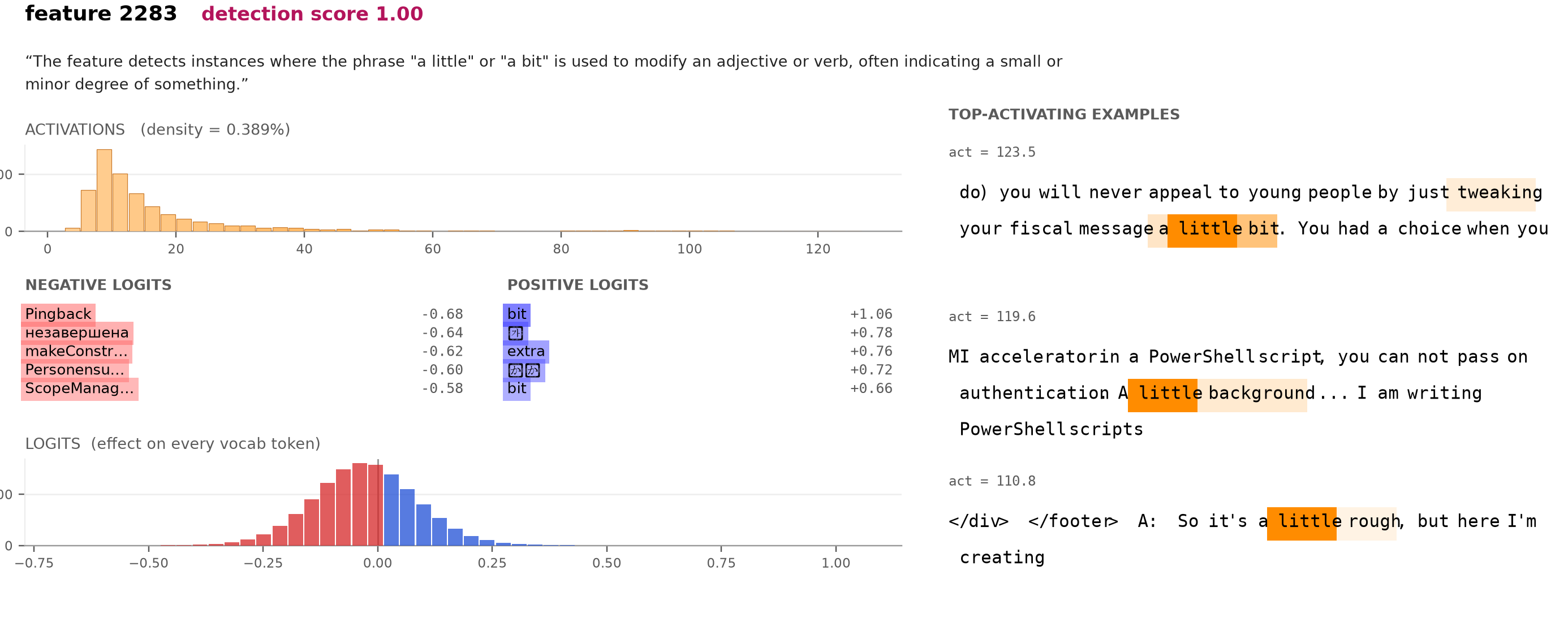}
        \caption{\scriptsize{Raw, feature 2283: ``a little bit''}}
        \label{fig:dash-a}
    \end{subfigure}
    \hfill
    \begin{subfigure}[b]{0.48\textwidth}
        \centering
        \includegraphics[width=\linewidth]{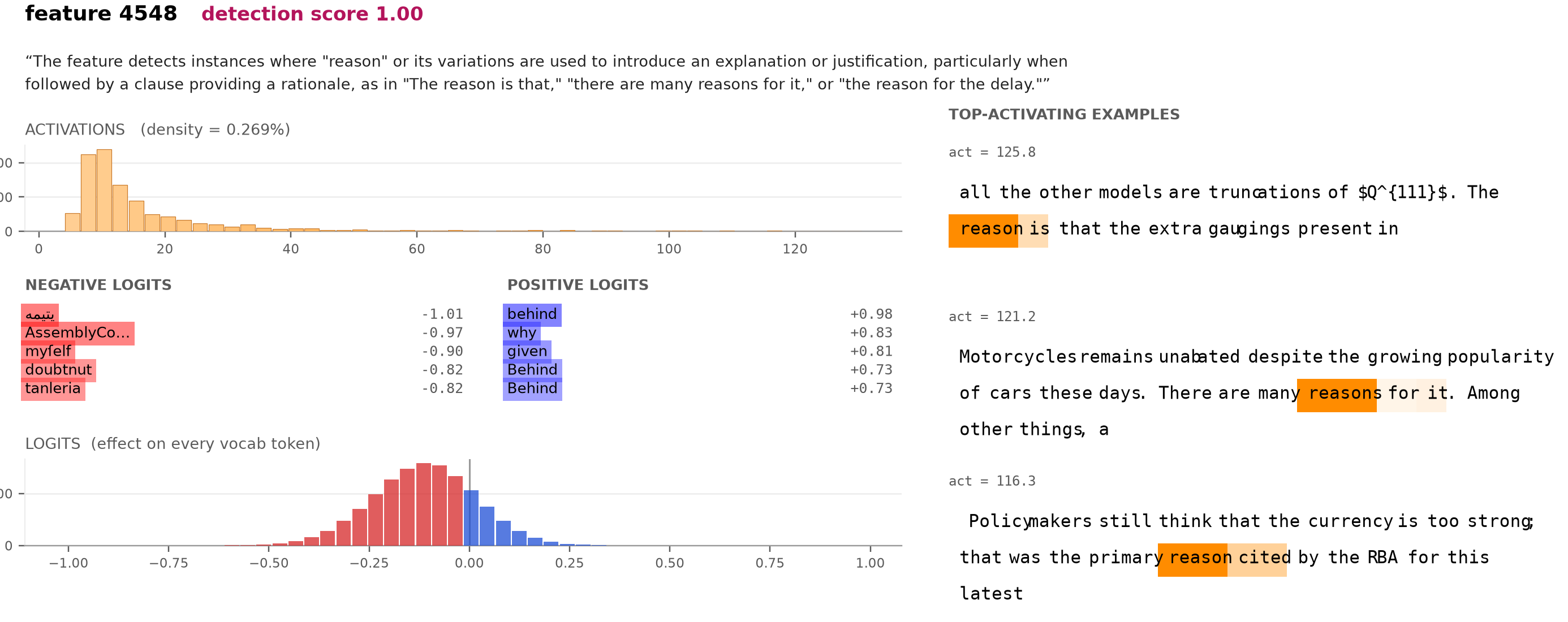}
        \caption{\scriptsize{Raw, feature 4548: ``the reason behind''}}
        \label{fig:dash-b}
    \end{subfigure}

    \vspace{6pt}

    \begin{subfigure}[b]{0.48\textwidth}
        \centering
        \includegraphics[width=\linewidth]{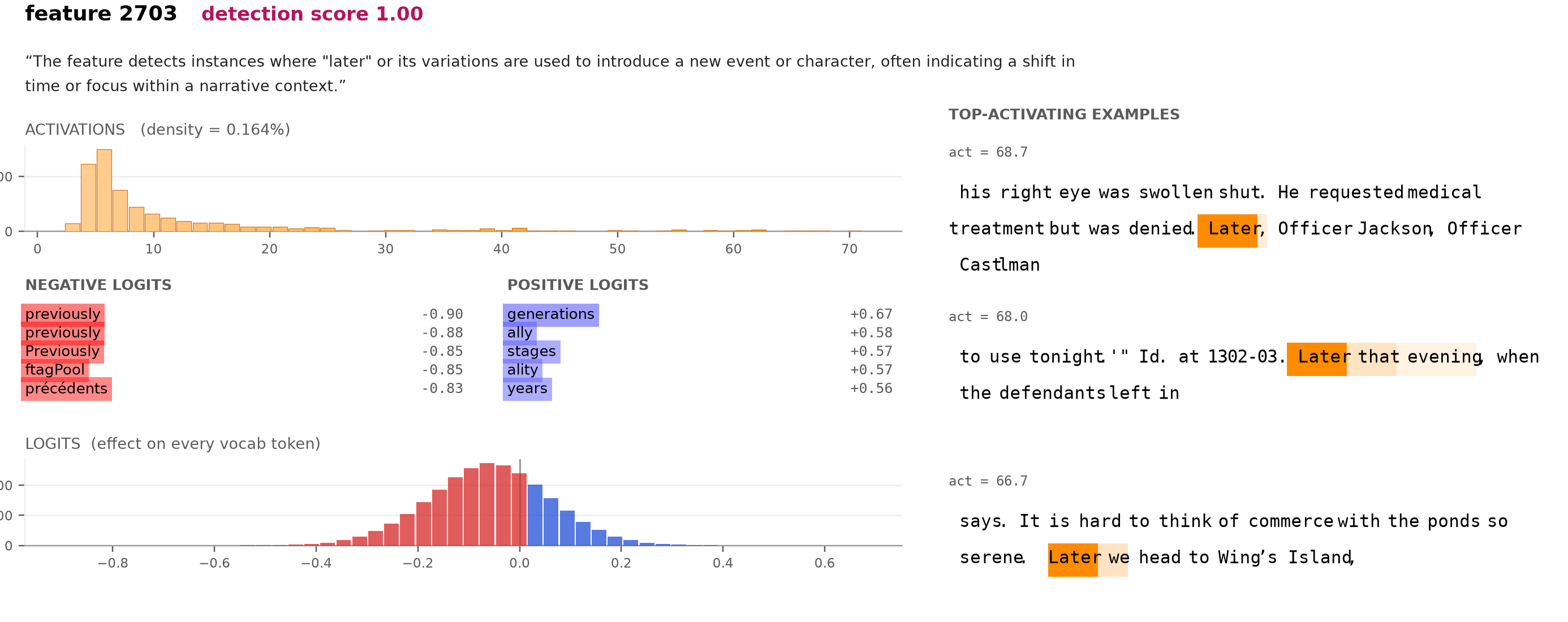}
        \caption{\scriptsize{Live, feature 2703: ``later''}}
        \label{fig:dash-c}
    \end{subfigure}
    \hfill
    \begin{subfigure}[b]{0.48\textwidth}
        \centering
        \includegraphics[width=\linewidth]{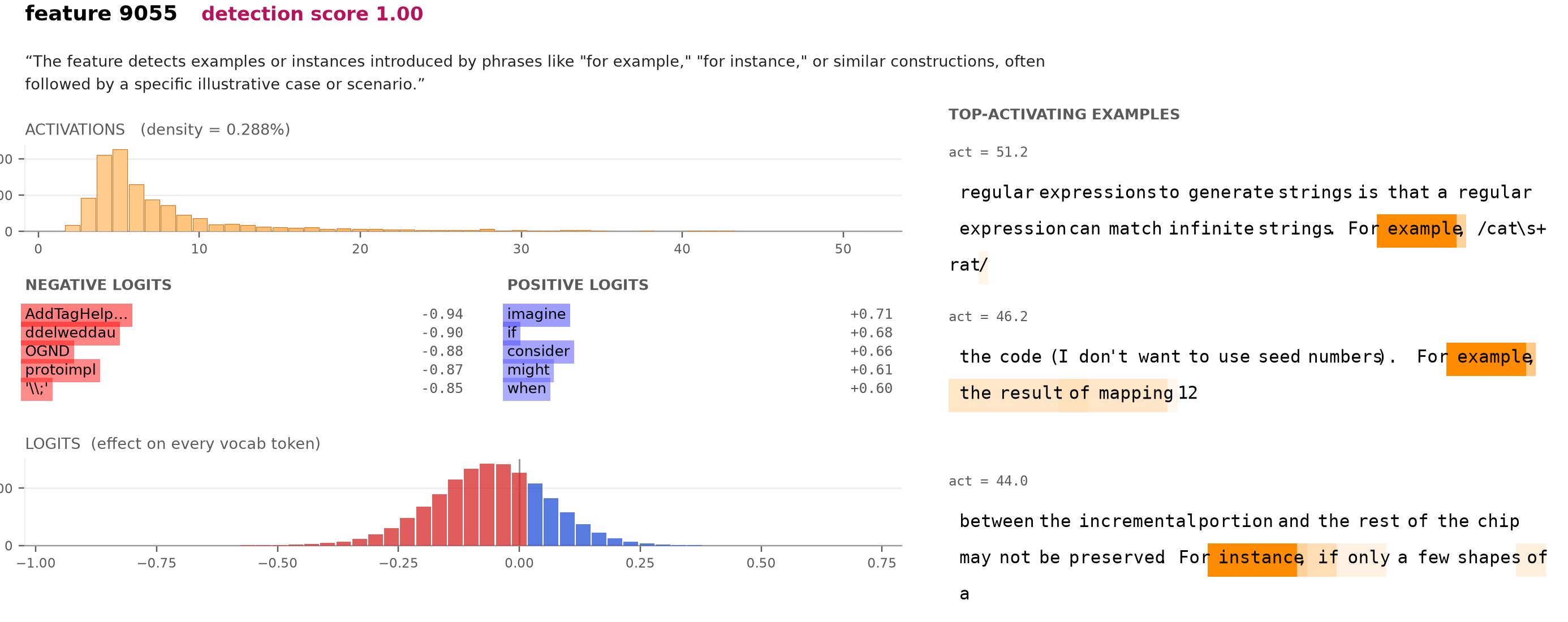}
        \caption{\scriptsize{Live, feature 9055: ``example''}}
        \label{fig:dash-d}
    \end{subfigure}

    \caption{\scriptsize{\textbf{Example feature dashboards (Gemma-2-2B, layer 20, 16K dictionary).} \textbf{(a,~b)} Features of the raw MF-SAE. \textbf{(c,~d)} Features of the live MF-SAE. Each dashboard shows the feature's top-activating contexts together with its LLM-generated explanation. Dashboards for all features with detection score above $0.7$, for layer 12 and other models, are available at the link in Section~\ref{sec:setup-eval}.}}
    \label{fig:dashboards}
\end{figure*}

\paragraph{Automated interpretability.} For each feature we collect its top-activating windows on $6000$ sequences of $128$ tokens from the Pile. Qwen2.5-7B-Instruct \cite{qwen2025qwen25} is shown $12$ activating windows and writes a one-line explanation. Given only the explanation, the same model then classifies $8$ held-out activating windows against $8$ top windows of randomly chosen other features, and we report the balanced accuracy as the detection score. Features with fewer than $20$ activating windows are not scored. Figure~\ref{fig:dashboards} shows example dashboards of features from layer 20. Dashboards for all features with detection score above $0.7$, including layer 12 and other models, are available at \url{https://huggingface.co/spaces/user1378916/MF-SAE/tree/main}.

\section{Results and Analysis}
\label{sec:results}
\subsection{Shaping without an SAE}
\label{sec:res-shaping}

\begin{figure*}[t]
    \centering
    \begin{subfigure}[b]{0.48\textwidth}
        \centering
        \includegraphics[width=\linewidth, height=0.5\linewidth]{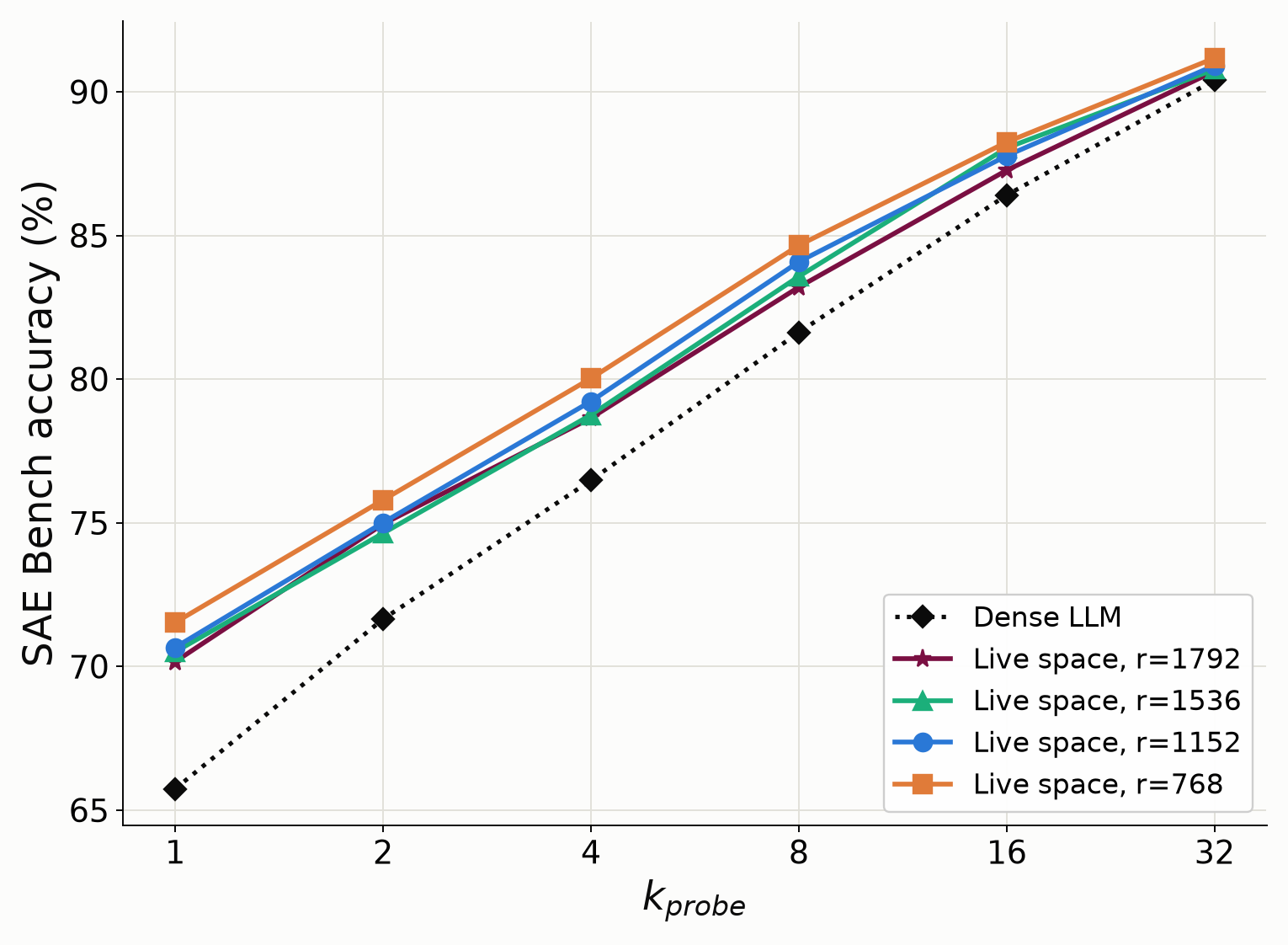}
        \caption{\scriptsize{SAEBench, layer 12}}
        \label{fig:nosae-a}
    \end{subfigure}
    \hfill
    \begin{subfigure}[b]{0.48\textwidth}
        \centering
        \includegraphics[width=\linewidth, height=0.5\linewidth]{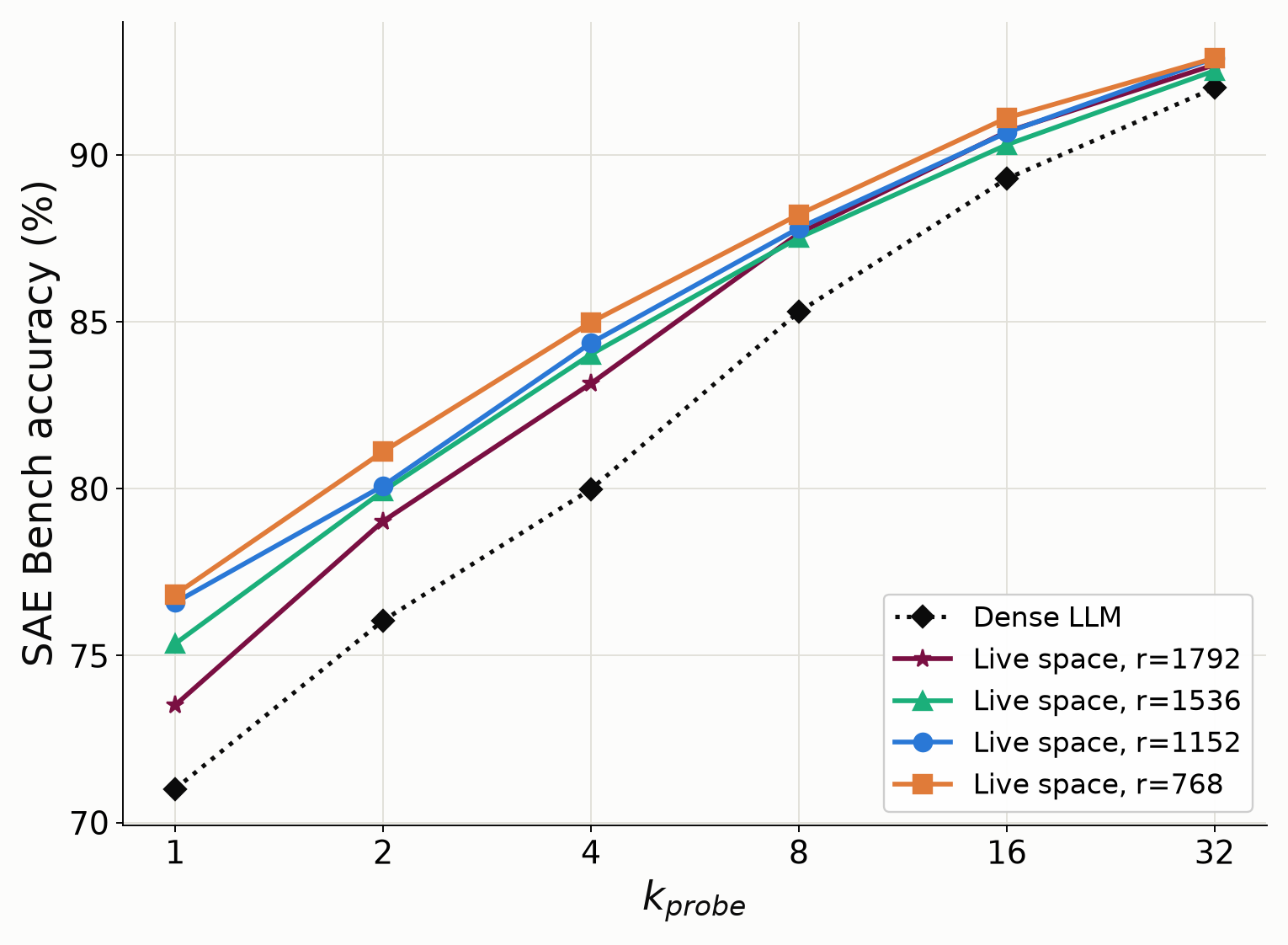}
        \caption{\scriptsize{SAEBench, layer 20}}
        \label{fig:nosae-b}
    \end{subfigure}

    \vspace{6pt}

    \begin{subfigure}[b]{0.48\textwidth}
        \centering
        \includegraphics[width=\linewidth, height=0.5\linewidth]{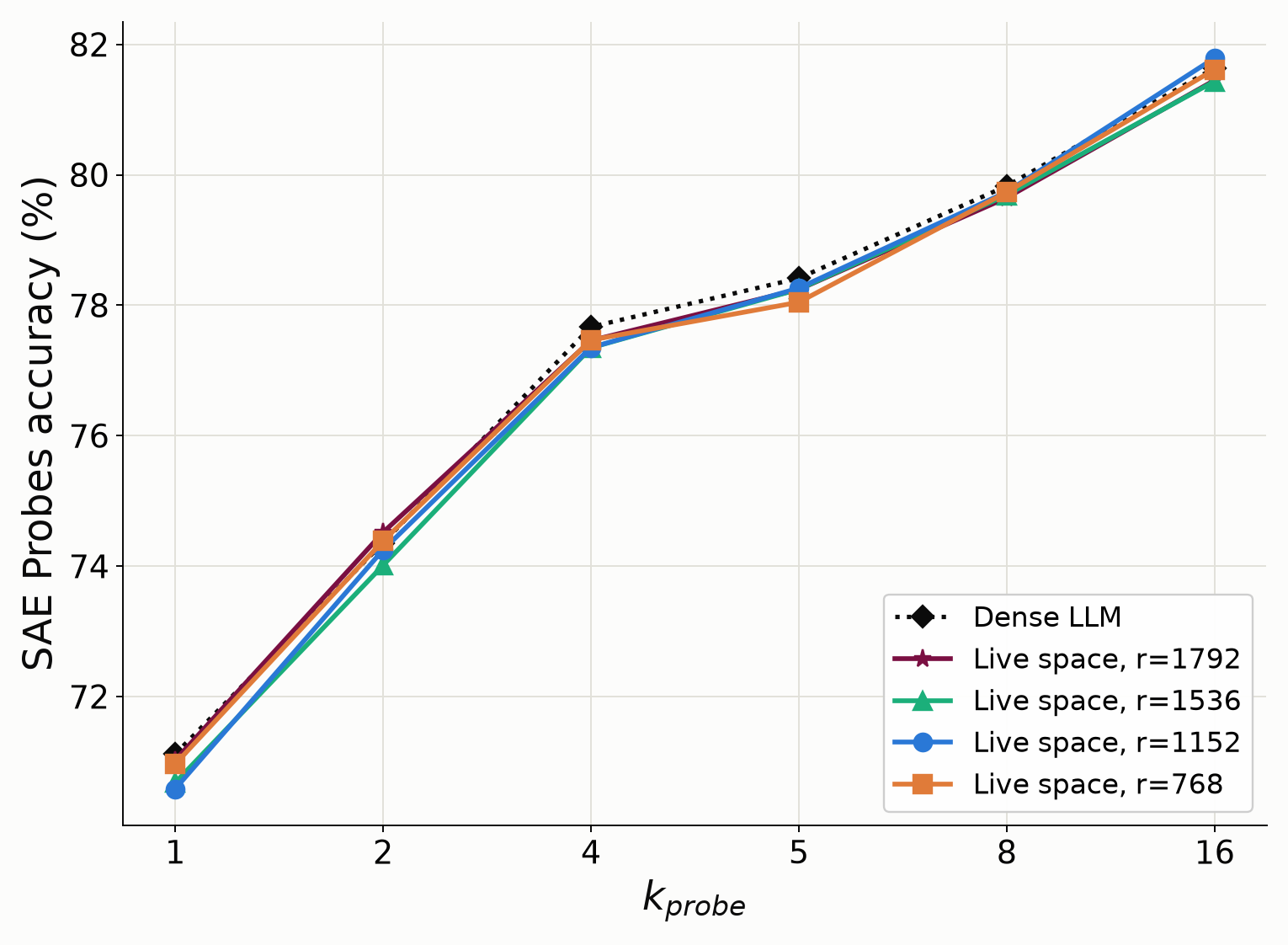}
        \caption{\scriptsize{SAE Probes, layer 12}}
        \label{fig:nosae-c}
    \end{subfigure}
    \hfill
    \begin{subfigure}[b]{0.48\textwidth}
        \centering
        \includegraphics[width=\linewidth, height=0.5\linewidth]{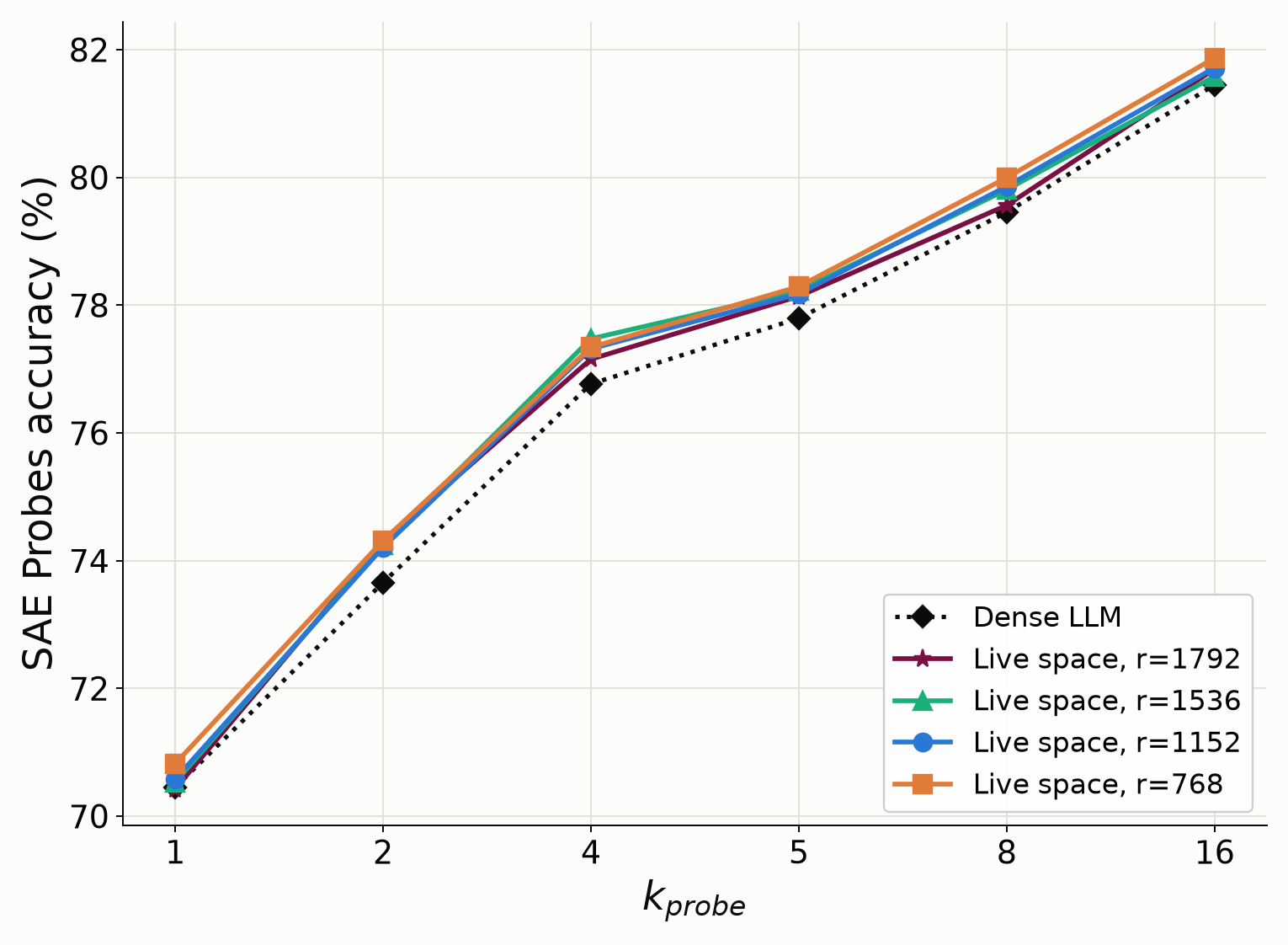}
        \caption{\scriptsize{SAE Probes, layer 20}}
        \label{fig:nosae-d}
    \end{subfigure}

    \caption{\scriptsize{\textbf{Probing shaped activations without an SAE (Gemma-2-2B).} Probe accuracy against $k_{\text{probe}}$ for the raw activation $h$ and for the shaped activation $h_{\text{live}}$ with $r\in\{768,1152,1536,1792\}$ live directions. \textbf{(a,~b)} On SAEBench, $h_{\text{live}}$ outperforms $h$ at both layers and for every $r$. \textbf{(c,~d)} On SAE Probes, $h_{\text{live}}$ is comparable to $h$ at layer 12 and slightly better at layer 20.}}
    \label{fig:NoSAE}
\end{figure*}

Before training any SAE, we test whether shaping preserves the information that probes rely on. We apply both probing benchmarks directly to the coordinates of the shaped activation $h_{\text{live}}$ for $r\in\{768,1152,1536,1792\}$ and compare with the raw activation $h$ (Figure~\ref{fig:NoSAE}).

On SAEBench, $h_{\text{live}}$ outperforms $h$ at both layers and for every live dimension $r$ (Figures~\ref{fig:nosae-a} and~\ref{fig:nosae-b}). Removing the quiet directions therefore does not discard the information these tasks need; it makes that information easier to find for a probe restricted to a few coordinates. On SAE Probes, $h_{\text{live}}$ is comparable to $h$ at layer 12 and slightly better at layer 20 (Figures~\ref{fig:nosae-c} and~\ref{fig:nosae-d}). Across both benchmarks, shaping never hurts probing, and the result is not sensitive to the choice of $r$. We use $r=1536$ throughout.

\begin{figure*}[t]
    \centering
    \begin{subfigure}[b]{0.48\textwidth}
        \centering
        \includegraphics[width=\linewidth, height=0.5\linewidth]{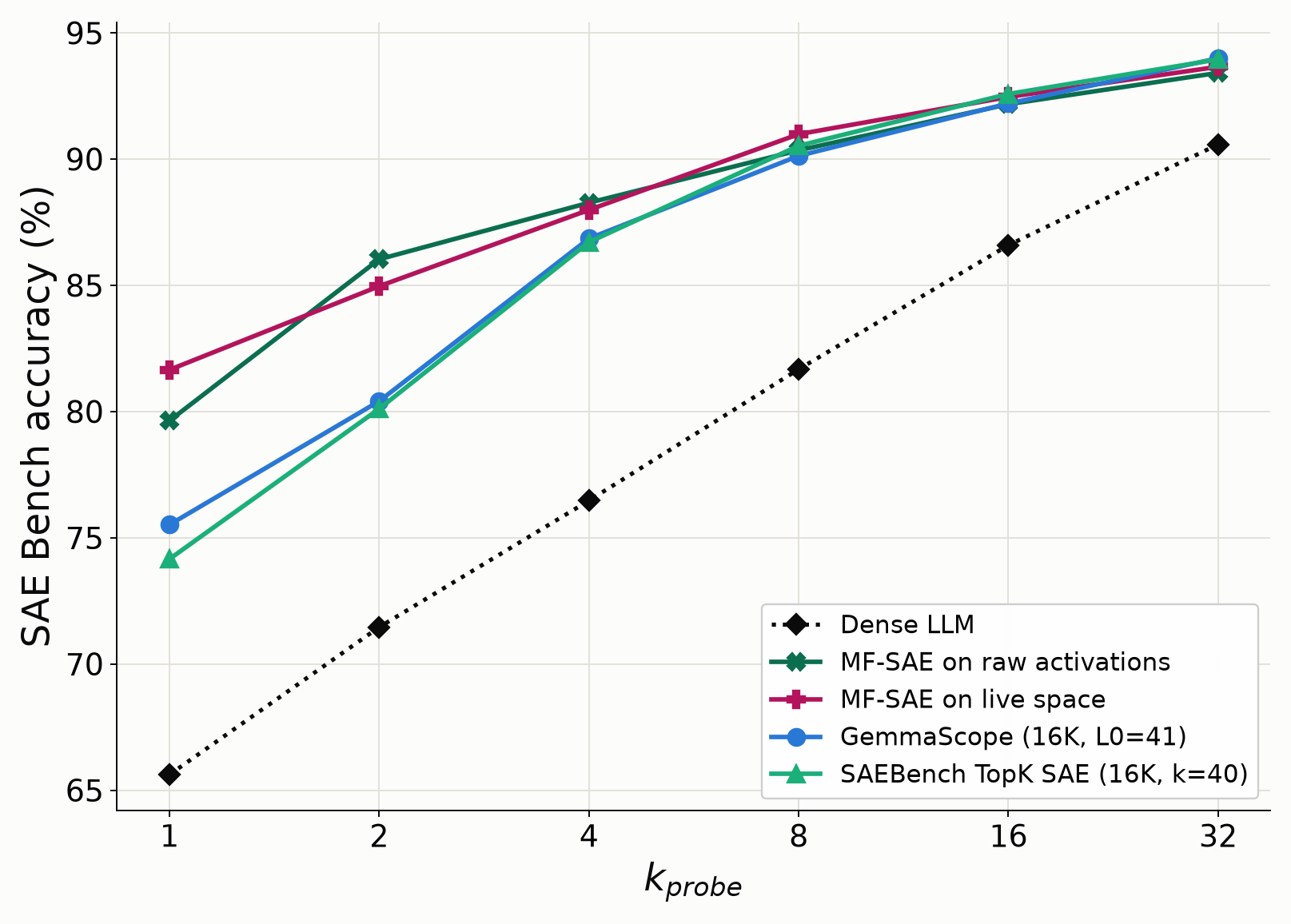}
        \caption{\scriptsize{Layer 12, $k=40$}}
        \label{fig:saebench-a}
    \end{subfigure}
    \hfill
    \begin{subfigure}[b]{0.48\textwidth}
        \centering
        \includegraphics[width=\linewidth, height=0.5\linewidth]{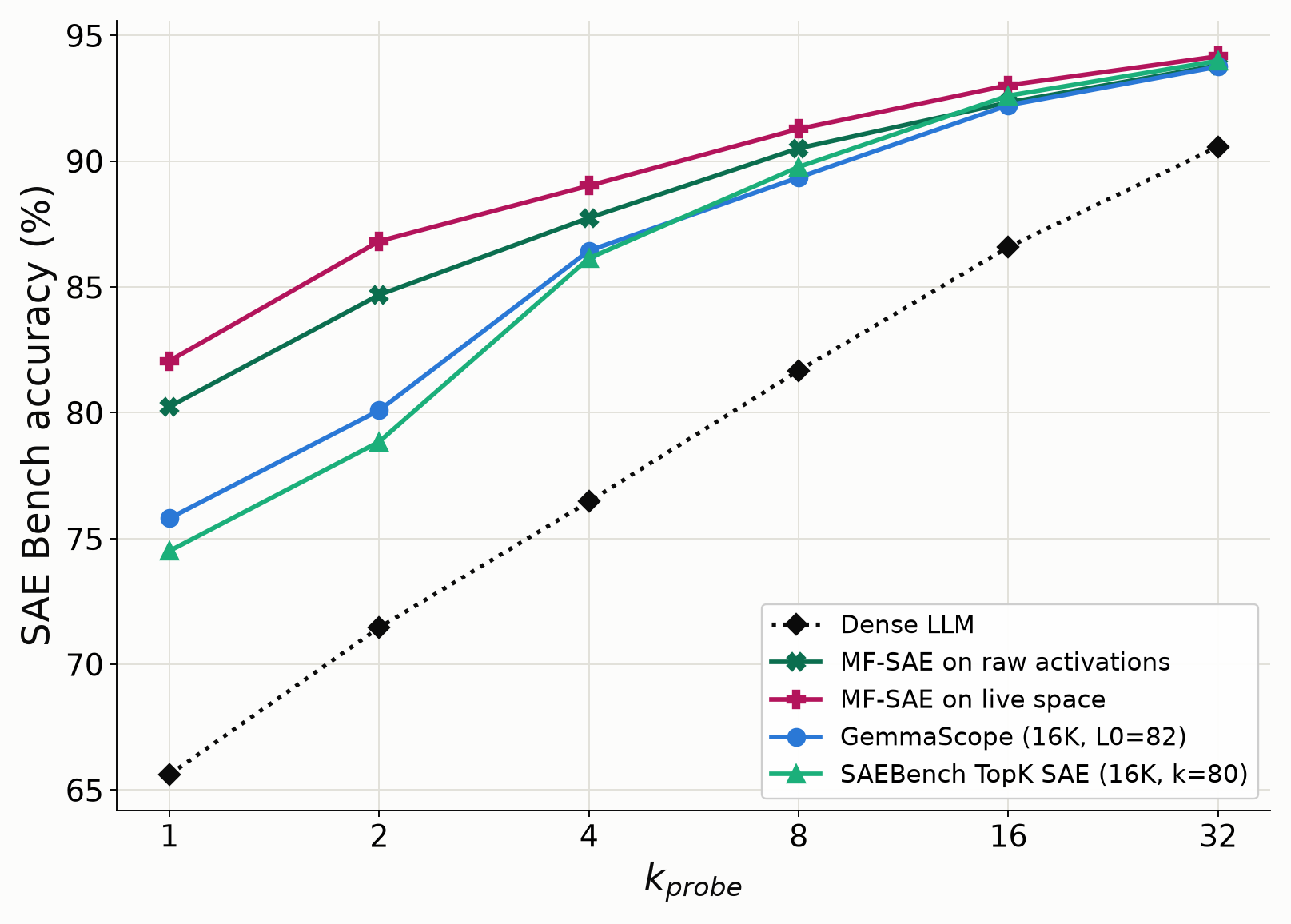}
        \caption{\scriptsize{Layer 12, $k=80$}}
        \label{fig:saebench-b}
    \end{subfigure}

    \vspace{6pt}

    \begin{subfigure}[b]{0.48\textwidth}
        \centering
        \includegraphics[width=\linewidth, height=0.5\linewidth]{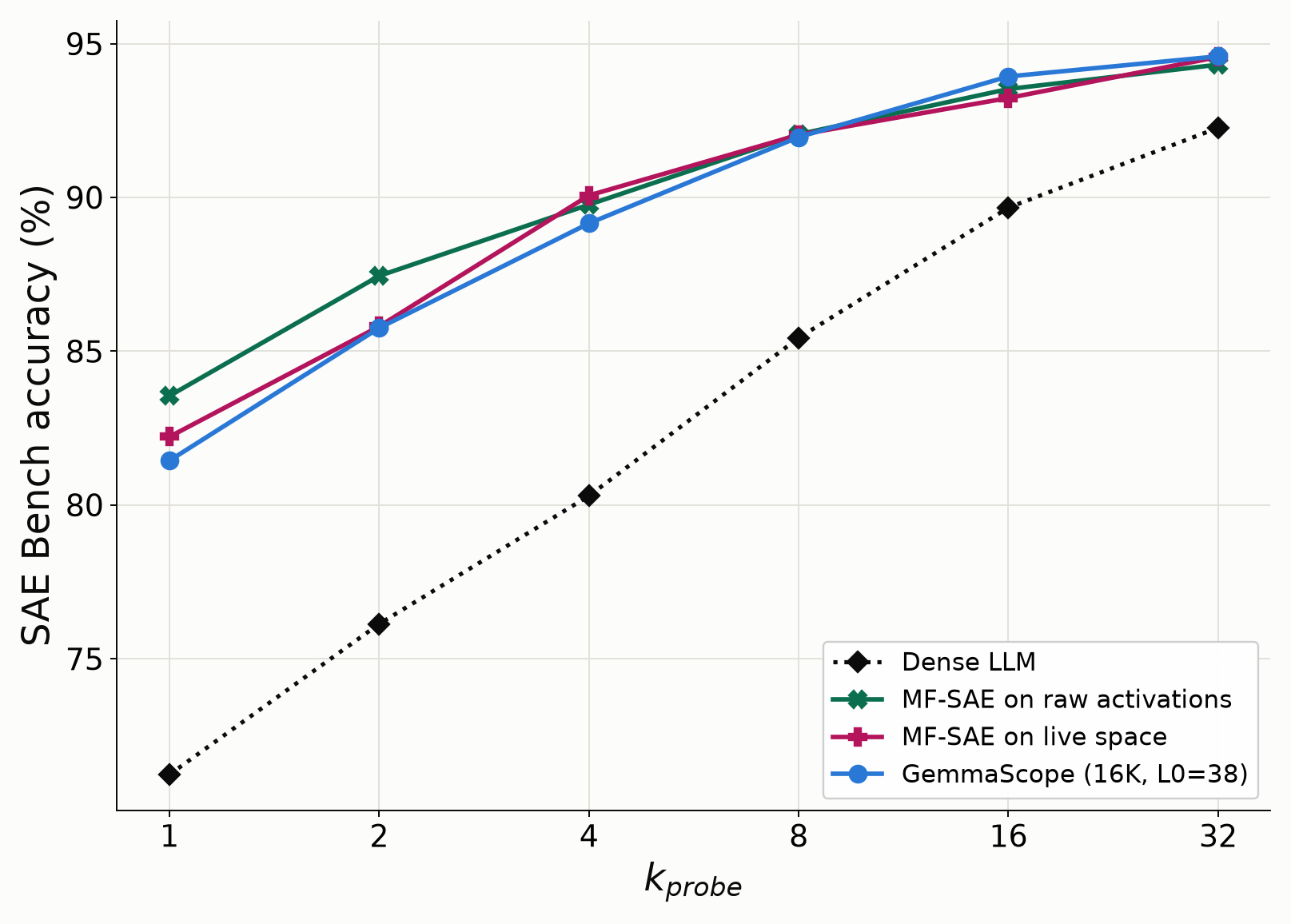}
        \caption{\scriptsize{Layer 20, $k=40$}}
        \label{fig:saebench-c}
    \end{subfigure}
    \hfill
    \begin{subfigure}[b]{0.48\textwidth}
        \centering
        \includegraphics[width=\linewidth, height=0.5\linewidth]{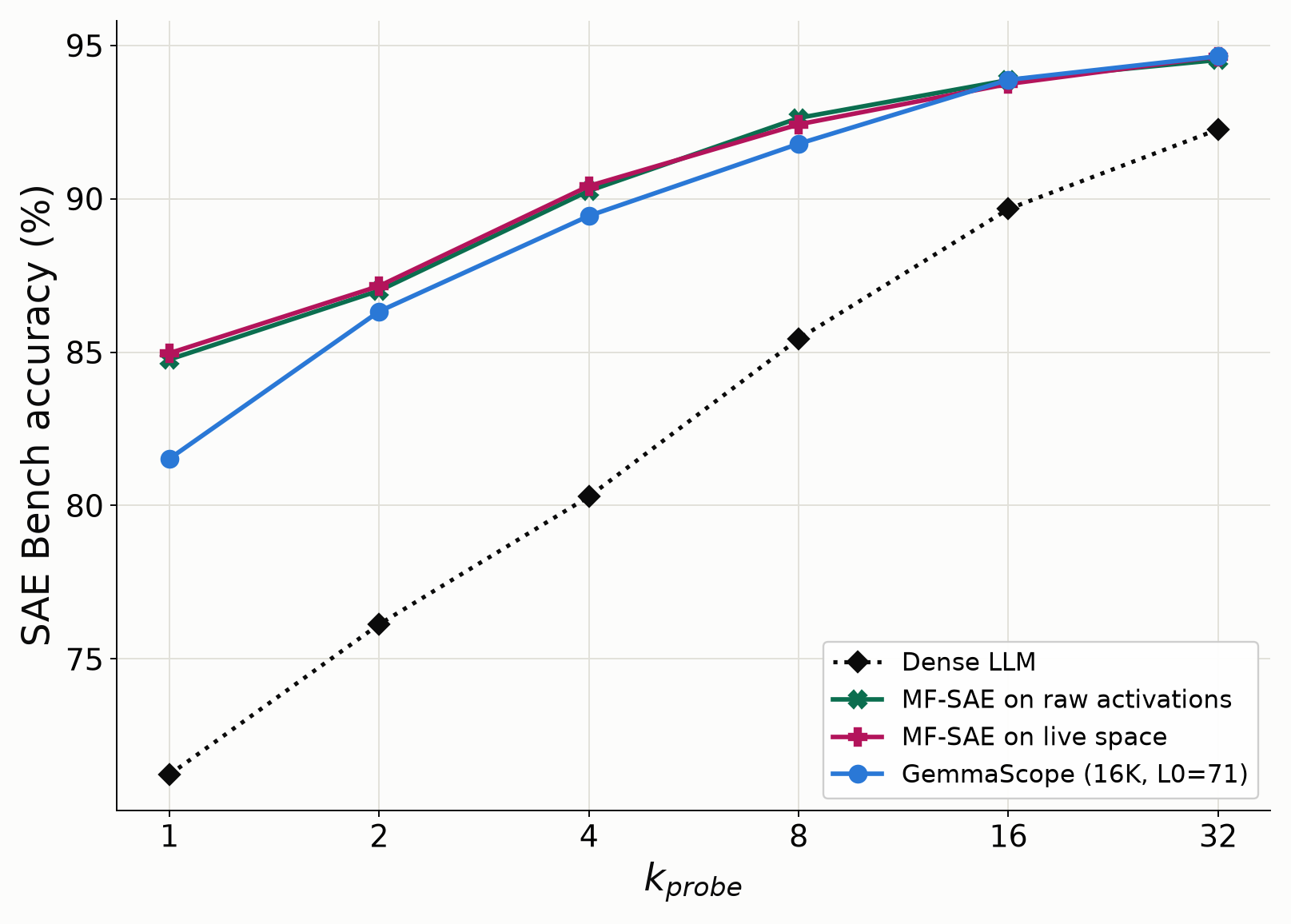}
        \caption{\scriptsize{Layer 20, $k=80$}}
        \label{fig:saebench-d}
    \end{subfigure}

    \caption{\scriptsize{\textbf{SAEBench sparse probing (Gemma-2-2B, 16K dictionary).} SB-Acc against $k_{\text{probe}}$ for the raw and live MF-SAEs and the available baseline SAEs. Rows are layers 12 and 20; columns are sparsity $k=40$ and $k=80$. Gemma Scope has no fixed sparsity, and the SAE with the closest average L0 is shown. Both MF-SAEs outperform the baselines in every panel. Raw and live are similar except in~(b), where the live MF-SAE is consistently better.}}
    \label{fig:SAEBench}
\end{figure*}

\begin{figure*}[t]
    \centering
    \begin{subfigure}[b]{0.48\textwidth}
        \centering
        \includegraphics[width=\linewidth, height=0.5\linewidth]{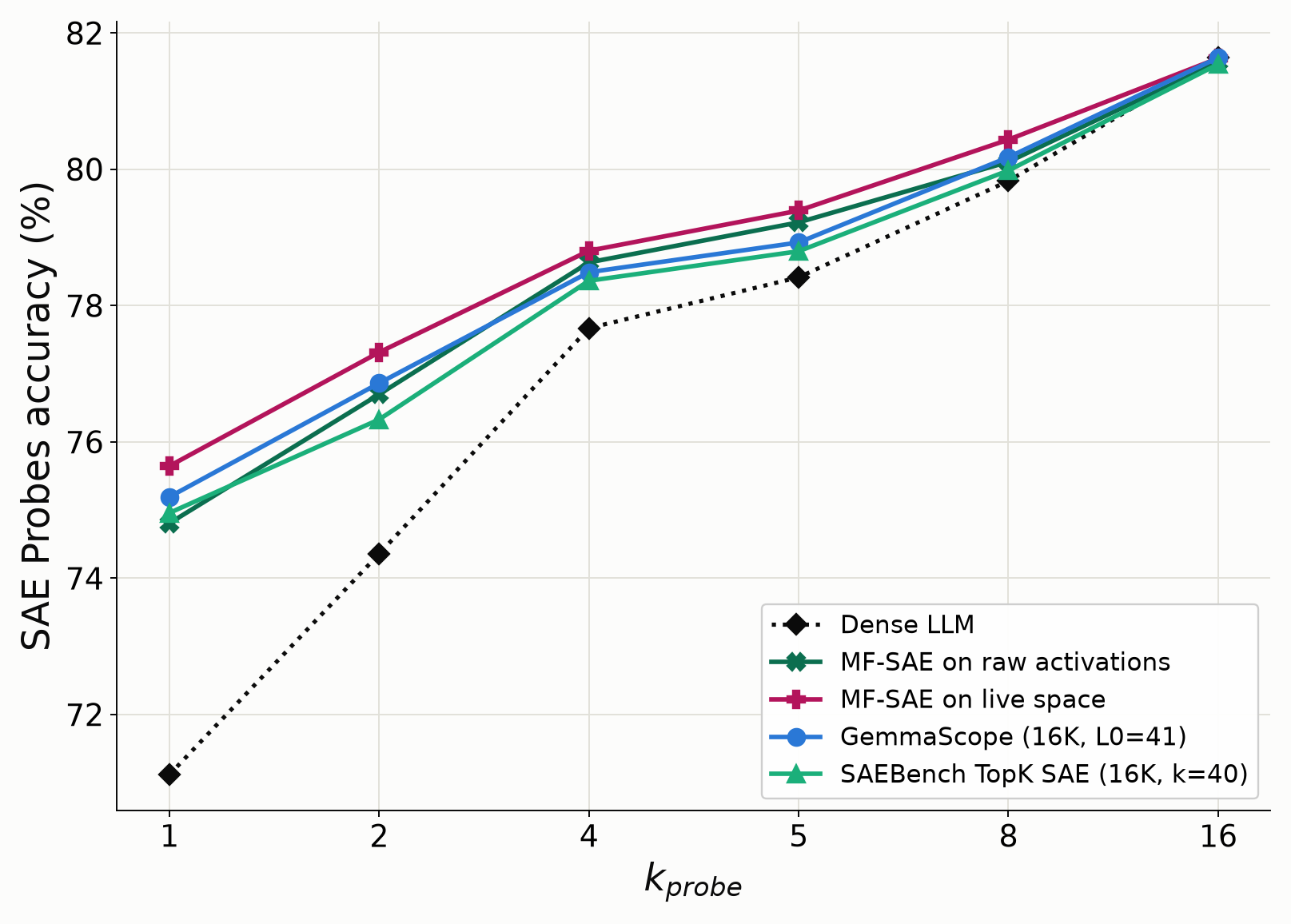}
        \caption{\scriptsize{Layer 12, $k=40$}}
        \label{fig:saeprobes-a}
    \end{subfigure}
    \hfill
    \begin{subfigure}[b]{0.48\textwidth}
        \centering
        \includegraphics[width=\linewidth, height=0.5\linewidth]{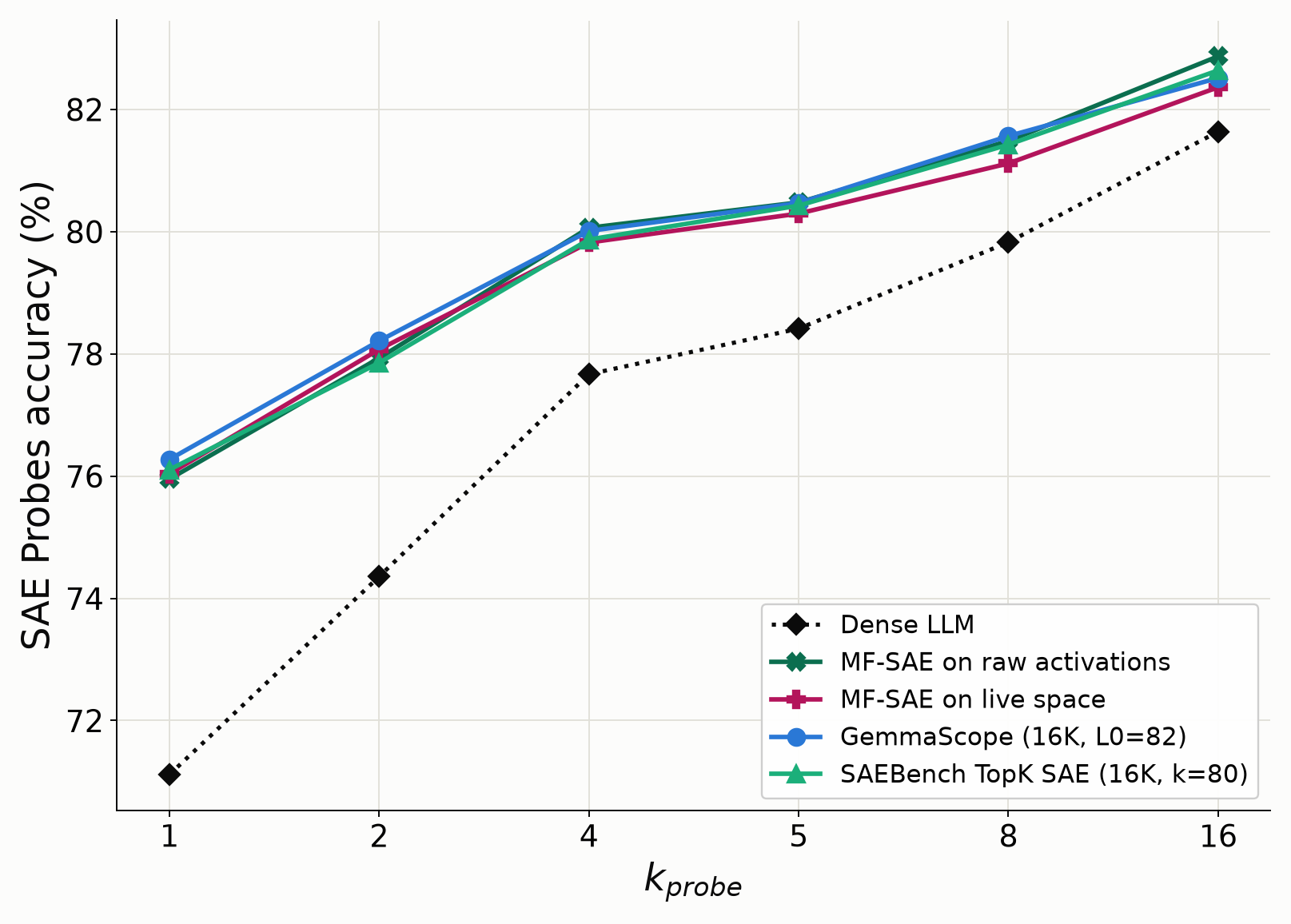}
        \caption{\scriptsize{Layer 12, $k=80$}}
        \label{fig:saeprobes-b}
    \end{subfigure}

    \vspace{6pt}

    \begin{subfigure}[b]{0.48\textwidth}
        \centering
        \includegraphics[width=\linewidth, height=0.5\linewidth]{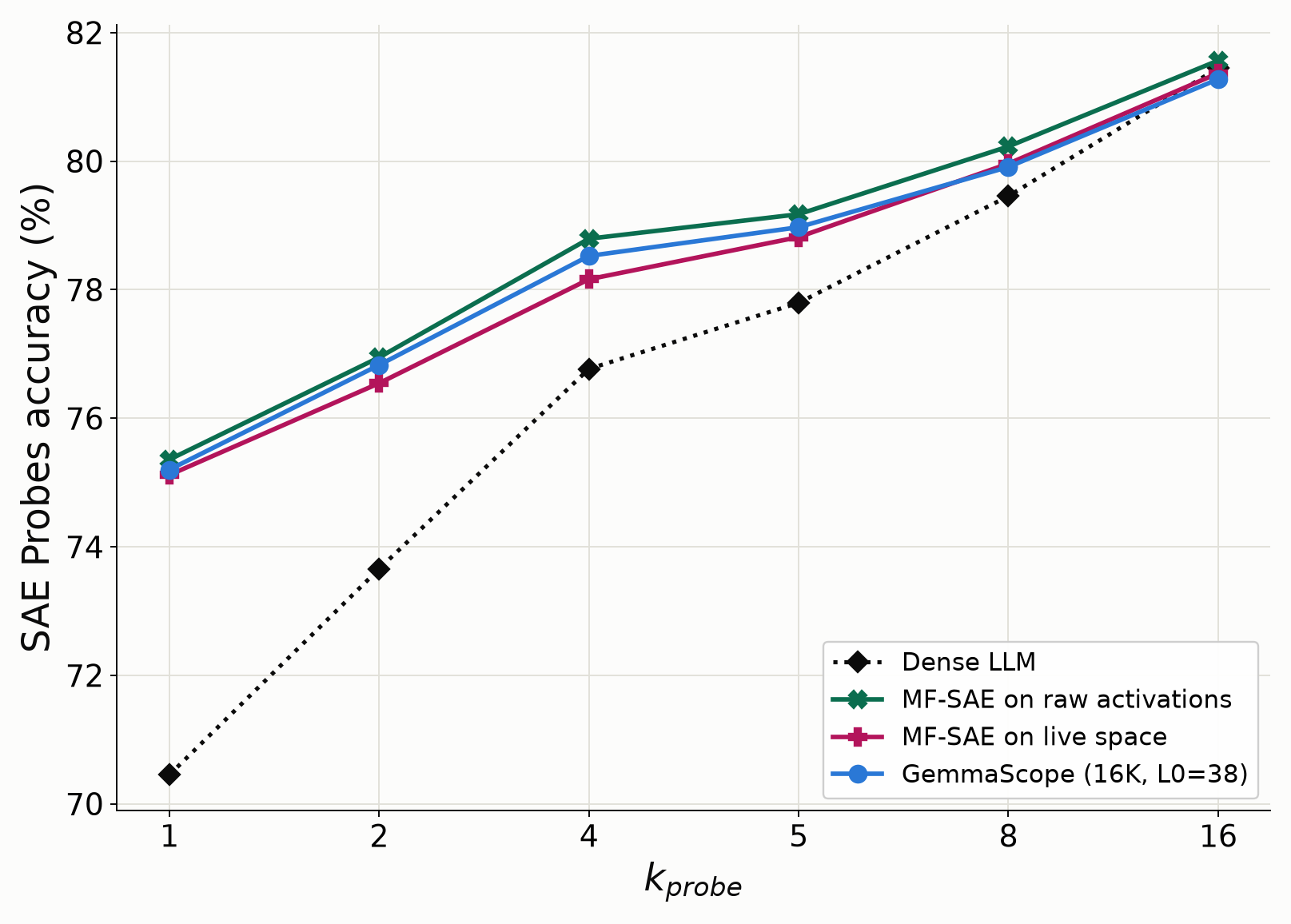}
        \caption{\scriptsize{Layer 20, $k=40$}}
        \label{fig:saeprobes-c}
    \end{subfigure}
    \hfill
    \begin{subfigure}[b]{0.48\textwidth}
        \centering
        \includegraphics[width=\linewidth, height=0.5\linewidth]{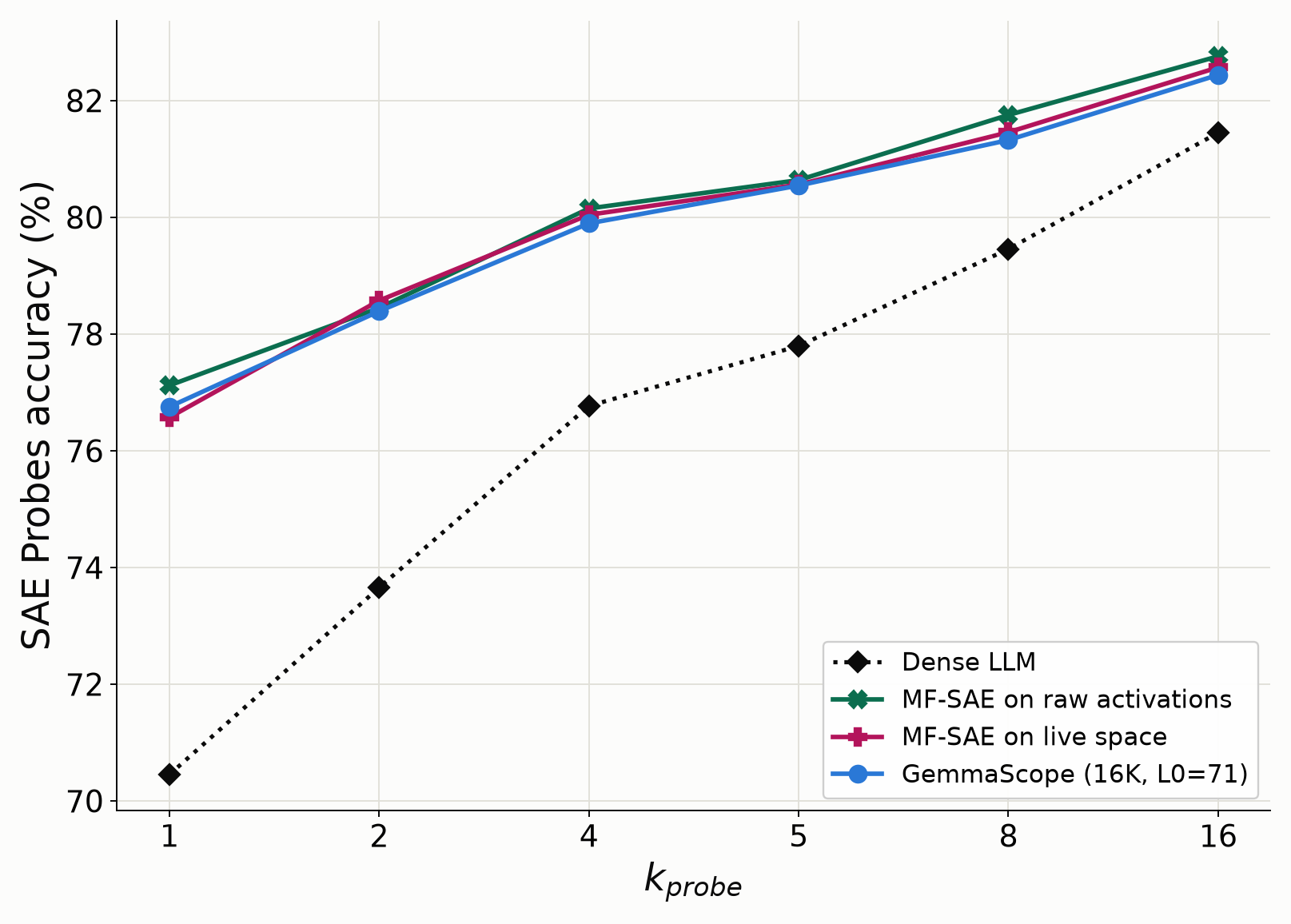}
        \caption{\scriptsize{Layer 20, $k=80$}}
        \label{fig:saeprobes-d}
    \end{subfigure}

    \caption{\scriptsize{\textbf{SAE Probes (Gemma-2-2B, 16K dictionary).} Mean test accuracy over the 113 datasets of \citet{kantamneni2025saeprobe} against $k_{\text{probe}}$, with the same layout and methods as Figure~\ref{fig:SAEBench}. The MF-SAEs improve over the baselines only marginally, and raw and live perform similarly.}}
    \label{fig:SAEProbes}
\end{figure*}

\subsection{Shaping without an SAE}
\label{sec:res-shaping}

Before training any SAE, we test whether shaping preserves the information that probes rely on. We apply both probing benchmarks directly to the coordinates of the shaped activation $h_{\text{live}}$ for $r\in\{768,1152,1536,1792\}$ and compare with the raw activation $h$ (Figure~\ref{fig:NoSAE}).

On SAEBench, $h_{\text{live}}$ outperforms $h$ at both layers and for every live dimension $r$ (Figures~\ref{fig:nosae-a} and~\ref{fig:nosae-b}). Removing the quiet directions therefore does not discard the information these tasks need; it makes that information easier to find for a probe restricted to a few coordinates. On SAE Probes, $h_{\text{live}}$ is comparable to $h$ at layer 12 and slightly better at layer 20 (Figures~\ref{fig:nosae-c} and~\ref{fig:nosae-d}). Across both benchmarks, shaping never hurts probing, and the result is not sensitive to the choice of $r$. We use $r=1536$ throughout.

\subsection{Sparse Probing}
\label{sec:res-probing}

We compare the raw and live MF-SAEs with the baseline SAEs of Section~\ref{sec:setup-baselines} at sparsity $k=40$ and $k=80$ with a 16K dictionary, at layers 12 and 20. For Gemma Scope, sparsity is not fixed; we use the SAE whose average L0 is closest to $k$, so this comparison is approximate.

\paragraph{SAEBench.} Both MF-SAEs outperform the available baselines at both layers and both sparsity levels (Figure~\ref{fig:SAEBench}). The raw and live MF-SAEs perform similarly in most settings. The exception is layer 12 at $k=80$, where the live MF-SAE is consistently better than the raw one across $k_{\text{probe}}$ (Figure~\ref{fig:saebench-b}). Most of the improvement over the baselines therefore comes from the MF-SAE itself, with shaping adding a further gain in some settings. The benefit of shaping without an SAE (Section~\ref{sec:res-shaping}) thus carries over only partially once sparse coding is applied.

\paragraph{SAE Probes.} On the 113 datasets of \cite{kantamneni2025saeprobe}, the MF-SAEs improve over the baselines only marginally, and the raw and live MF-SAEs are again similar (Figure~\ref{fig:SAEProbes}). This is consistent with the finding of \cite{kantamneni2025saeprobe} that SAE features offer little advantage on this broader benchmark, and indicates that the gains on SAEBench do not transfer fully to it.

\subsection{Reconstruction and Downstream Loss}
\label{sec:res-recon}

Table~\ref{tab:recon} reports the reconstruction metrics of Section~\ref{sec:setup-eval} for the raw and live MF-SAEs with a 16K dictionary. Since the live PVE is computed on $h_{\text{live}}$ rather than $h$, PVE is comparable only within each of the raw and live groups. For the live MF-SAE, the gap between \emph{live only} and \emph{live+quiet} measures the contribution of the discarded quiet component, while \emph{projection only} upper-bounds the live-only value. The raw MF-SAE recovers 97--99\% of the CE loss. The live MF-SAE recovers 89--94\% when patched alone, close to the 92--95\% bound set by dropping the quiet component without any SAE (projection only), so most of this gap comes from the projection rather than from sparse coding. Restoring the quiet component recovers 99\% in every setting. The live MF-SAE also uses more of its dictionary, with 81--99\% of atoms alive compared with 62--89\% for the raw MF-SAE, and the fraction of alive atoms increases with $k$ for both.

\begin{table}[t]
    \centering
    \caption{\textbf{Reconstruction and downstream loss (Gemma-2-2B, 16K dictionary).} PVE of the MF-SAE input, fraction of alive atoms, and CE loss recovered \eqref{eq:ce-rec}. For the live MF-SAE, CE recovered is reported for the three patching modes of Section~\ref{sec:setup-eval}: live only (L), live+quiet (L+Q), and projection only (P).}
    \label{tab:recon}
    \small
    \setlength{\tabcolsep}{3.5pt}
    \begin{tabular}{cclcccc}
        \toprule
        & & & & & \multicolumn{2}{c}{CE recovered} \\
        \cmidrule(lr){6-7}
        Layer & $k$ & Input & PVE & Alive & Raw / L & L+Q \,|\, P \\
        \midrule
        12 & 40 & Raw  & 0.91 & 0.62 & 0.98 & -- \\
           &    & Live & 0.84 & 0.81 & 0.93 & 0.99 \,|\, 0.95 \\
           & 80 & Raw  & 0.92 & 0.81 & 0.99 & -- \\
           &    & Live & 0.86 & 0.96 & 0.94 & 0.99 \,|\, 0.95 \\
        \midrule
        20 & 40 & Raw  & 0.86 & 0.77 & 0.97 & -- \\
           &    & Live & 0.79 & 0.97 & 0.89 & 0.99 \,|\, 0.92 \\
           & 80 & Raw  & 0.88 & 0.89 & 0.98 & -- \\
           &    & Live & 0.82 & 0.99 & 0.90 & 0.99 \,|\, 0.92 \\
        \bottomrule
    \end{tabular}
\end{table}

\section{Discussion and Limitations}
\label{sec:discussion}
Our results illustrate the potential gains from shaping activations to enhance semantic selectivity under tight feature budgets. We have chosen to illustrate our ideas through transformations for which we can offer intuitive interpretations. Projecting onto the live J-space selects directions with larger downstream gain, while the matched filter SAE is inspired by concepts of whitening and matched filtering from communication theory. Apart from performing well on benchmarks, the MF-SAE formulation results in a large fraction of useful atoms without requiring additional mechanisms to address dead atoms. We hope our work stimulates further research into the concept of shaping via linear or nonlinear transformations.

\paragraph{Limitations.}
\emph{Shaping mechanisms.} Our current work shows that, on SAEBench, semantic shaping performs better than dense activations, and MF-SAE performs better than state-of-the-art SAEs. However, combining semantic shaping with the MF-SAE yields gains only at tight feature budgets. Enhancing performance by combining shaping mechanisms remains an important topic for future work.

\emph{Jacobian analysis.} While we have empirically shown the benefits of projecting onto the live J-space, a more detailed analysis of the live and quiet components is an important topic for future work, complementary to the directions pursued in \cite{gurnee2026workspace}. The quiet space is of particular interest: the Jacobian is a linearization averaged over contexts and target positions, so a quiet direction with small average output-layer gain may matter in specific contexts or at intermediate layers.

\emph{Scope.} Our main results are on Gemma-2-2B at two layers, with GPT-2 Small in the appendix; we have not tested larger models.

\emph{Baselines.} Gemma Scope and the Matryoshka SAE were trained on different data and token budgets, and their sparsity is only approximately matched to ours. SAE Probes results are available for these baselines at a single sparsity level.

\emph{Automated interpretability.} Explanations and detection scores come from a single 7B model, and negatives are drawn from other features' activating windows rather than from non-activating text.

\bibliographystyle{plainnat}
\bibliography{iclr2027_conference}

\clearpage
\appendix
\section{Available baseline SAEs}
\label{app:baselines}

Tables~\ref{tab:baselines-gemma} and~\ref{tab:baselines-gpt2} list the open-source SAEs we compare against at each model and layer. JumpReLU SAEs have no fixed sparsity; their L0 is the average number of active latents per token, and we pair each MF-SAE with the JumpReLU SAE of closest average L0. The Matryoshka SAE exists only with a 32K dictionary at L0 $=40$, so it appears only in the 32K comparisons of Appendix~\ref{app:dict-32k} and not in the main text.

\begin{table*}[t]
    \centering
    \caption{\textbf{Baseline SAEs on Gemma-2-2B.} For JumpReLU SAEs, L0 is the average number of active latents per token.}
    \label{tab:baselines-gemma}
    \begin{tabular}{clllc}
        \toprule
        Layer & SAE & Architecture & Width & Sparsity \\
        \midrule
        12 & SAEBench \cite{karvonen2025saebench} & Top-K & 16K & $k=40,80,160,320$ \\
           & Gemma Scope \cite{lieberum2024gemmascope} & JumpReLU & 16K & L0 $=22,41,82,176,445$ \\
           & Gemma Scope \cite{lieberum2024gemmascope} & JumpReLU & 32K & L0 $=12,22,40,76,155,360$ \\
           & Matryoshka \cite{bussmann2025matryoshka} & BatchTopK & 32K & L0 $=40$ \\
        \midrule
        20 & Gemma Scope \cite{lieberum2024gemmascope} & JumpReLU & 16K & L0 $=22,38,71,139,294$ \\
           & Matryoshka \cite{bussmann2025matryoshka} & BatchTopK & 32K & L0 $=40$ \\
        \bottomrule
    \end{tabular}
\end{table*}

\begin{table*}[t]
    \centering
    \caption{\textbf{Baseline SAEs on GPT-2 Small.} Our MF-SAEs are trained at the same hook point, with the same dictionary size and matched sparsity.}
    \label{tab:baselines-gpt2}
    \begin{tabular}{clllc}
        \toprule
        Layer & SAE & Hook & Width & Sparsity \\
        \midrule
        7 & ReLU, L1 \cite{bloom2024gpt2small} & \texttt{resid\_pre} & 24{,}576 & L0 $\approx 54$ \\
        8 & Top-K \cite{gao2025scaling} & \texttt{resid\_post} & 32{,}768 & $k=32$ \\
        \bottomrule
    \end{tabular}
\end{table*}

\section{Sparsity level ablation, 16K dictionary}
\label{app:sparsity-16k}

\begin{figure*}[t]
    \centering
    \begin{subfigure}[b]{0.48\textwidth}
        \centering
        \includegraphics[width=\linewidth, height=0.5\linewidth]{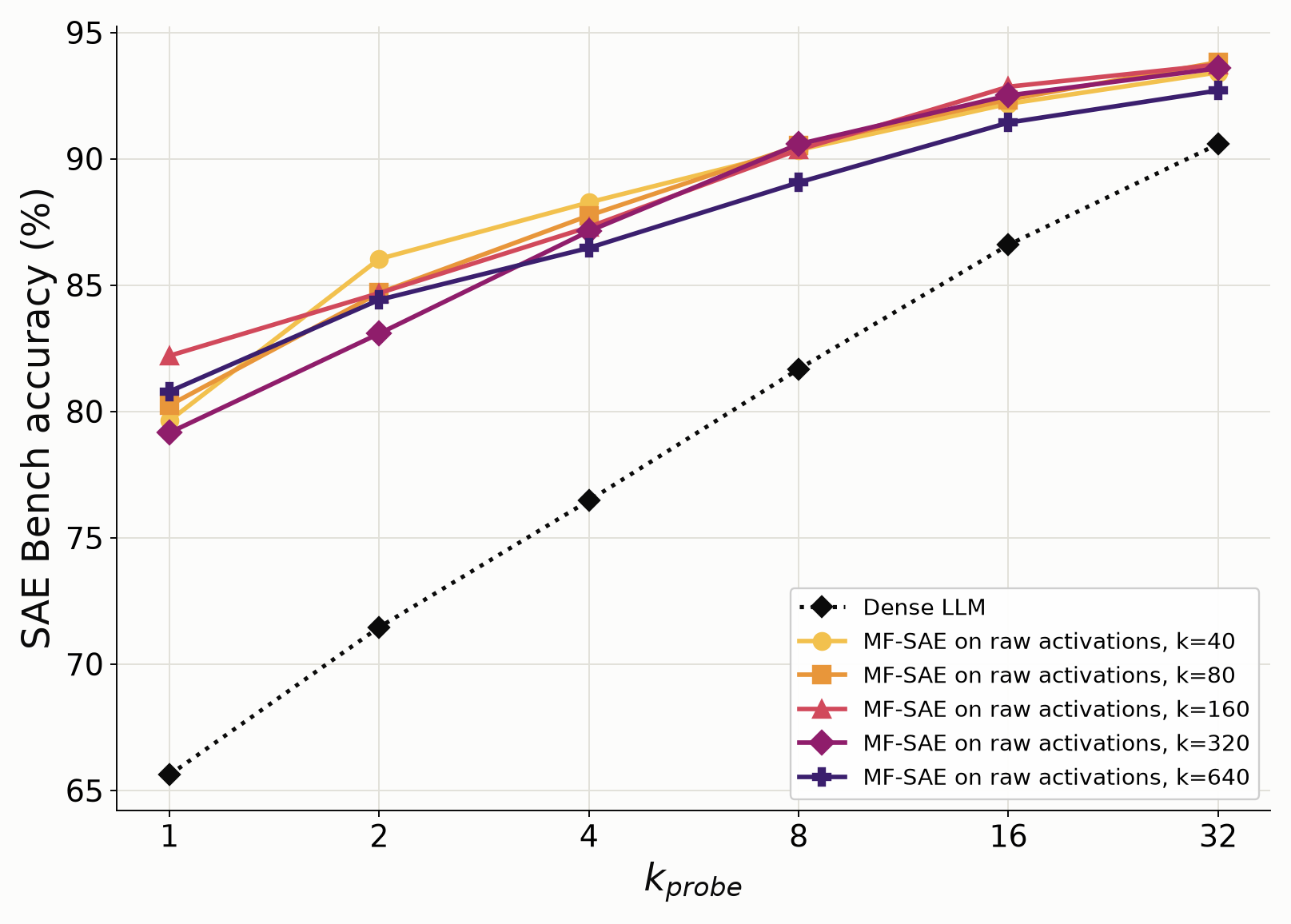}
        \caption{Raw, layer 12}
        \label{fig:a16-mf-sb-a}
    \end{subfigure}
    \hfill
    \begin{subfigure}[b]{0.48\textwidth}
        \centering
        \includegraphics[width=\linewidth, height=0.5\linewidth]{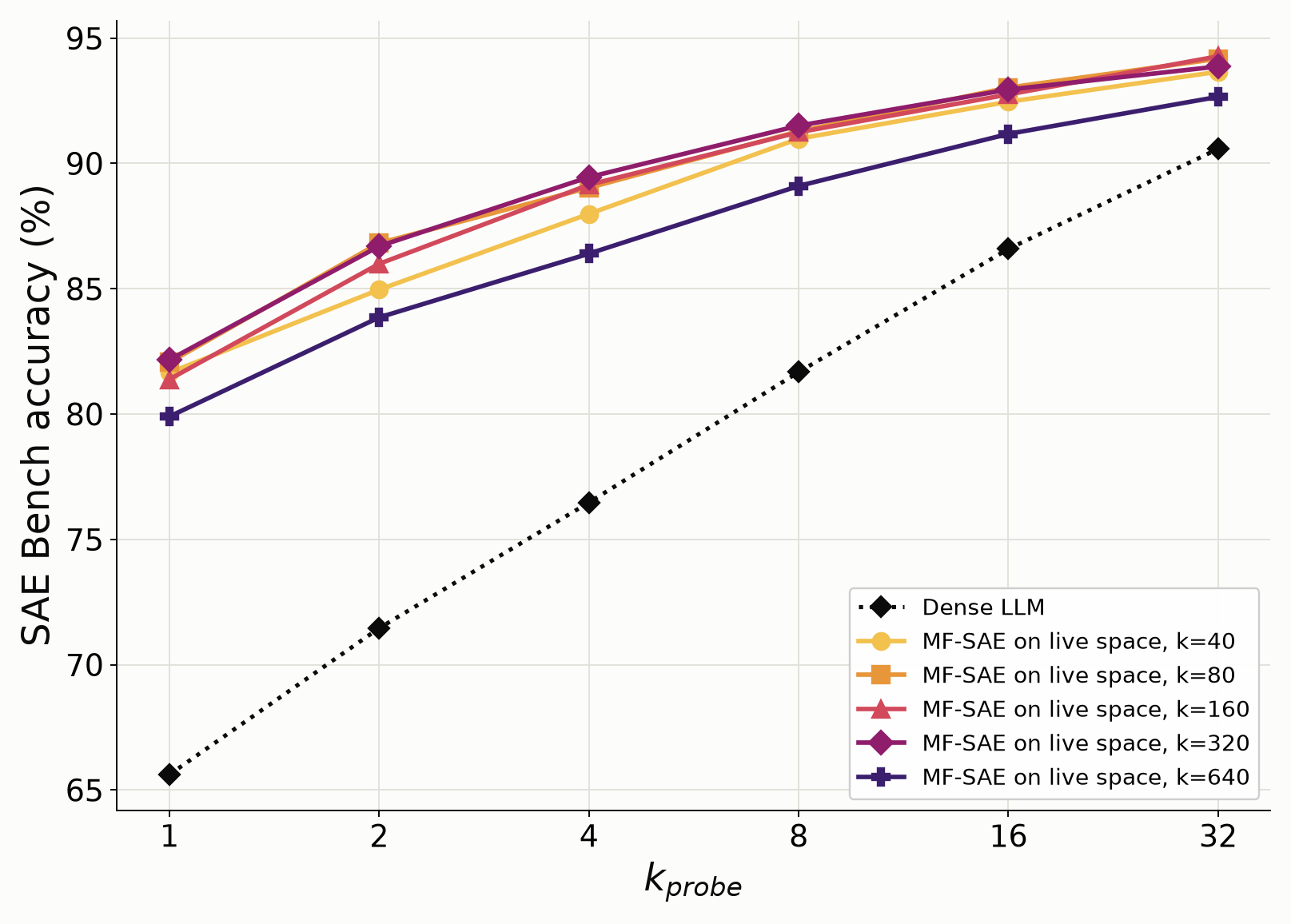}
        \caption{Live, layer 12}
        \label{fig:a16-mf-sb-b}
    \end{subfigure}

    \vspace{6pt}

    \begin{subfigure}[b]{0.48\textwidth}
        \centering
        \includegraphics[width=\linewidth, height=0.5\linewidth]{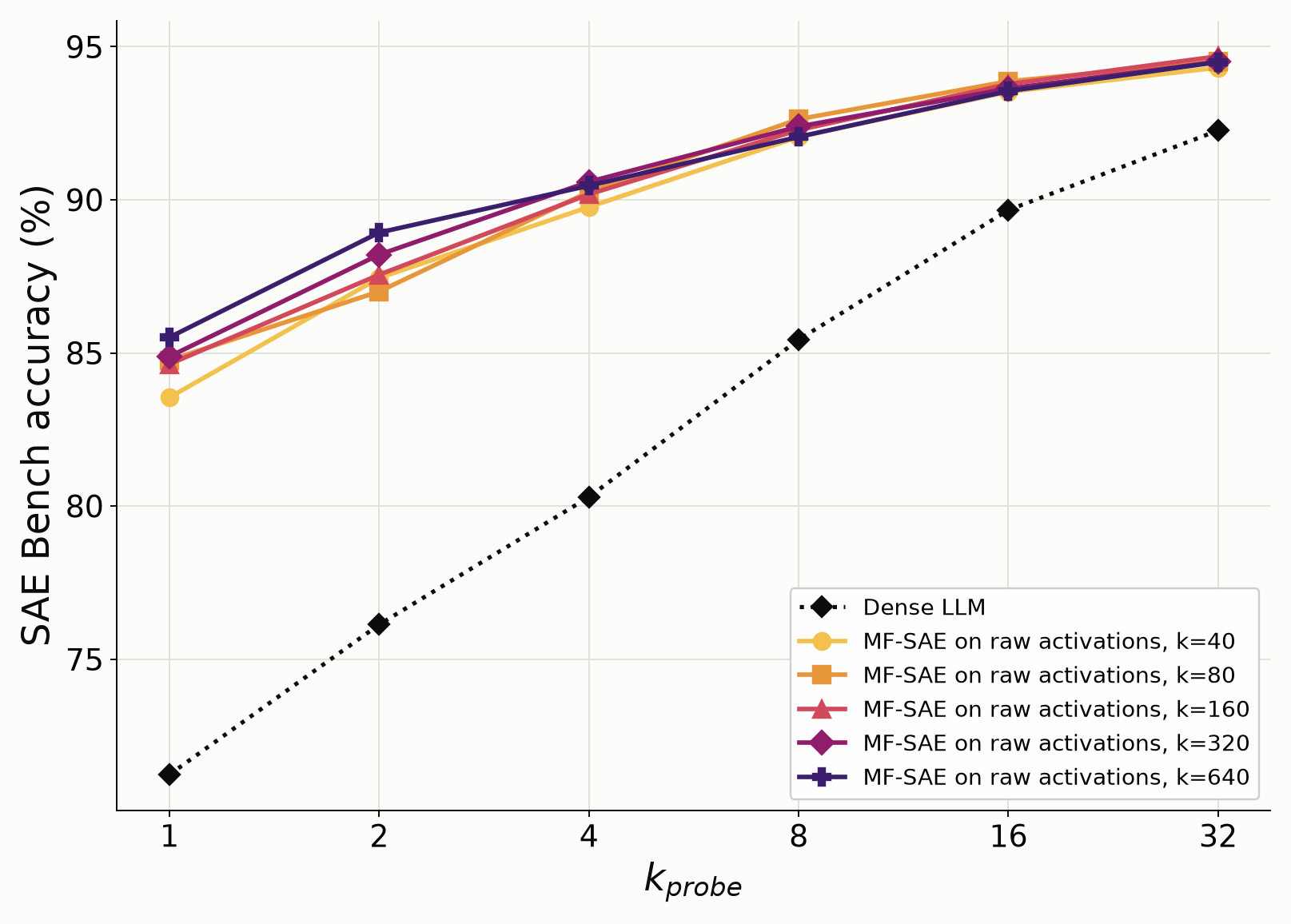}
        \caption{Raw, layer 20}
        \label{fig:a16-mf-sb-c}
    \end{subfigure}
    \hfill
    \begin{subfigure}[b]{0.48\textwidth}
        \centering
        \includegraphics[width=\linewidth, height=0.5\linewidth]{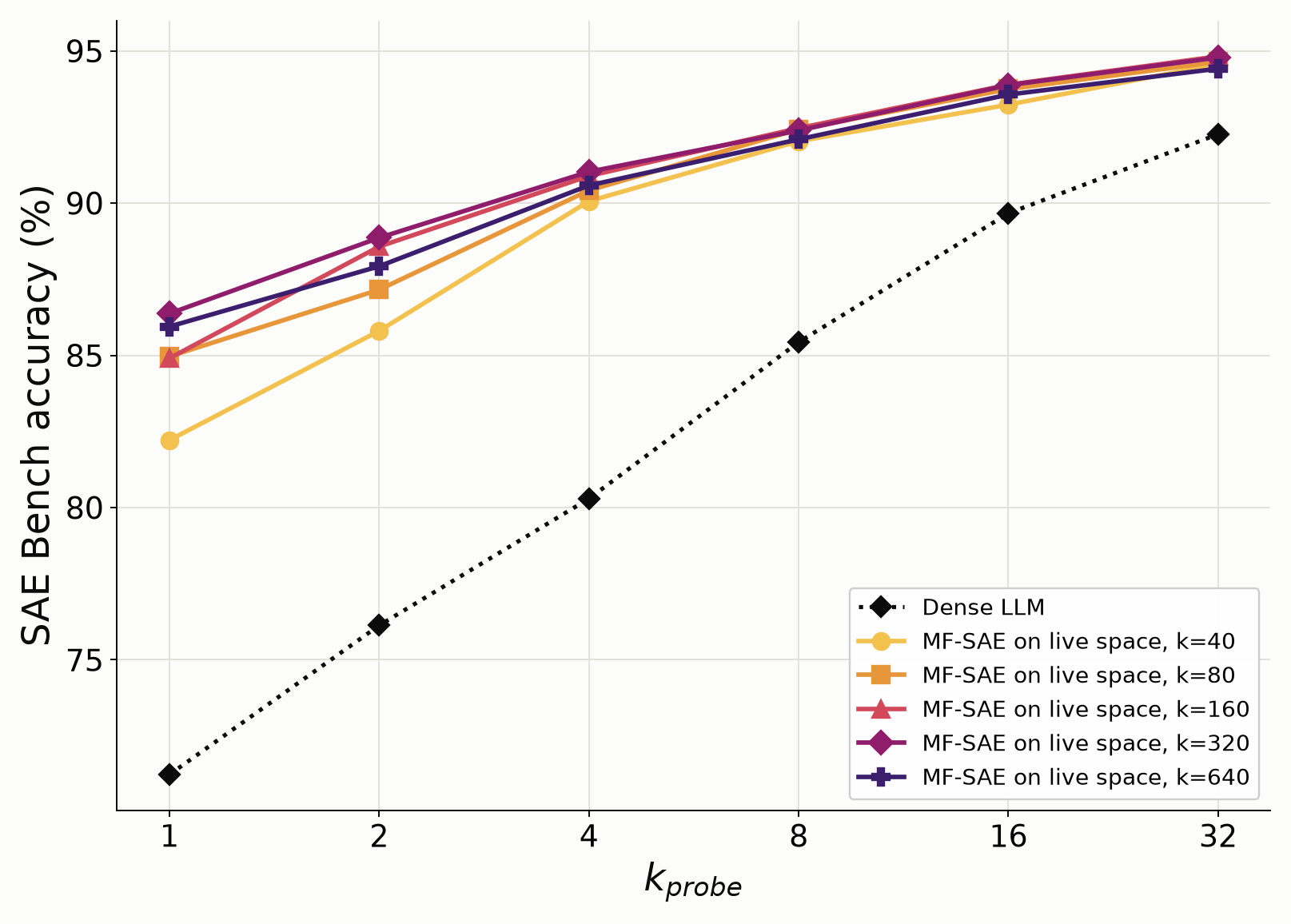}
        \caption{Live, layer 20}
        \label{fig:a16-mf-sb-d}
    \end{subfigure}
    \caption{\textbf{SAEBench, MF-SAE across sparsity levels (Gemma-2-2B, 16K dictionary).} SB-Acc against $k_{\text{probe}}$ for the MF-SAE at $k\in\{40,80,160,320\}$. Columns are the raw and live MF-SAE; rows are layers 12 and 20. $k=320$ is among the best sparsity levels but not the best in every panel.}
    \label{fig:a16-mf-sb}
\end{figure*}

\begin{figure*}[t]
    \centering
    \begin{subfigure}[b]{0.48\textwidth}
        \centering
        \includegraphics[width=\linewidth, height=0.5\linewidth]{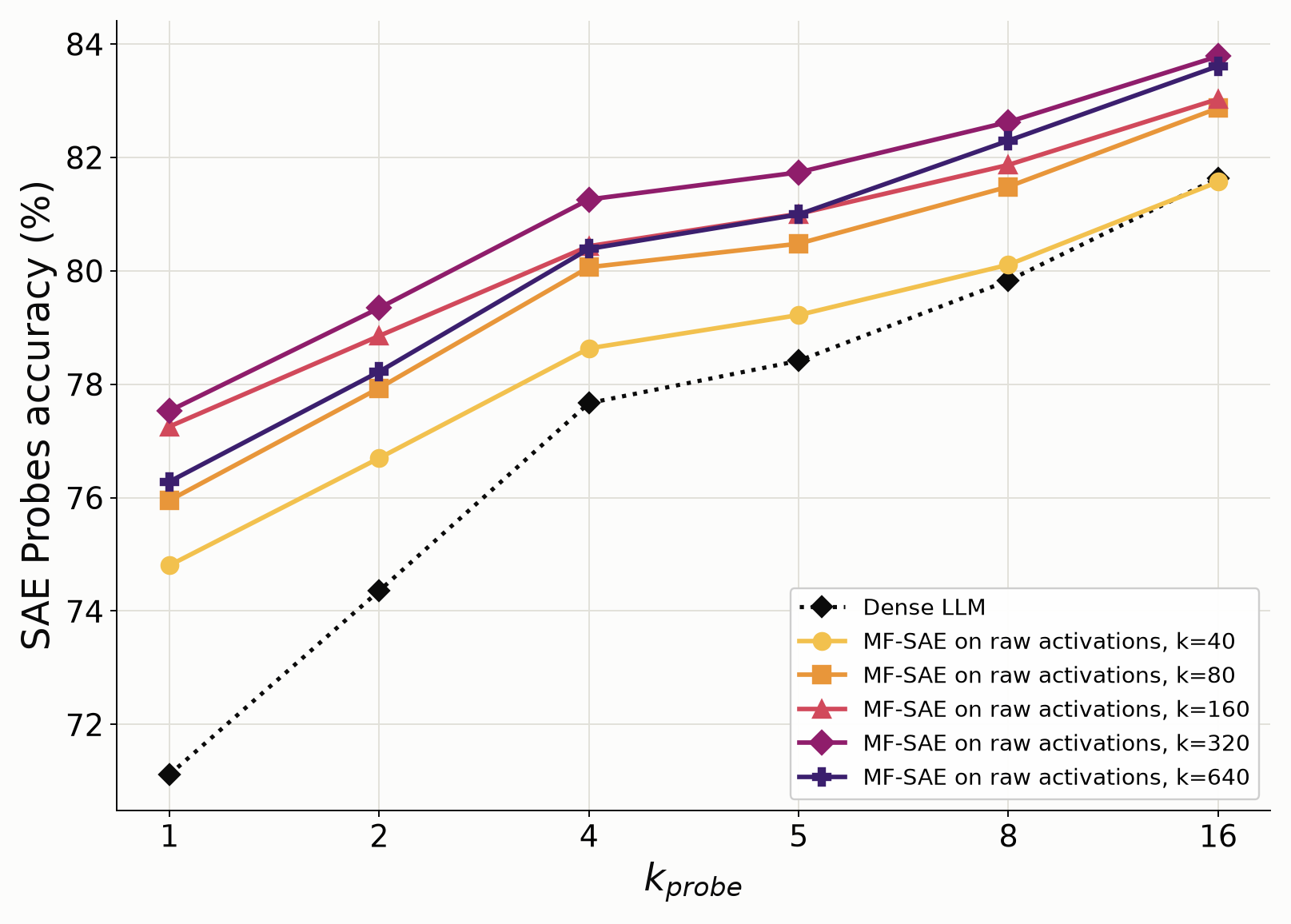}
        \caption{Raw, layer 12}
        \label{fig:a16-mf-sp-a}
    \end{subfigure}
    \hfill
    \begin{subfigure}[b]{0.48\textwidth}
        \centering
        \includegraphics[width=\linewidth, height=0.5\linewidth]{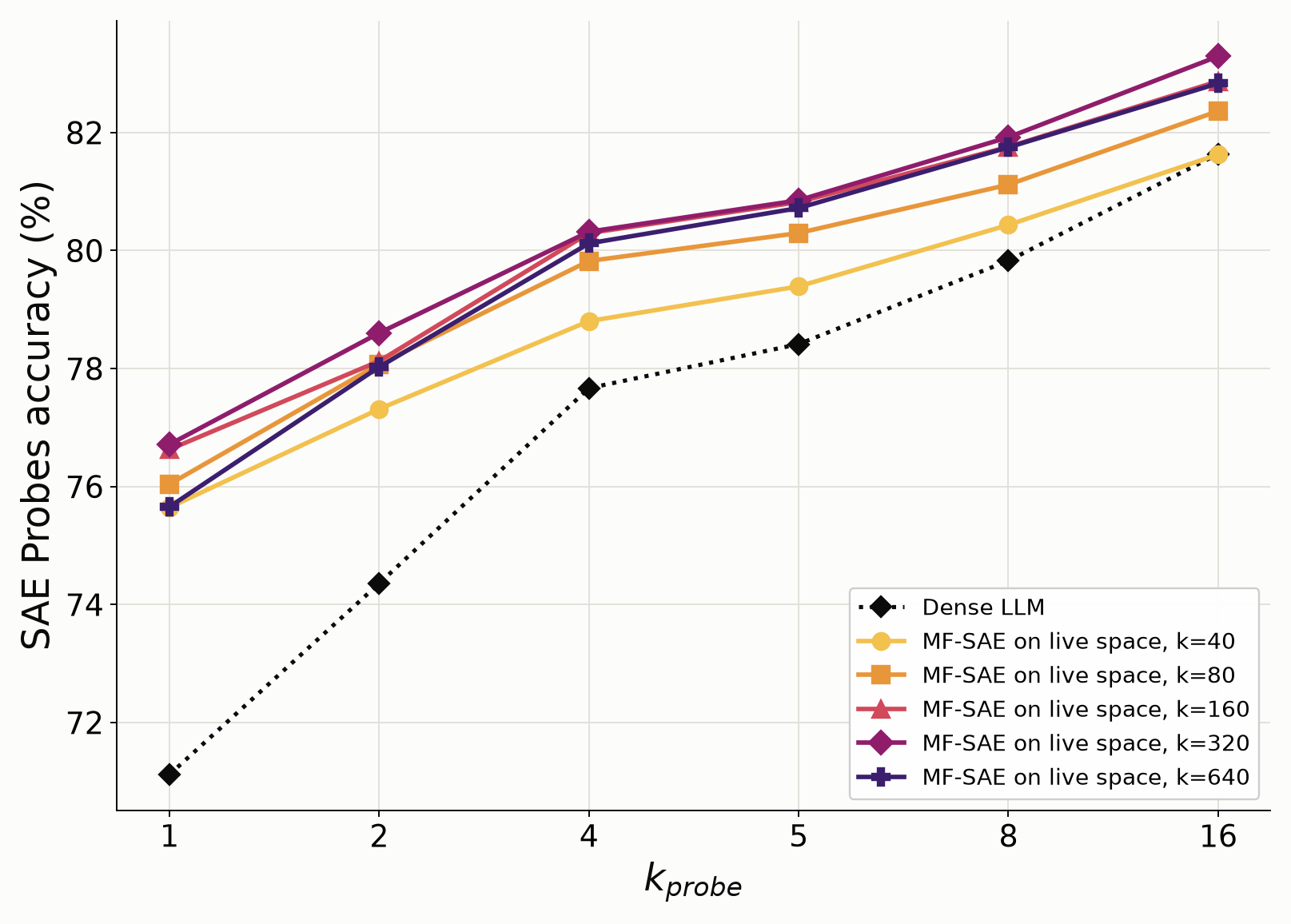}
        \caption{Live, layer 12}
        \label{fig:a16-mf-sp-b}
    \end{subfigure}

    \vspace{6pt}

    \begin{subfigure}[b]{0.48\textwidth}
        \centering
        \includegraphics[width=\linewidth, height=0.5\linewidth]{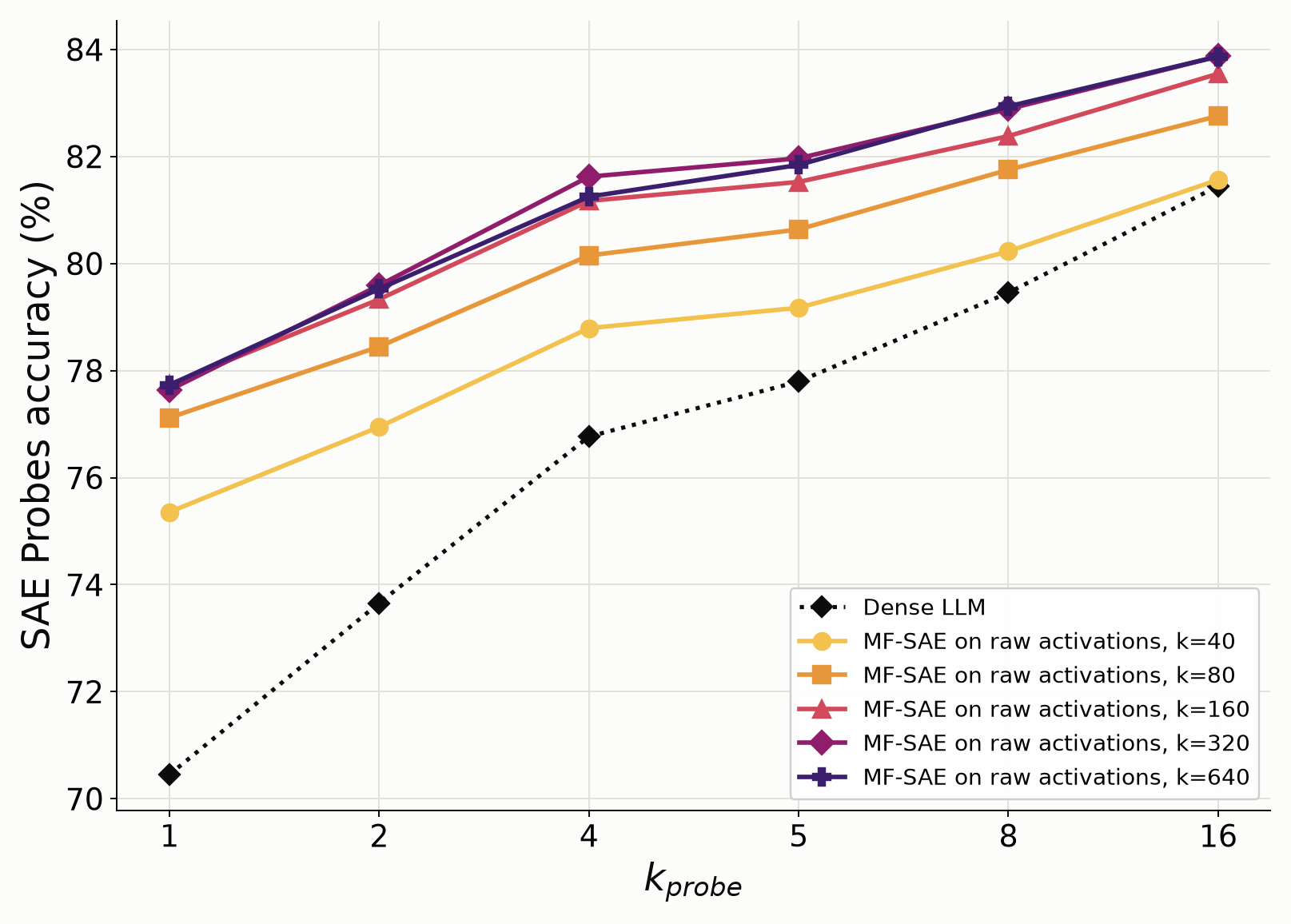}
        \caption{Raw, layer 20}
        \label{fig:a16-mf-sp-c}
    \end{subfigure}
    \hfill
    \begin{subfigure}[b]{0.48\textwidth}
        \centering
        \includegraphics[width=\linewidth, height=0.5\linewidth]{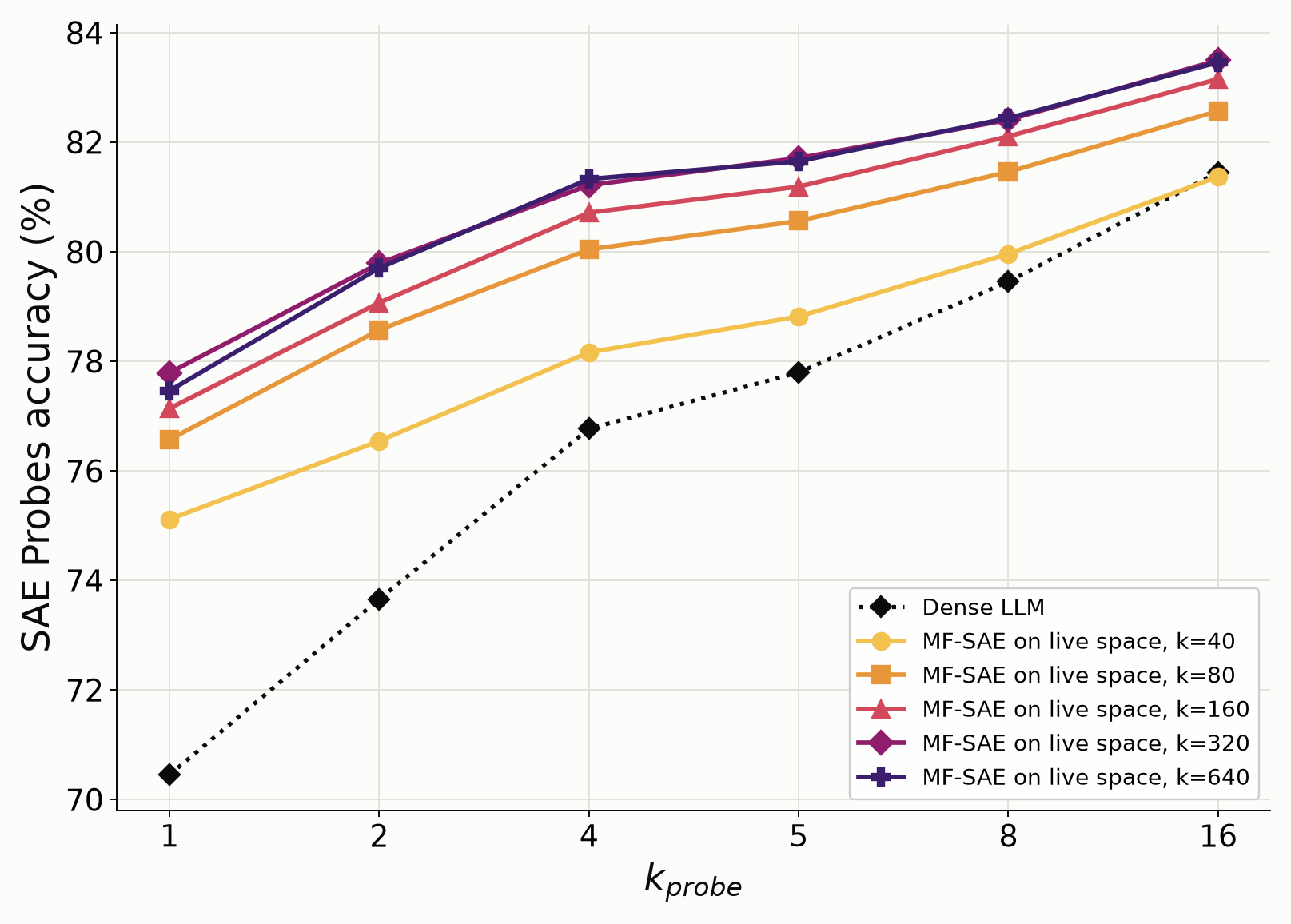}
        \caption{Live, layer 20}
        \label{fig:a16-mf-sp-d}
    \end{subfigure}
    \caption{\textbf{SAE Probes, MF-SAE across sparsity levels (Gemma-2-2B, 16K dictionary).} Mean test accuracy against $k_{\text{probe}}$ for the MF-SAE at $k\in\{40,80,160,320\}$, with the same layout as Figure~\ref{fig:a16-mf-sb}. $k=320$ is consistently the best sparsity level.}
    \label{fig:a16-mf-sp}
\end{figure*}

\begin{figure*}[t]
    \centering
    \begin{subfigure}[b]{0.48\textwidth}
        \centering
        \includegraphics[width=\linewidth, height=0.5\linewidth]{Figures/SAE_Bench/Gemma_Layer12/16K/gemma2_l12_16k_k40_saebench.png}
        \caption{$k=40$}
        \label{fig:a16-l12-sb-a}
    \end{subfigure}
    \hfill
    \begin{subfigure}[b]{0.48\textwidth}
        \centering
        \includegraphics[width=\linewidth, height=0.5\linewidth]{Figures/SAE_Bench/Gemma_Layer12/16K/gemma2_l12_16k_k80_saebench.png}
        \caption{$k=80$}
        \label{fig:a16-l12-sb-b}
    \end{subfigure}

    \vspace{6pt}

    \begin{subfigure}[b]{0.48\textwidth}
        \centering
        \includegraphics[width=\linewidth, height=0.5\linewidth]{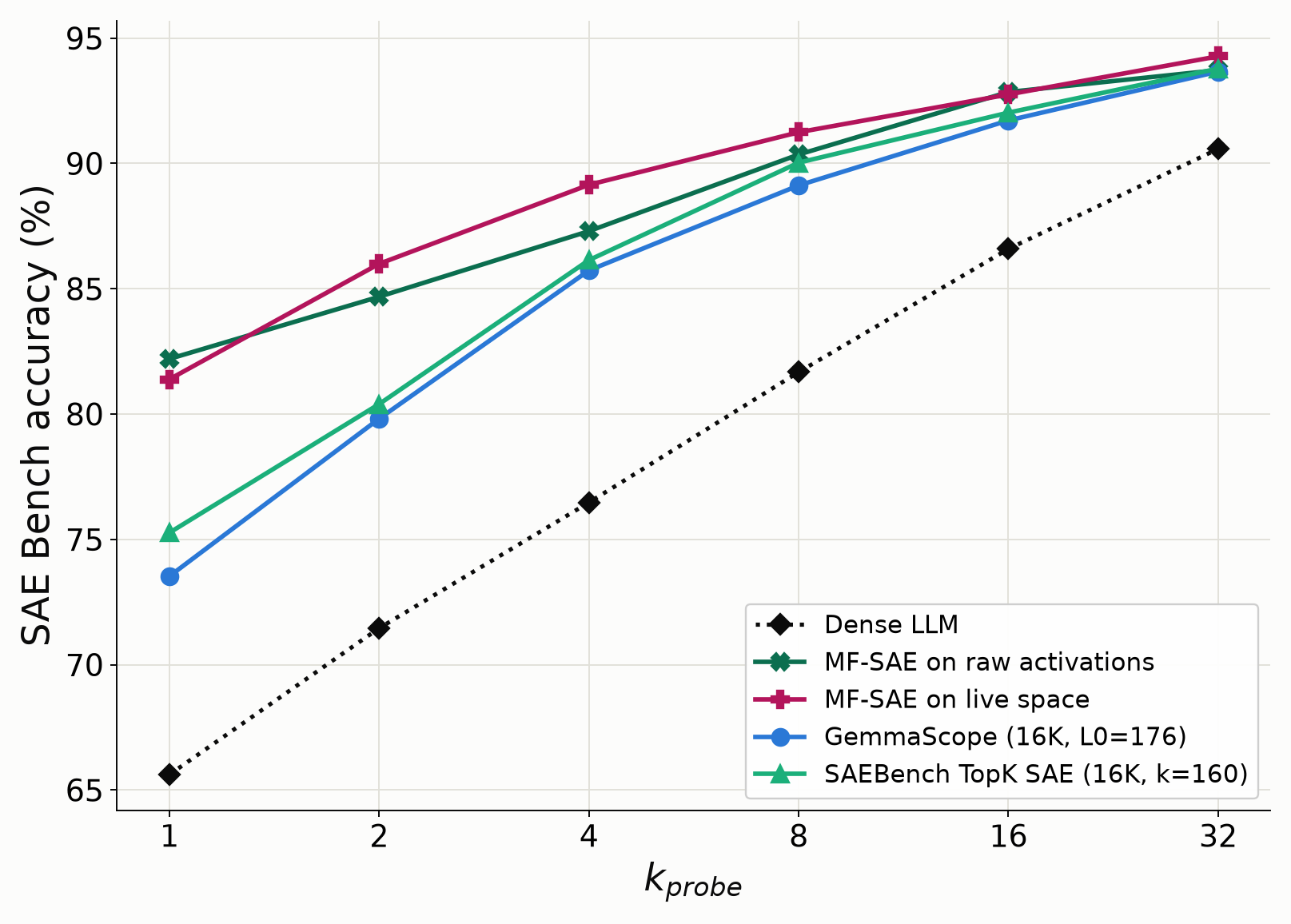}
        \caption{$k=160$}
        \label{fig:a16-l12-sb-c}
    \end{subfigure}
    \hfill
    \begin{subfigure}[b]{0.48\textwidth}
        \centering
        \includegraphics[width=\linewidth, height=0.5\linewidth]{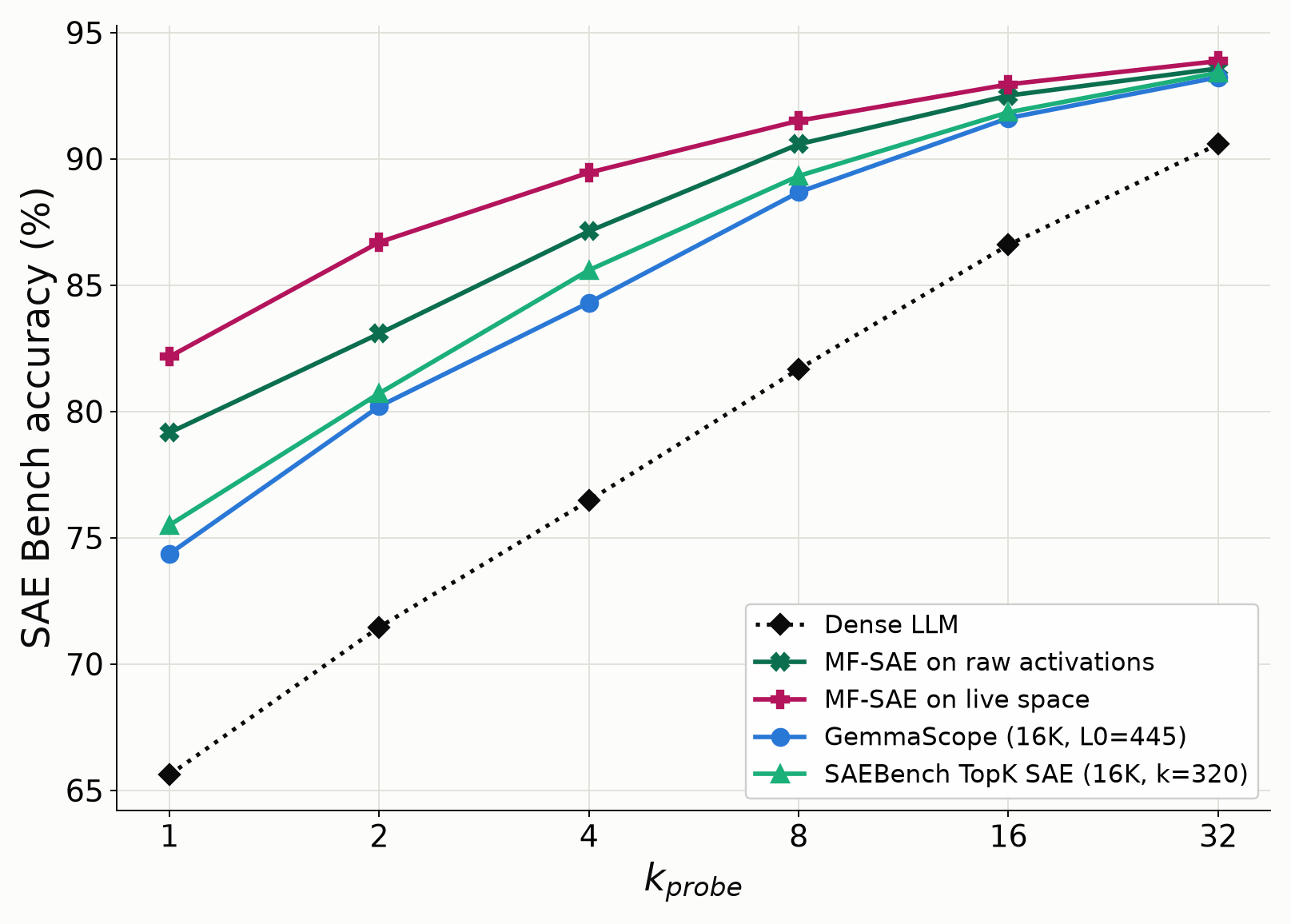}
        \caption{$k=320$}
        \label{fig:a16-l12-sb-d}
    \end{subfigure}
    \caption{\textbf{SAEBench, layer 12, 16K dictionary.} SB-Acc against $k_{\text{probe}}$ for the raw and live MF-SAEs, the SAEBench Top-K SAE at the same $k$, and the Gemma Scope 16K SAE of closest average L0.}
    \label{fig:a16-l12-sb}
\end{figure*}

\begin{figure*}[t]
    \centering
    \begin{subfigure}[b]{0.48\textwidth}
        \centering
        \includegraphics[width=\linewidth, height=0.5\linewidth]{Figures/SAE_Probes/Gemma_Layer12/16K/gemma2_l12_16k_k40_saeprobes.png}
        \caption{$k=40$}
        \label{fig:a16-l12-sp-a}
    \end{subfigure}
    \hfill
    \begin{subfigure}[b]{0.48\textwidth}
        \centering
        \includegraphics[width=\linewidth, height=0.5\linewidth]{Figures/SAE_Probes/Gemma_Layer12/16K/gemma2_l12_16k_k80_saeprobes.png}
        \caption{$k=80$}
        \label{fig:a16-l12-sp-b}
    \end{subfigure}

    \vspace{6pt}

    \begin{subfigure}[b]{0.48\textwidth}
        \centering
        \includegraphics[width=\linewidth, height=0.5\linewidth]{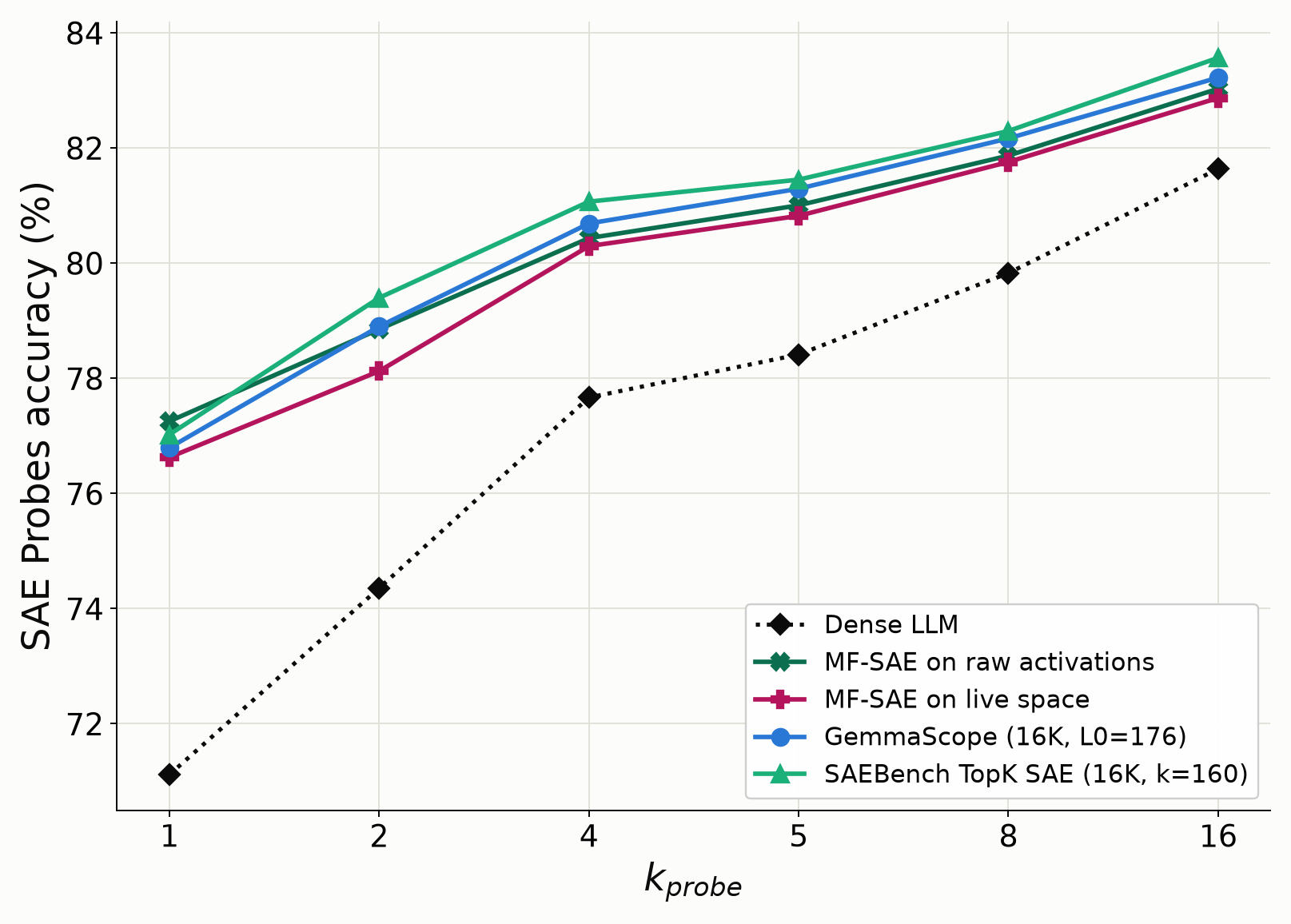}
        \caption{$k=160$}
        \label{fig:a16-l12-sp-c}
    \end{subfigure}
    \hfill
    \begin{subfigure}[b]{0.48\textwidth}
        \centering
        \includegraphics[width=\linewidth, height=0.5\linewidth]{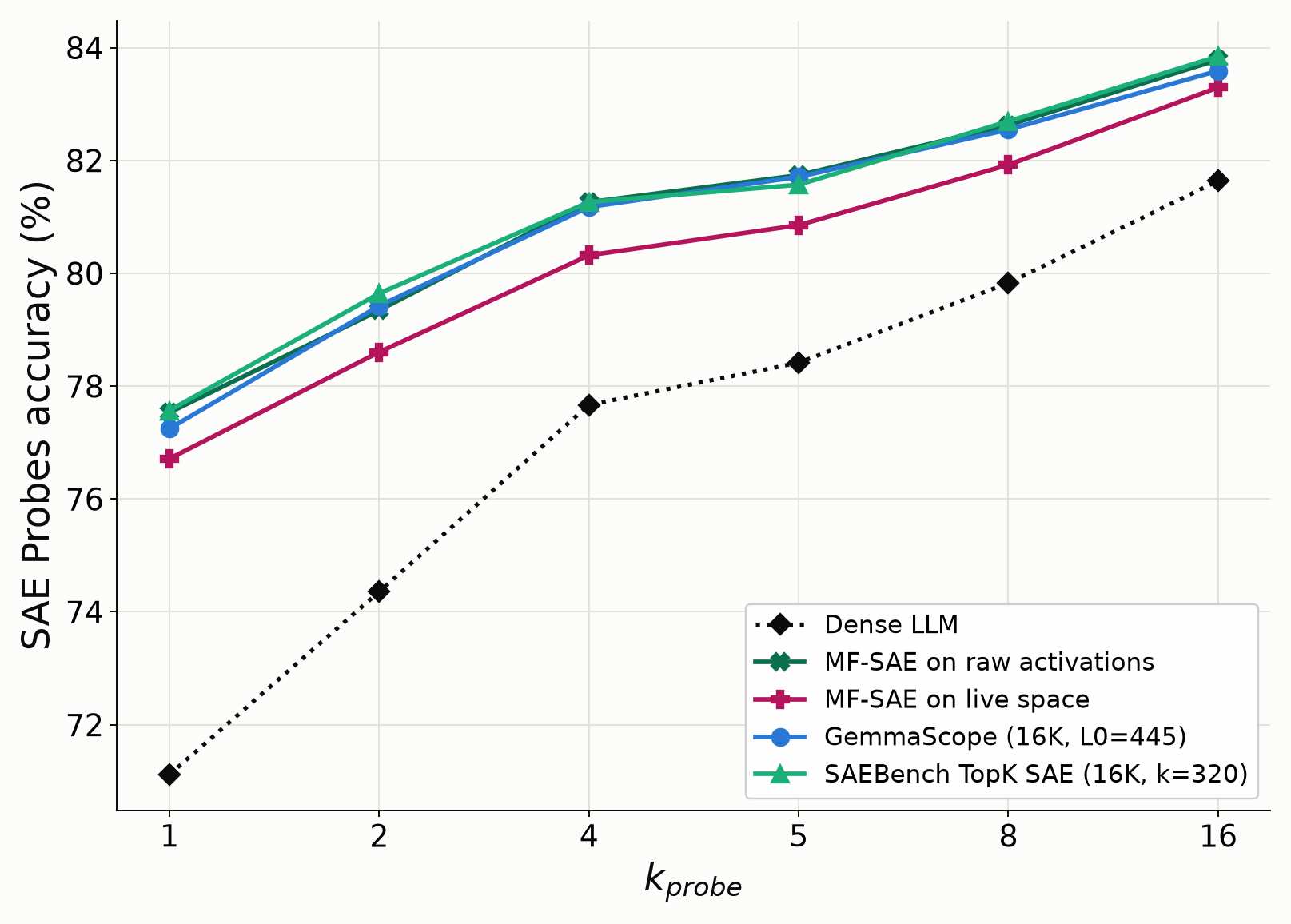}
        \caption{$k=320$}
        \label{fig:a16-l12-sp-d}
    \end{subfigure}
    \caption{\textbf{SAE Probes, layer 12, 16K dictionary.} Mean test accuracy against $k_{\text{probe}}$, with the same methods and layout as Figure~\ref{fig:a16-l12-sb}.}
    \label{fig:a16-l12-sp}
\end{figure*}

\begin{figure*}[t]
    \centering
    \begin{subfigure}[b]{0.48\textwidth}
        \centering
        \includegraphics[width=\linewidth, height=0.5\linewidth]{Figures/SAE_Bench/Gemma_Layer20/16K/gemma2_l20_16k_k40_saebench.png}
        \caption{$k=40$}
        \label{fig:a16-l20-sb-a}
    \end{subfigure}
    \hfill
    \begin{subfigure}[b]{0.48\textwidth}
        \centering
        \includegraphics[width=\linewidth, height=0.5\linewidth]{Figures/SAE_Bench/Gemma_Layer20/16K/gemma2_l20_16k_k80_saebench.png}
        \caption{$k=80$}
        \label{fig:a16-l20-sb-b}
    \end{subfigure}

    \vspace{6pt}

    \begin{subfigure}[b]{0.48\textwidth}
        \centering
        \includegraphics[width=\linewidth, height=0.5\linewidth]{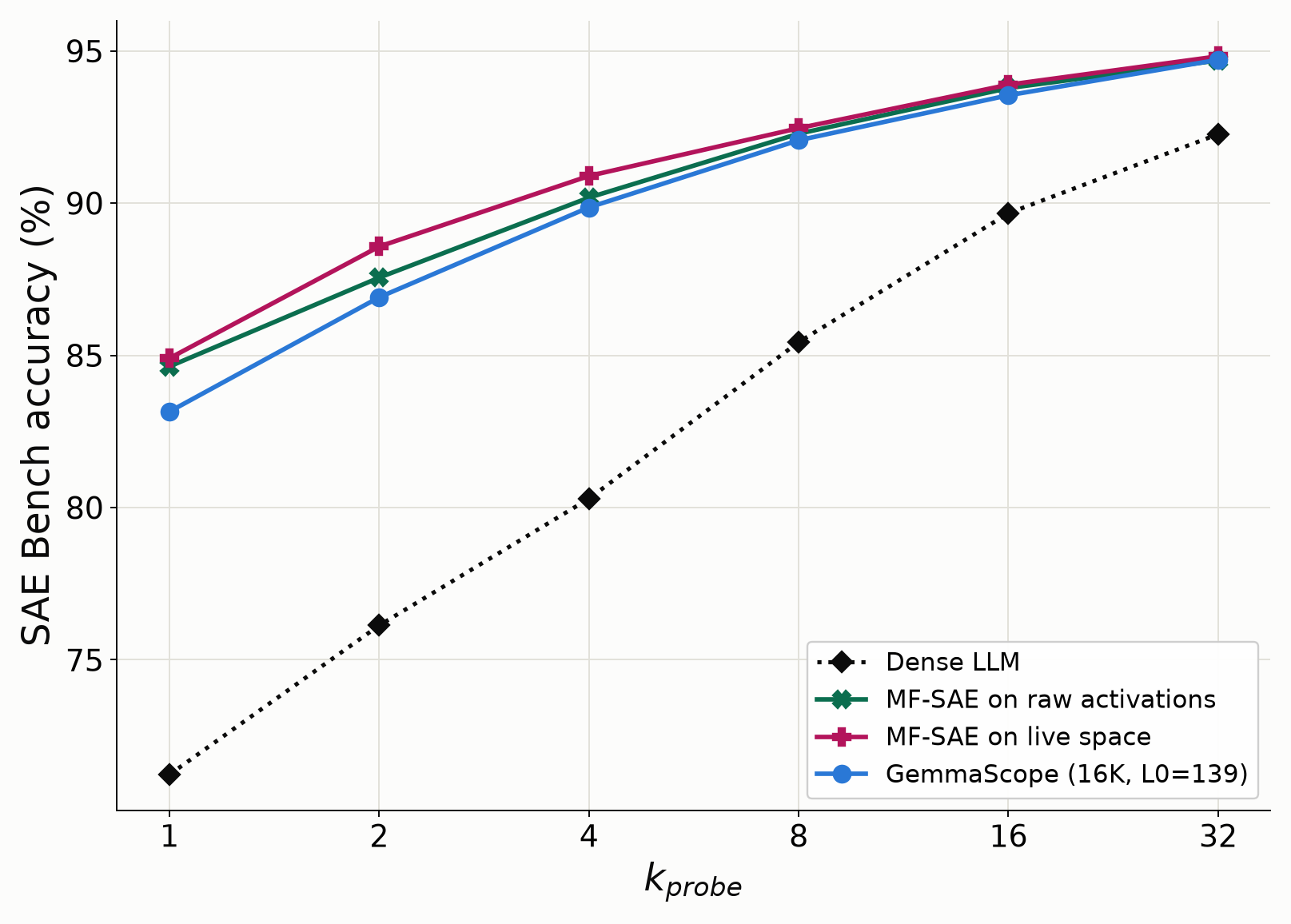}
        \caption{$k=160$}
        \label{fig:a16-l20-sb-c}
    \end{subfigure}
    \hfill
    \begin{subfigure}[b]{0.48\textwidth}
        \centering
        \includegraphics[width=\linewidth, height=0.5\linewidth]{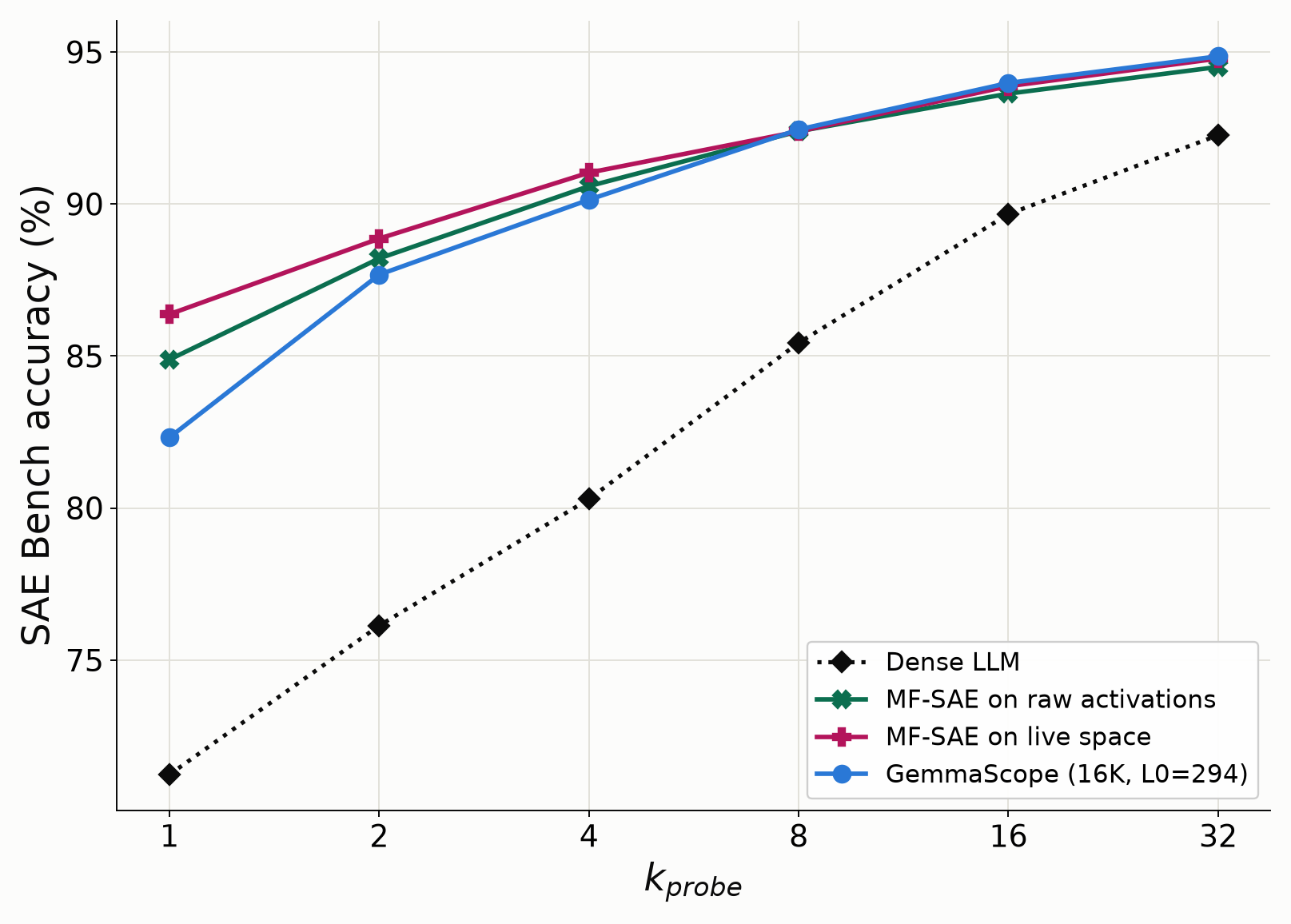}
        \caption{$k=320$}
        \label{fig:a16-l20-sb-d}
    \end{subfigure}
    \caption{\textbf{SAEBench, layer 20, 16K dictionary.} SB-Acc against $k_{\text{probe}}$ for the raw and live MF-SAEs and the Gemma Scope 16K SAE of closest average L0 (38, 71, 139 and 294 for $k=40,80,160,320$). No SAEBench SAEs are available at this layer.}
    \label{fig:a16-l20-sb}
\end{figure*}

\begin{figure*}[t]
    \centering
    \begin{subfigure}[b]{0.48\textwidth}
        \centering
        \includegraphics[width=\linewidth, height=0.5\linewidth]{Figures/SAE_Probes/Gemma_Layer20/16K/gemma2_l20_16k_k40_saeprobes.png}
        \caption{$k=40$}
        \label{fig:a16-l20-sp-a}
    \end{subfigure}
    \hfill
    \begin{subfigure}[b]{0.48\textwidth}
        \centering
        \includegraphics[width=\linewidth, height=0.5\linewidth]{Figures/SAE_Probes/Gemma_Layer20/16K/gemma2_l20_16k_k80_saeprobes.png}
        \caption{$k=80$}
        \label{fig:a16-l20-sp-b}
    \end{subfigure}

    \vspace{6pt}

    \begin{subfigure}[b]{0.48\textwidth}
        \centering
        \includegraphics[width=\linewidth, height=0.5\linewidth]{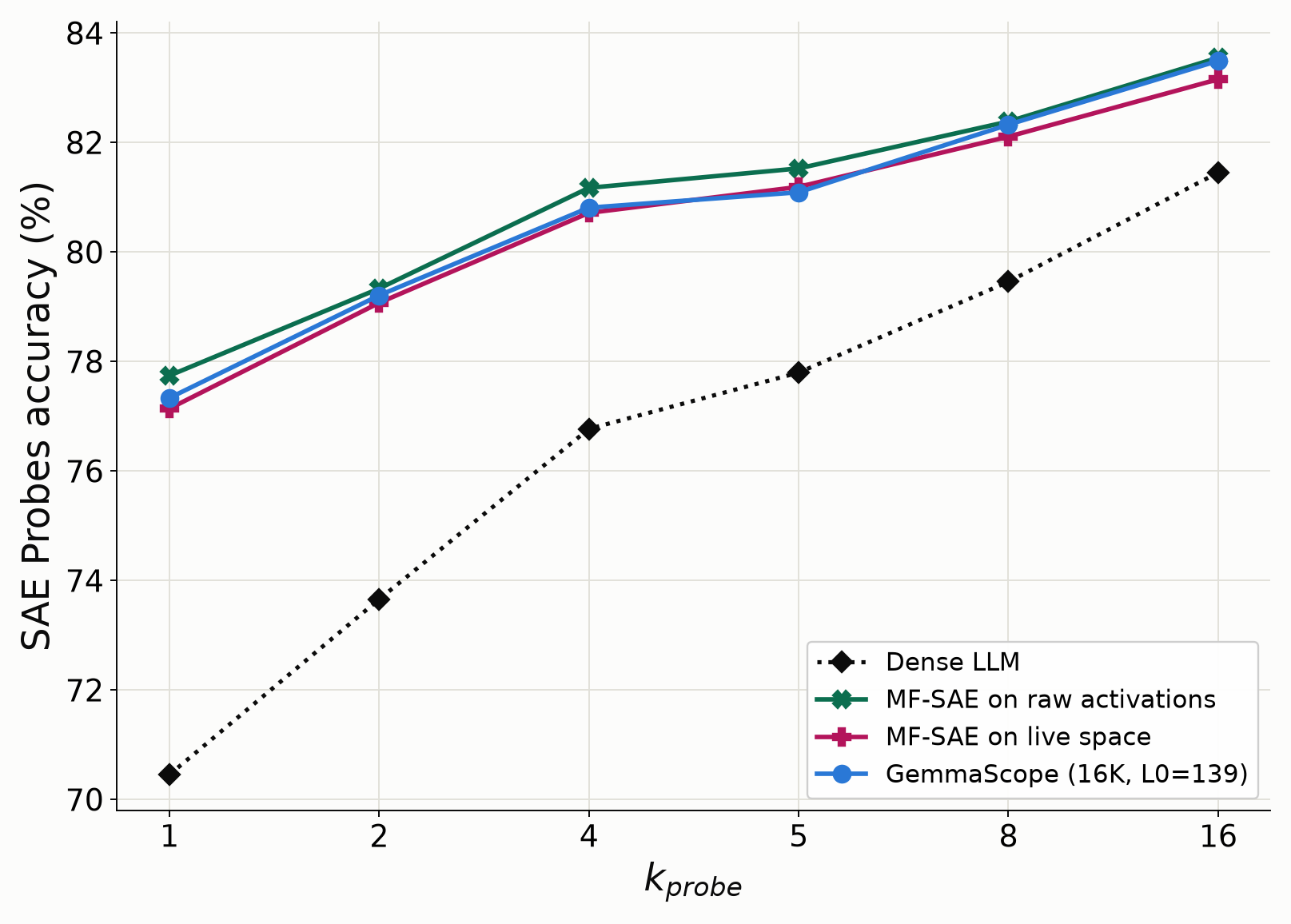}
        \caption{$k=160$}
        \label{fig:a16-l20-sp-c}
    \end{subfigure}
    \hfill
    \begin{subfigure}[b]{0.48\textwidth}
        \centering
        \includegraphics[width=\linewidth, height=0.5\linewidth]{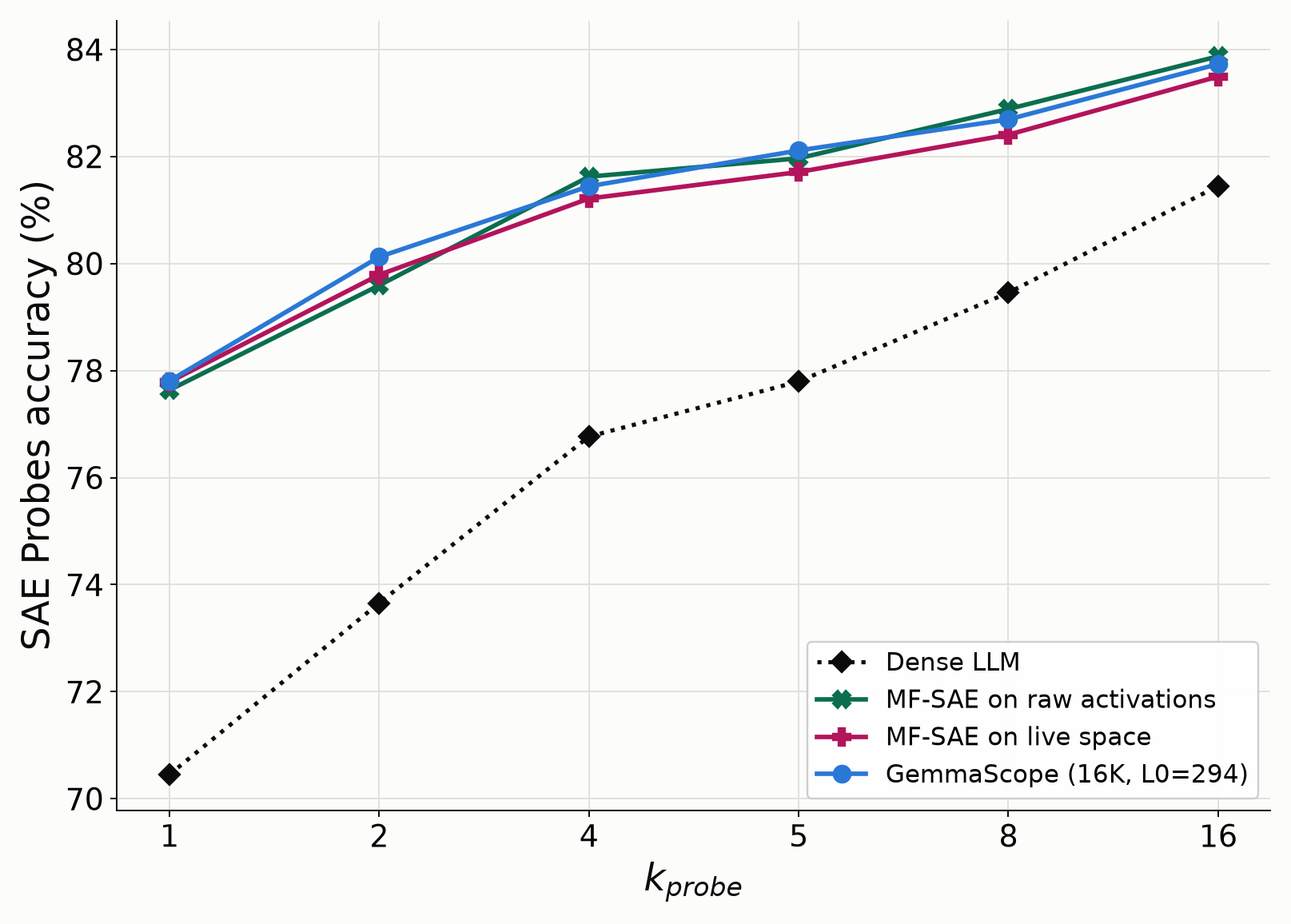}
        \caption{$k=320$}
        \label{fig:a16-l20-sp-d}
    \end{subfigure}
    \caption{\textbf{SAE Probes, layer 20, 16K dictionary.} Mean test accuracy against $k_{\text{probe}}$, with the same methods and layout as Figure~\ref{fig:a16-l20-sb}.}
    \label{fig:a16-l20-sp}
\end{figure*}

This section extends Section~\ref{sec:res-probing} to all sparsity levels $k\in\{40,80,160,320\}$ with a 16K dictionary. Across sparsity levels, $k=320$ gives the best MF-SAE on SAE Probes in every setting, while on SAEBench it is among the best sparsity levels but not always the best (Figures~\ref{fig:a16-mf-sb} and~\ref{fig:a16-mf-sp}). At layer 12 the baselines are the SAEBench Top-K and Gemma Scope 16K SAEs; at layer 20, only Gemma Scope 16K is available (Table~\ref{tab:baselines-gemma}). The comparisons with these baselines are shown in Figures~\ref{fig:a16-l12-sb} and~\ref{fig:a16-l12-sp} for layer 12 and Figures~\ref{fig:a16-l20-sb} and~\ref{fig:a16-l20-sp} for layer 20.

\section{Dictionary size ablation, 32K dictionary}
\label{app:dict-32k}

\begin{figure*}[t]
    \centering
    \begin{subfigure}[b]{0.48\textwidth}
        \centering
        \includegraphics[width=\linewidth, height=0.5\linewidth]{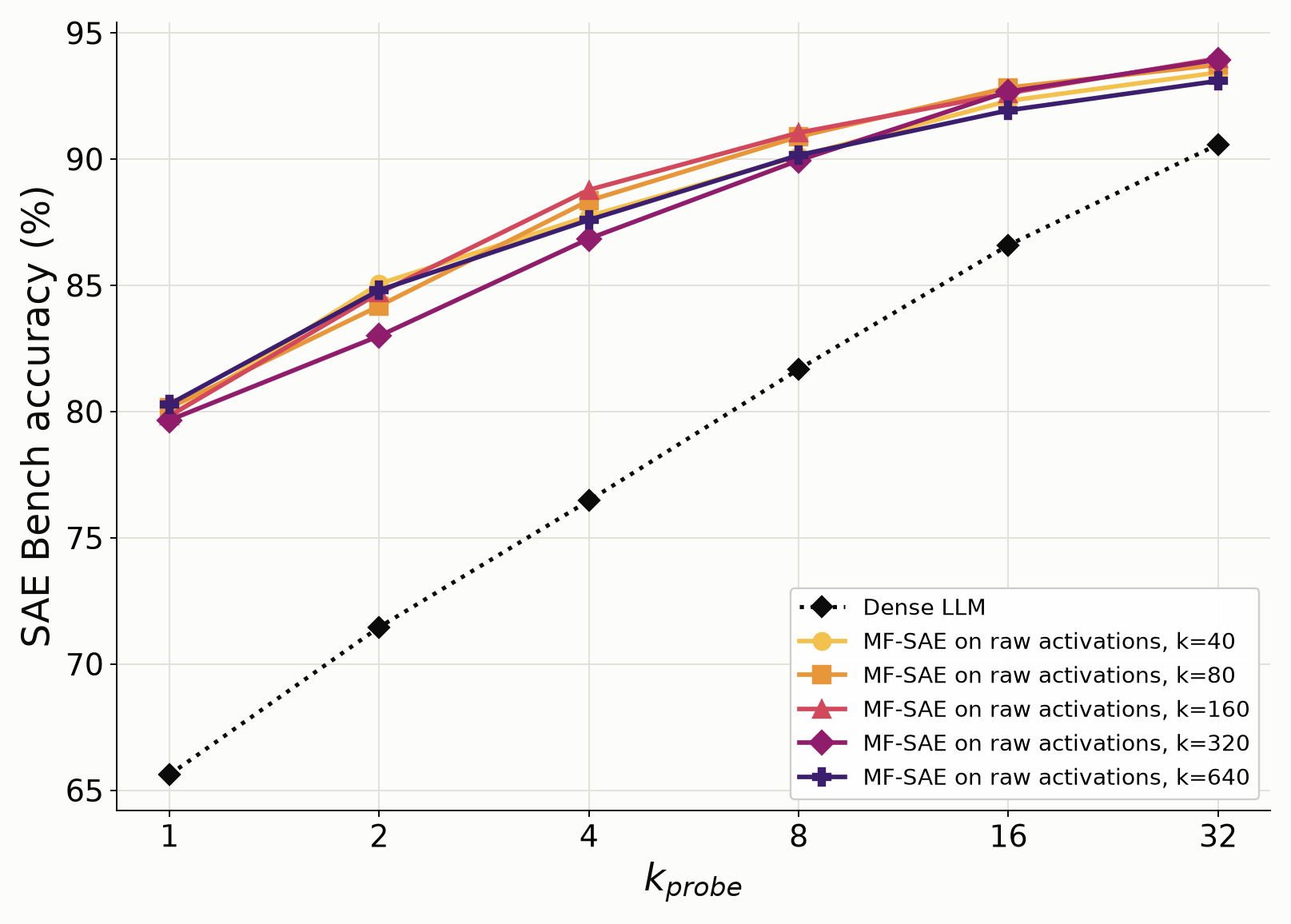}
        \caption{Raw, layer 12}
        \label{fig:a32-mf-sb-a}
    \end{subfigure}
    \hfill
    \begin{subfigure}[b]{0.48\textwidth}
        \centering
        \includegraphics[width=\linewidth, height=0.5\linewidth]{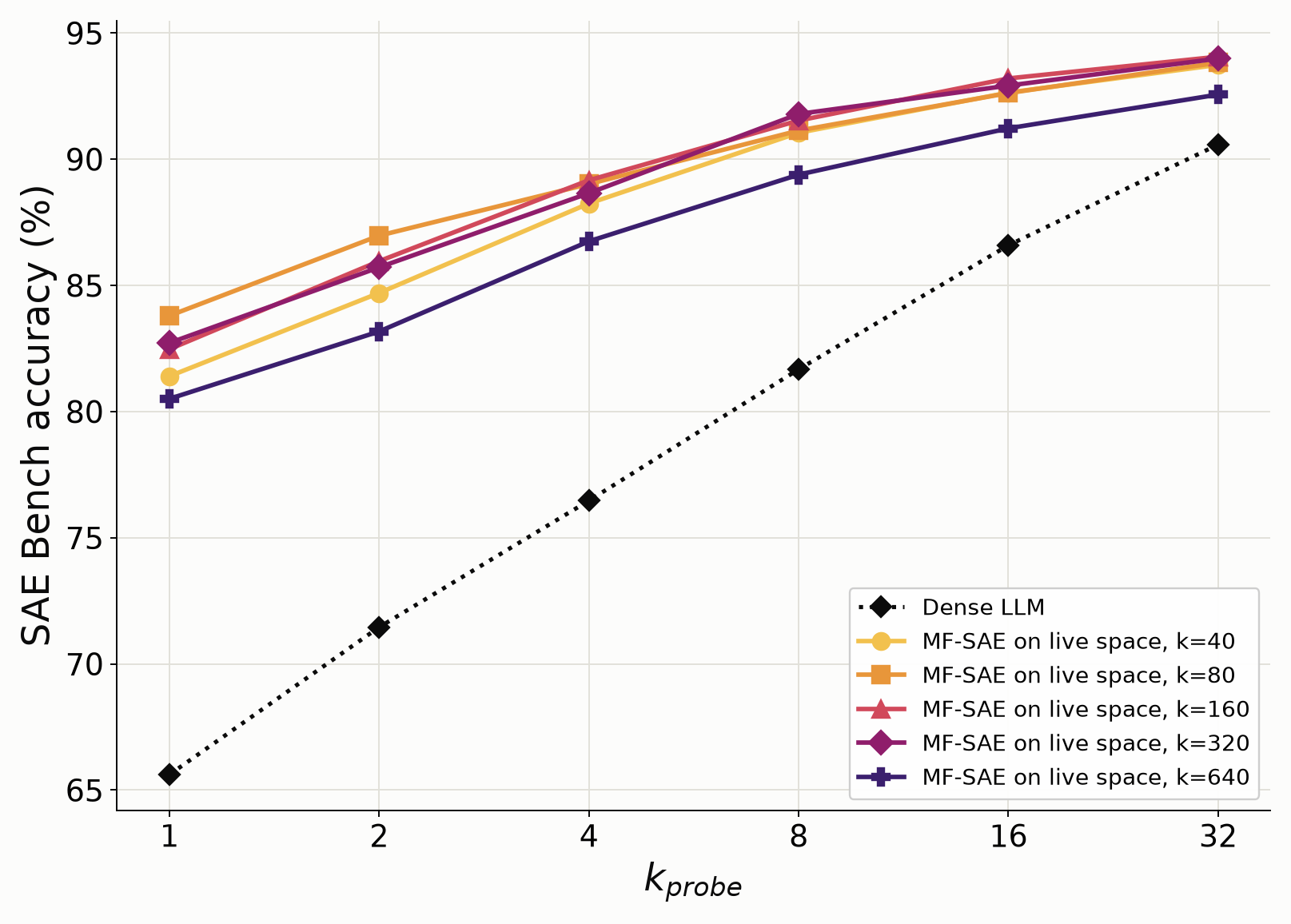}
        \caption{Live, layer 12}
        \label{fig:a32-mf-sb-b}
    \end{subfigure}

    \vspace{6pt}

    \begin{subfigure}[b]{0.48\textwidth}
        \centering
        \includegraphics[width=\linewidth, height=0.5\linewidth]{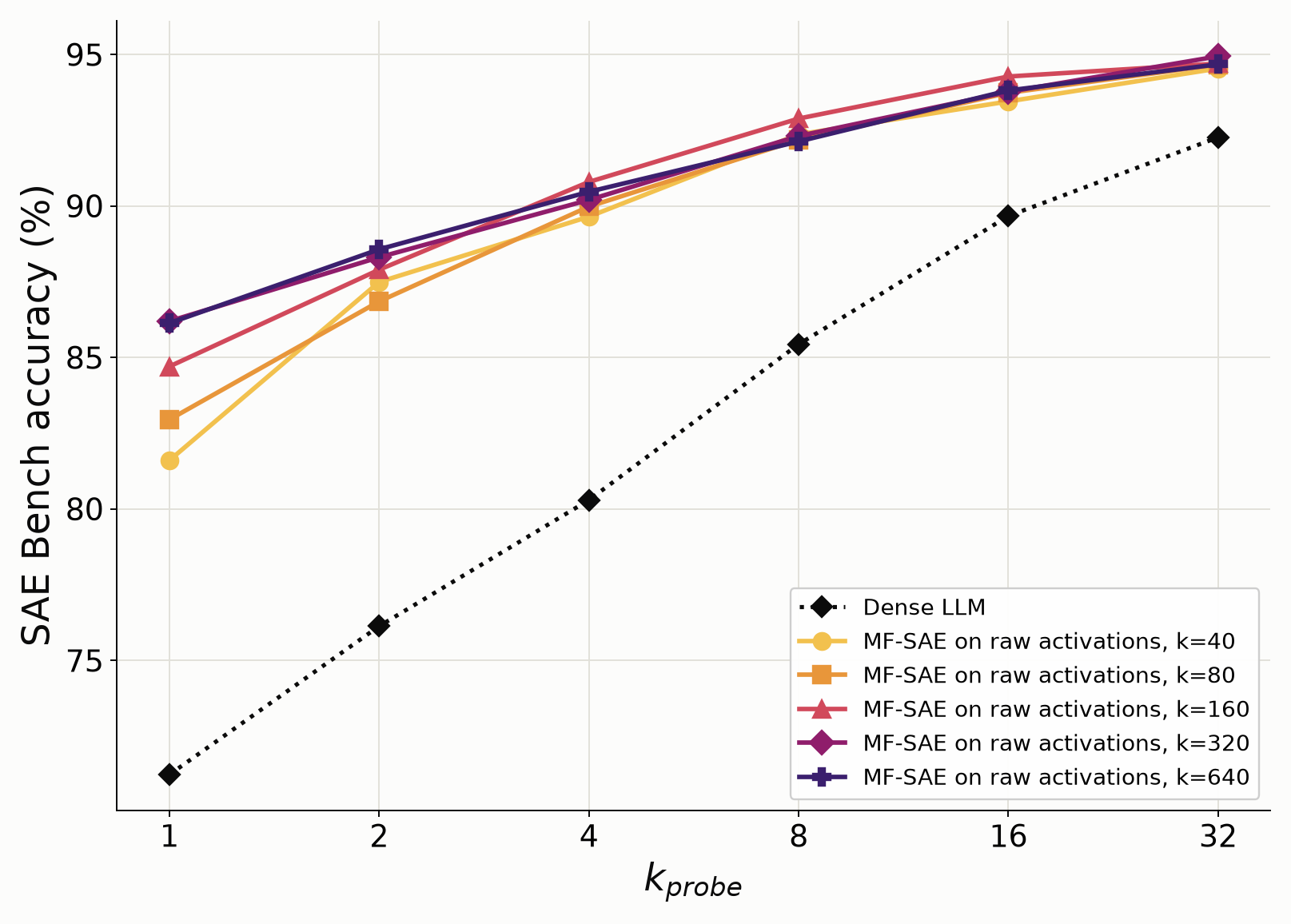}
        \caption{Raw, layer 20}
        \label{fig:a32-mf-sb-c}
    \end{subfigure}
    \hfill
    \begin{subfigure}[b]{0.48\textwidth}
        \centering
        \includegraphics[width=\linewidth, height=0.5\linewidth]{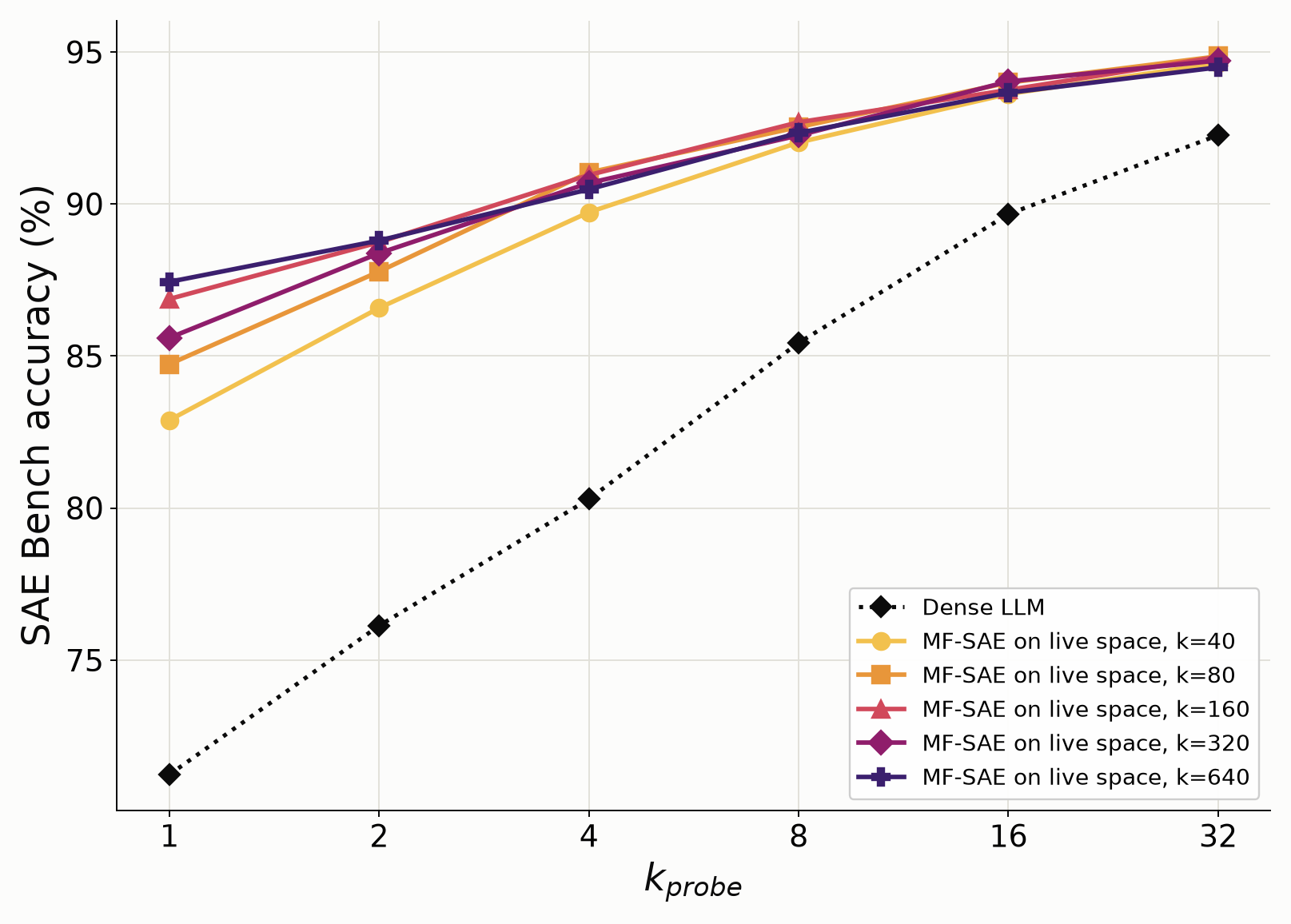}
        \caption{Live, layer 20}
        \label{fig:a32-mf-sb-d}
    \end{subfigure}
    \caption{\textbf{SAEBench, MF-SAE across sparsity levels (Gemma-2-2B, 32K dictionary).} SB-Acc against $k_{\text{probe}}$ for the MF-SAE at $k\in\{40,80,160,320\}$. Columns are the raw and live MF-SAE; rows are layers 12 and 20. $k=320$ is among the best sparsity levels but not the best in every panel.}
    \label{fig:a32-mf-sb}
\end{figure*}

\begin{figure*}[t]
    \centering
    \begin{subfigure}[b]{0.48\textwidth}
        \centering
        \includegraphics[width=\linewidth, height=0.5\linewidth]{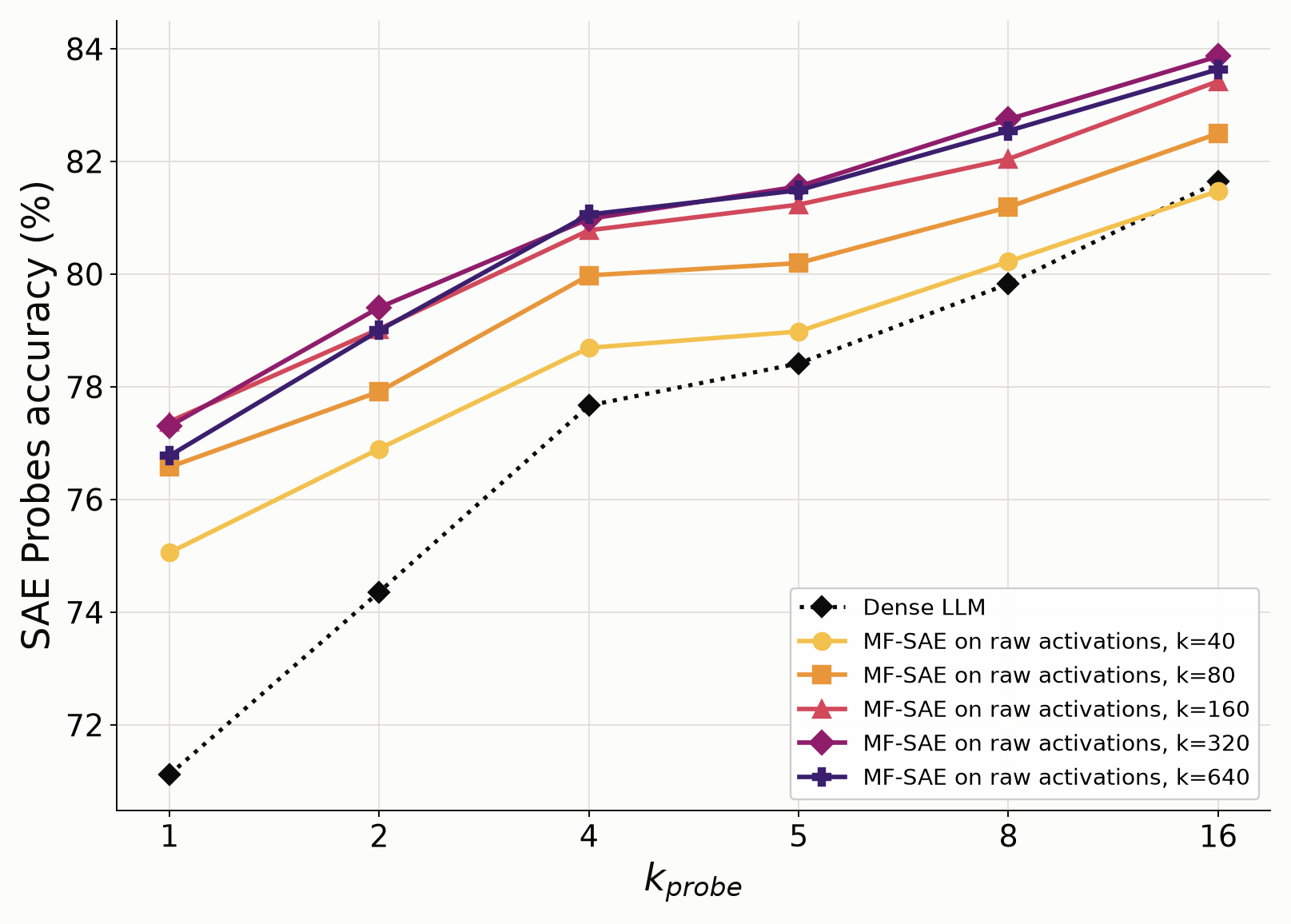}
        \caption{Raw, layer 12}
        \label{fig:a32-mf-sp-a}
    \end{subfigure}
    \hfill
    \begin{subfigure}[b]{0.48\textwidth}
        \centering
        \includegraphics[width=\linewidth, height=0.5\linewidth]{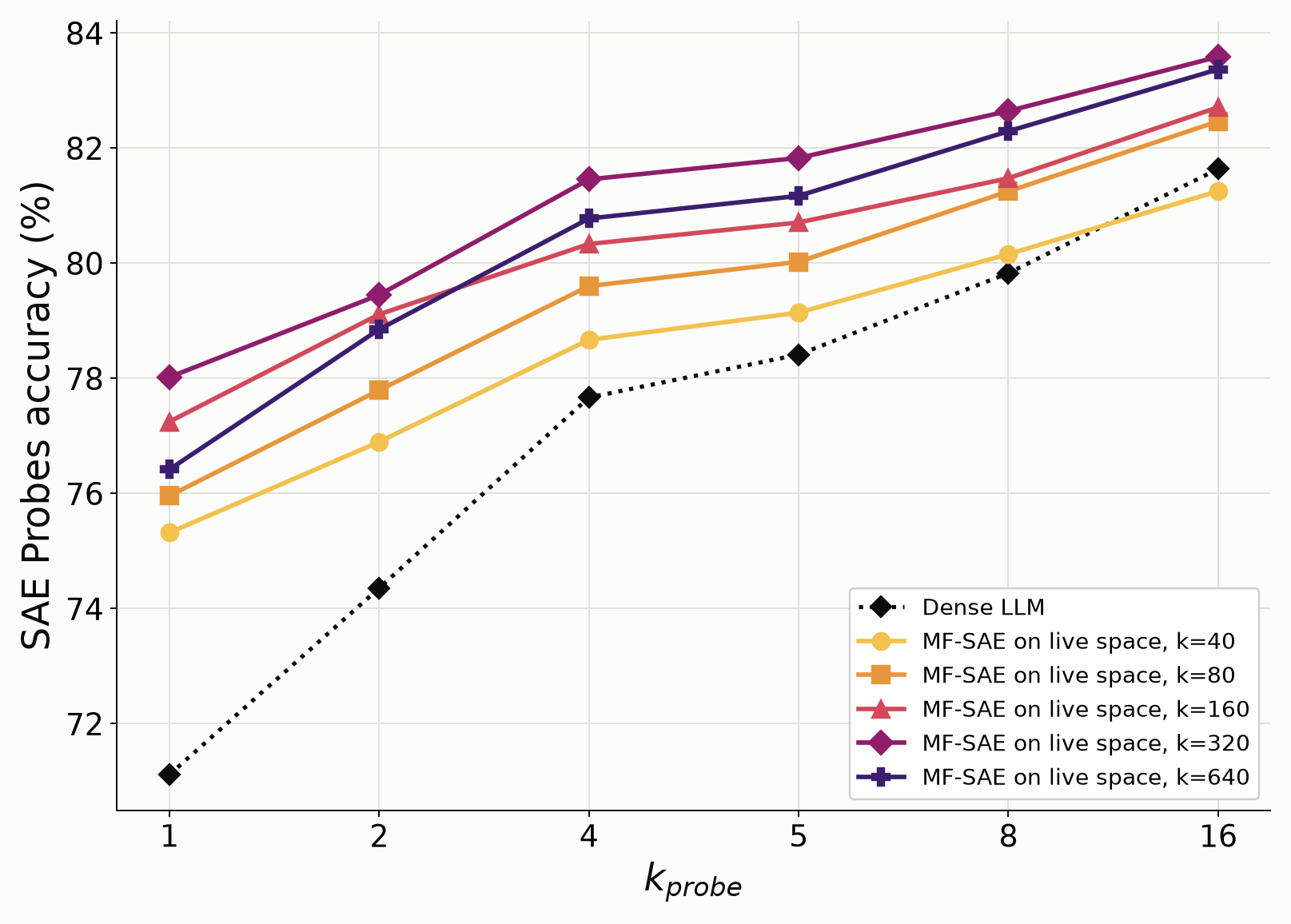}
        \caption{Live, layer 12}
        \label{fig:a32-mf-sp-b}
    \end{subfigure}

    \vspace{6pt}

    \begin{subfigure}[b]{0.48\textwidth}
        \centering
        \includegraphics[width=\linewidth, height=0.5\linewidth]{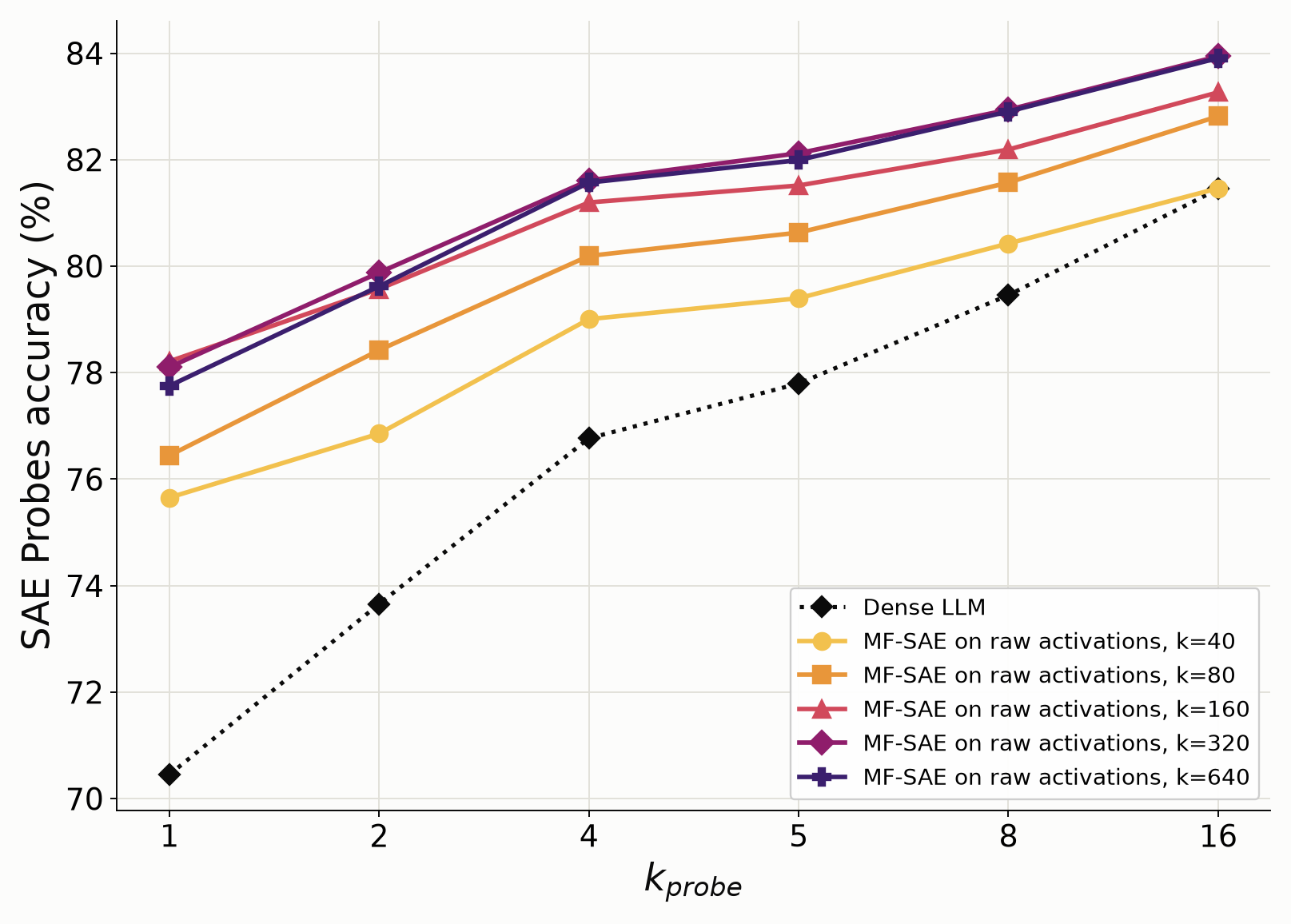}
        \caption{Raw, layer 20}
        \label{fig:a32-mf-sp-c}
    \end{subfigure}
    \hfill
    \begin{subfigure}[b]{0.48\textwidth}
        \centering
        \includegraphics[width=\linewidth, height=0.5\linewidth]{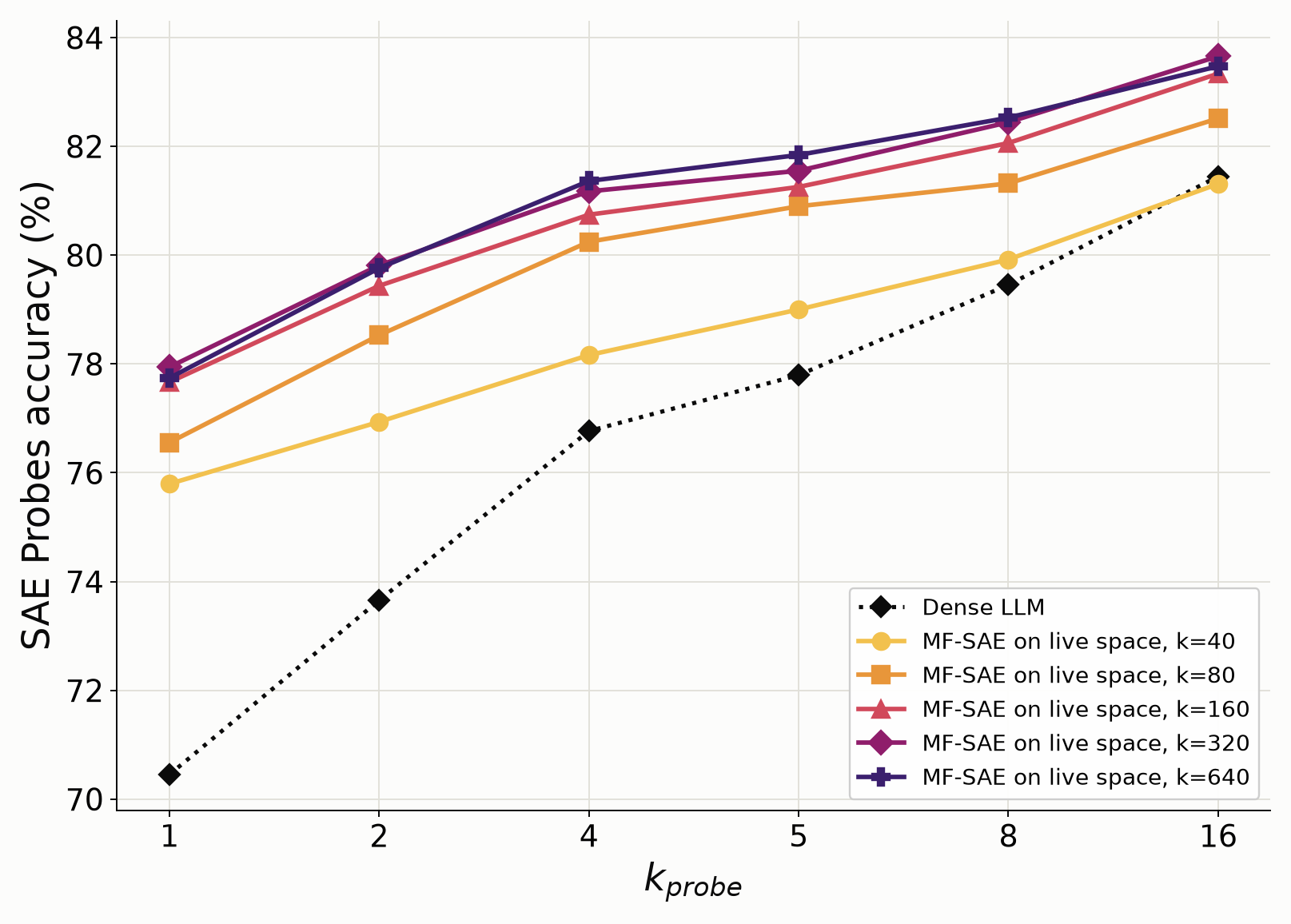}
        \caption{Live, layer 20}
        \label{fig:a32-mf-sp-d}
    \end{subfigure}
    \caption{\textbf{SAE Probes, MF-SAE across sparsity levels (Gemma-2-2B, 32K dictionary).} Mean test accuracy against $k_{\text{probe}}$ for the MF-SAE at $k\in\{40,80,160,320\}$, with the same layout as Figure~\ref{fig:a32-mf-sb}. $k=320$ is consistently the best sparsity level.}
    \label{fig:a32-mf-sp}
\end{figure*}

\begin{figure*}[t]
    \centering
    \begin{subfigure}[b]{0.48\textwidth}
        \centering
        \includegraphics[width=\linewidth, height=0.5\linewidth]{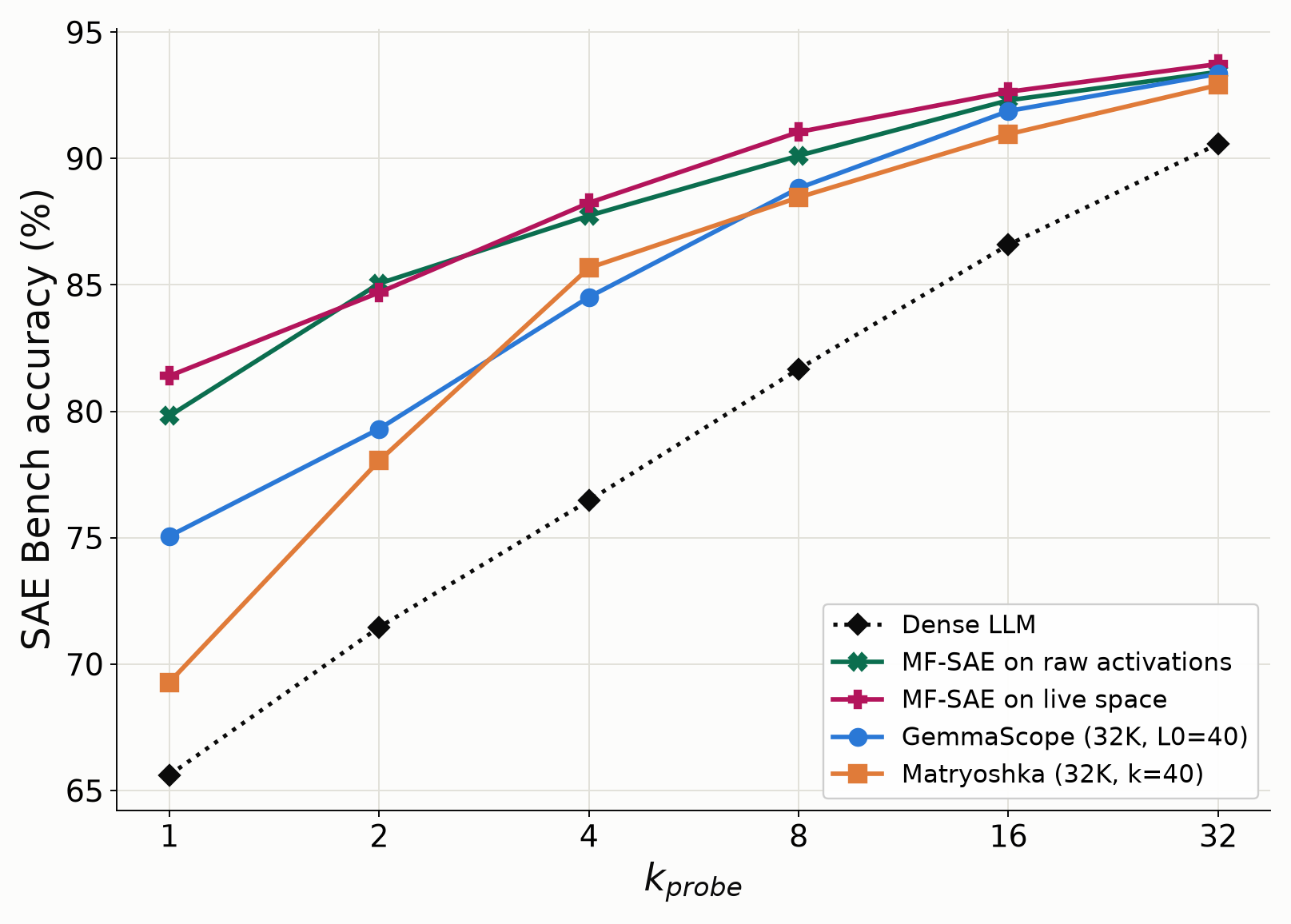}
        \caption{$k=40$}
        \label{fig:a32-l12-sb-a}
    \end{subfigure}
    \hfill
    \begin{subfigure}[b]{0.48\textwidth}
        \centering
        \includegraphics[width=\linewidth, height=0.5\linewidth]{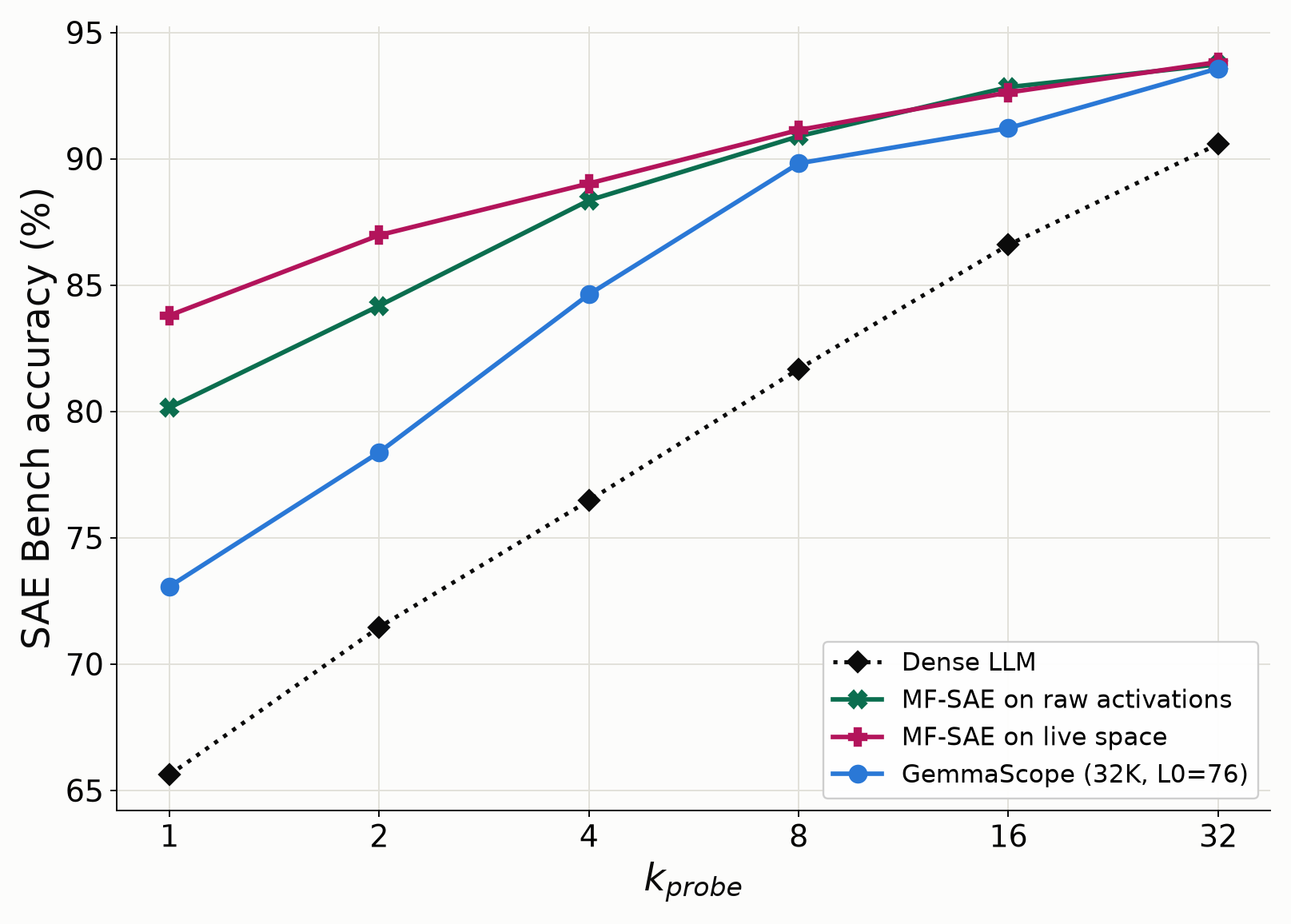}
        \caption{$k=80$}
        \label{fig:a32-l12-sb-b}
    \end{subfigure}

    \vspace{6pt}

    \begin{subfigure}[b]{0.48\textwidth}
        \centering
        \includegraphics[width=\linewidth, height=0.5\linewidth]{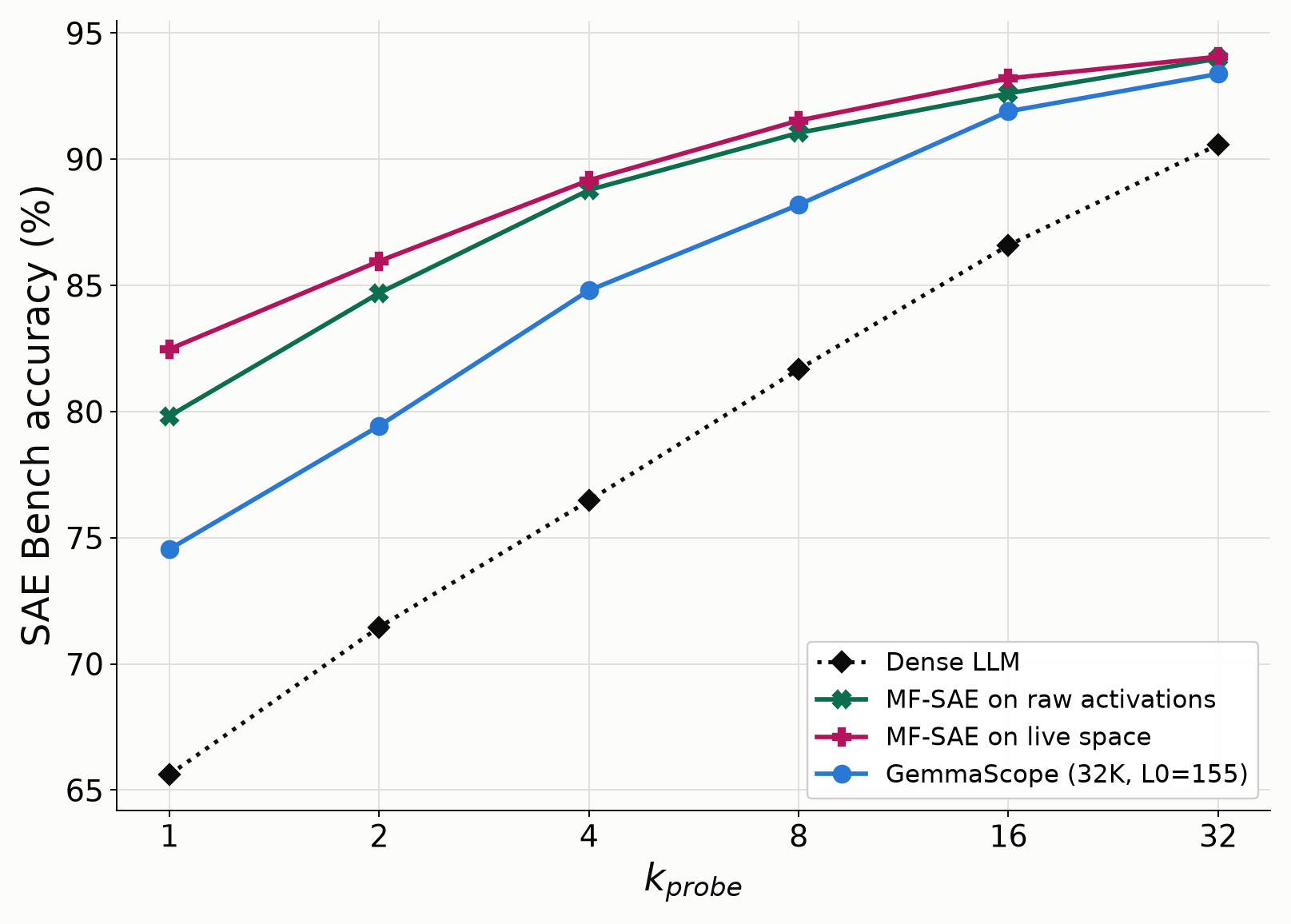}
        \caption{$k=160$}
        \label{fig:a32-l12-sb-c}
    \end{subfigure}
    \hfill
    \begin{subfigure}[b]{0.48\textwidth}
        \centering
        \includegraphics[width=\linewidth, height=0.5\linewidth]{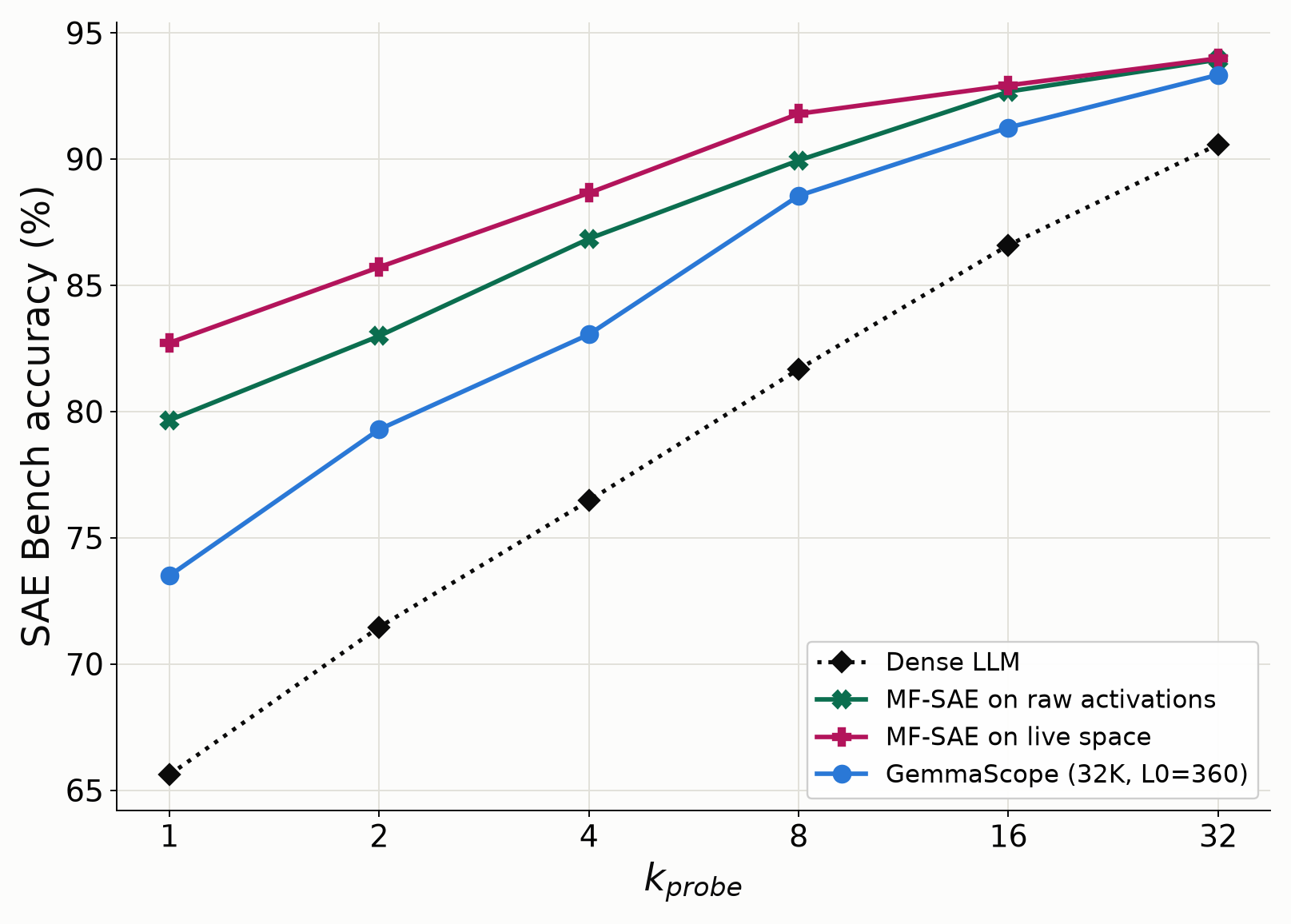}
        \caption{$k=320$}
        \label{fig:a32-l12-sb-d}
    \end{subfigure}
    \caption{\textbf{SAEBench, layer 12, 32K dictionary.} SB-Acc against $k_{\text{probe}}$ for the raw and live MF-SAEs, the Gemma Scope 32K SAE of closest average L0, and, in (a), the Matryoshka SAE.}
    \label{fig:a32-l12-sb}
\end{figure*}

\begin{figure*}[t]
    \centering
    \begin{subfigure}[b]{0.48\textwidth}
        \centering
        \includegraphics[width=\linewidth, height=0.5\linewidth]{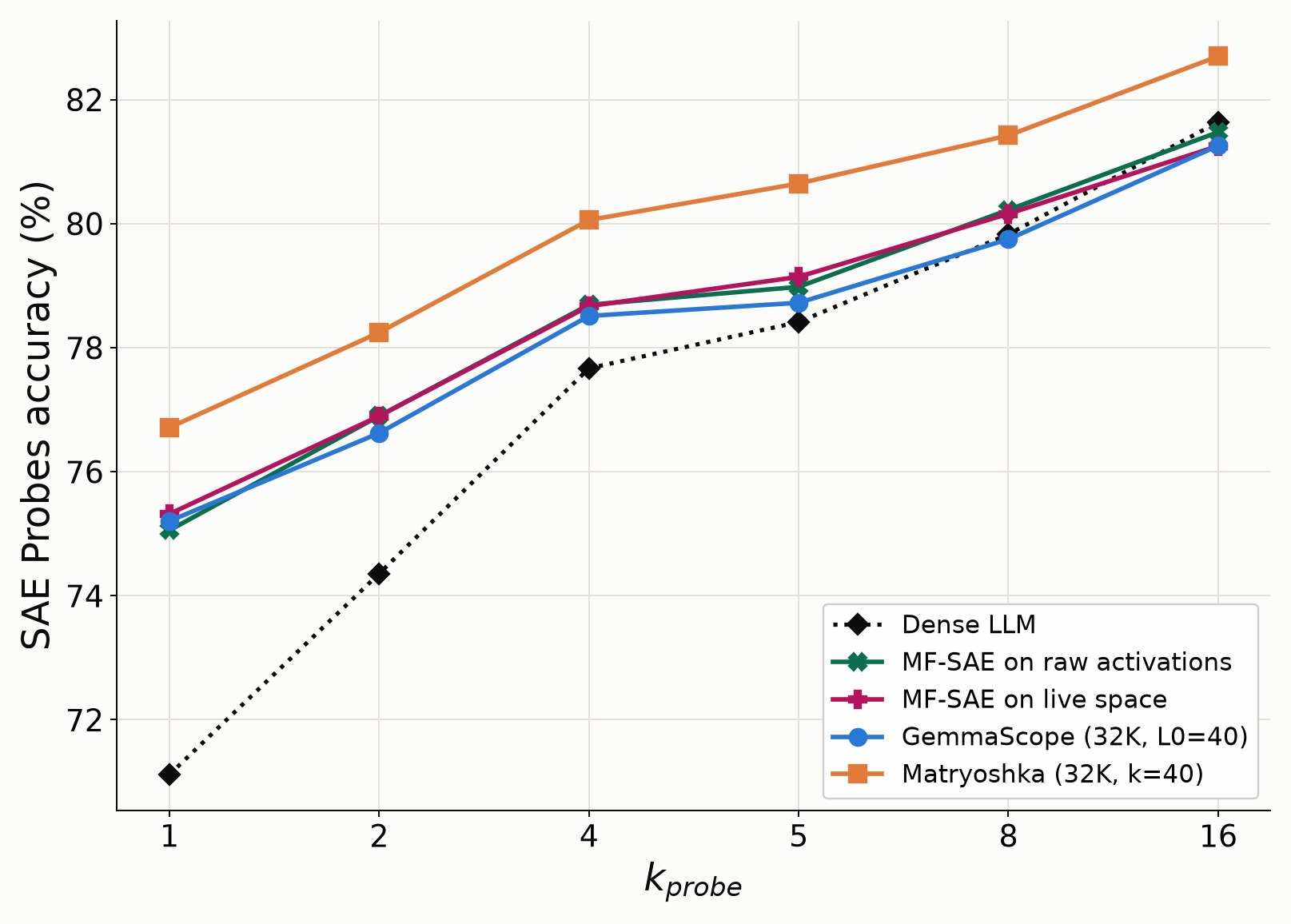}
        \caption{$k=40$}
        \label{fig:a32-l12-sp-a}
    \end{subfigure}
    \hfill
    \begin{subfigure}[b]{0.48\textwidth}
        \centering
        \includegraphics[width=\linewidth, height=0.5\linewidth]{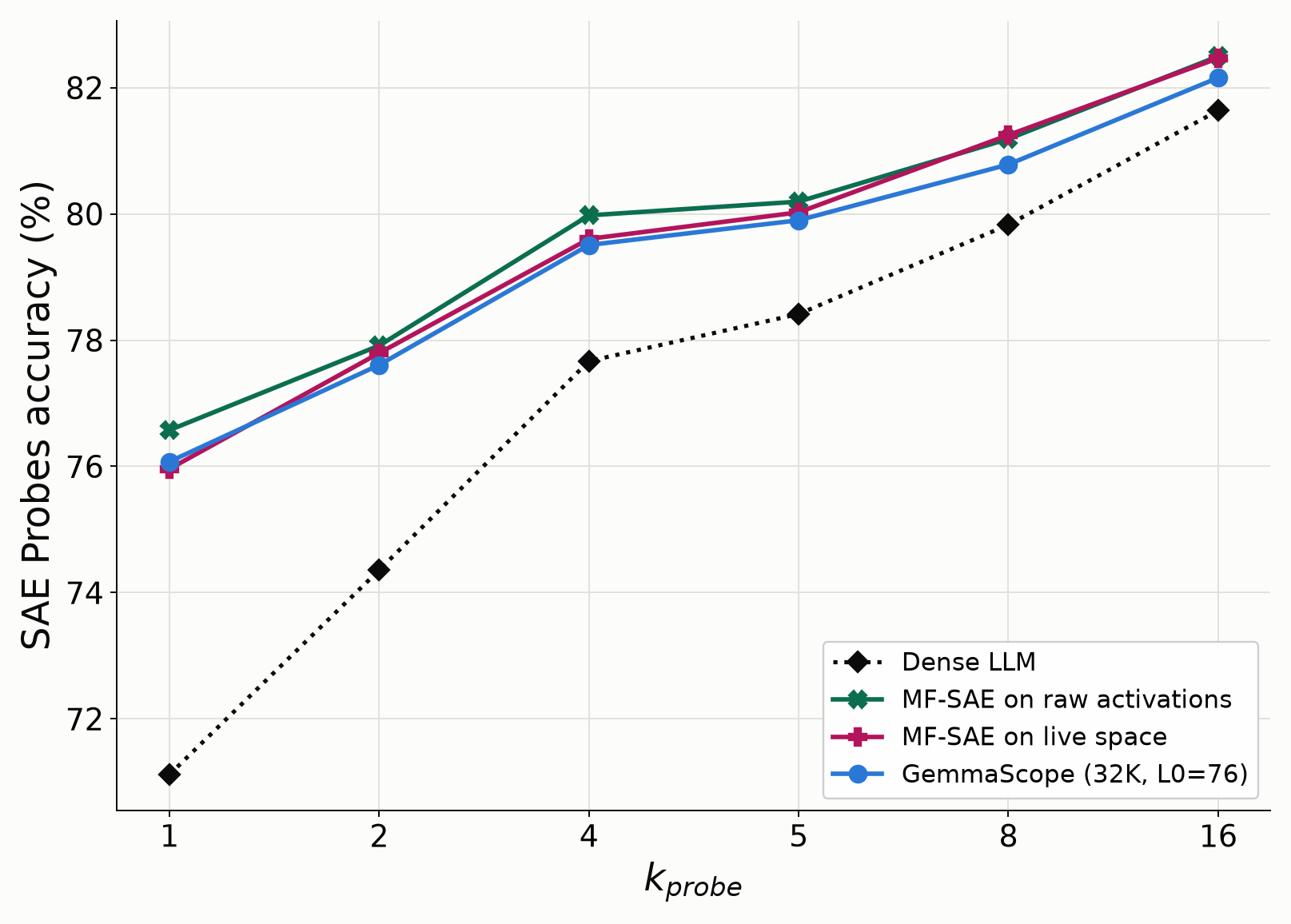}
        \caption{$k=80$}
        \label{fig:a32-l12-sp-b}
    \end{subfigure}

    \vspace{6pt}

    \begin{subfigure}[b]{0.48\textwidth}
        \centering
        \includegraphics[width=\linewidth, height=0.5\linewidth]{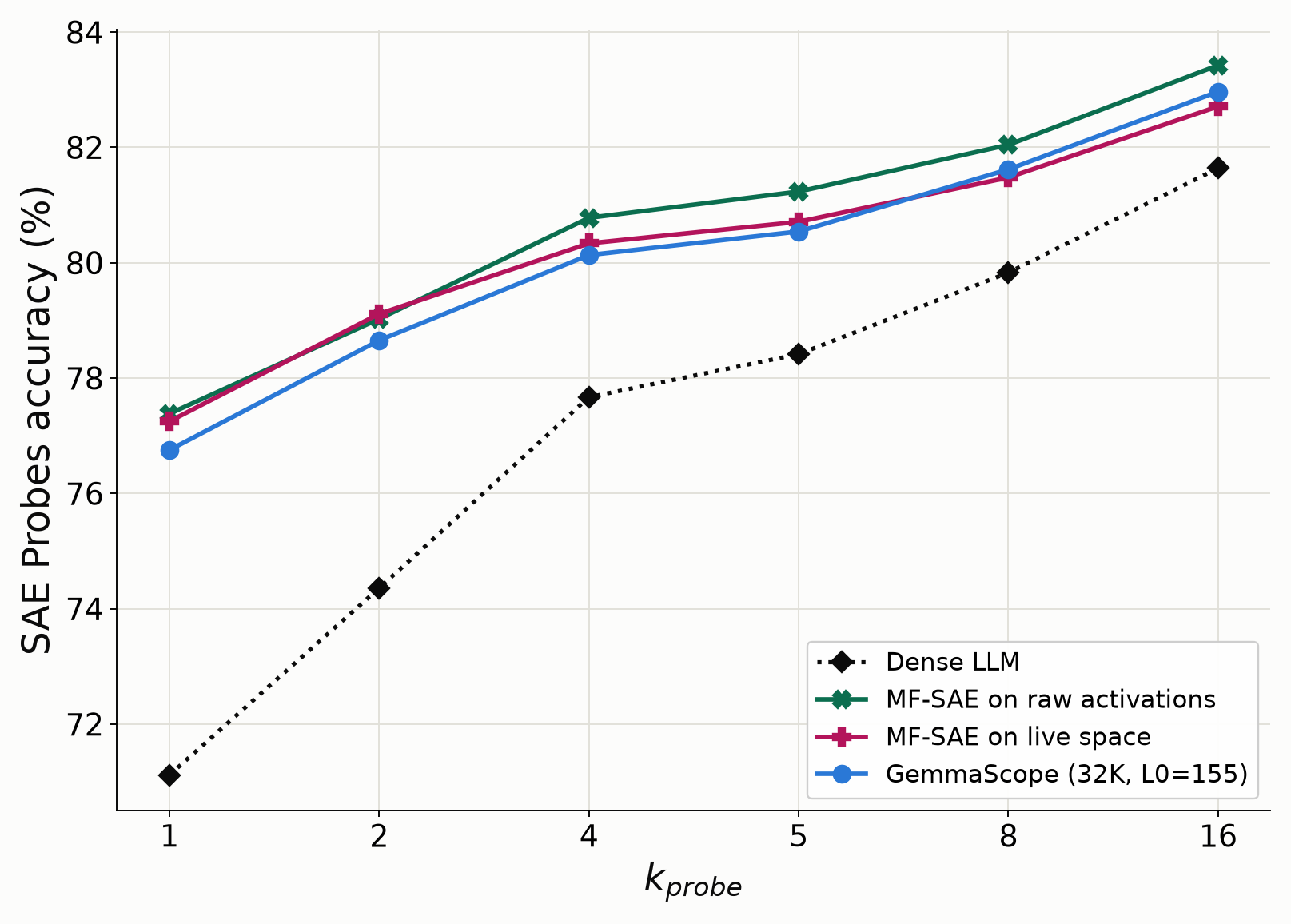}
        \caption{$k=160$}
        \label{fig:a32-l12-sp-c}
    \end{subfigure}
    \hfill
    \begin{subfigure}[b]{0.48\textwidth}
        \centering
        \includegraphics[width=\linewidth, height=0.5\linewidth]{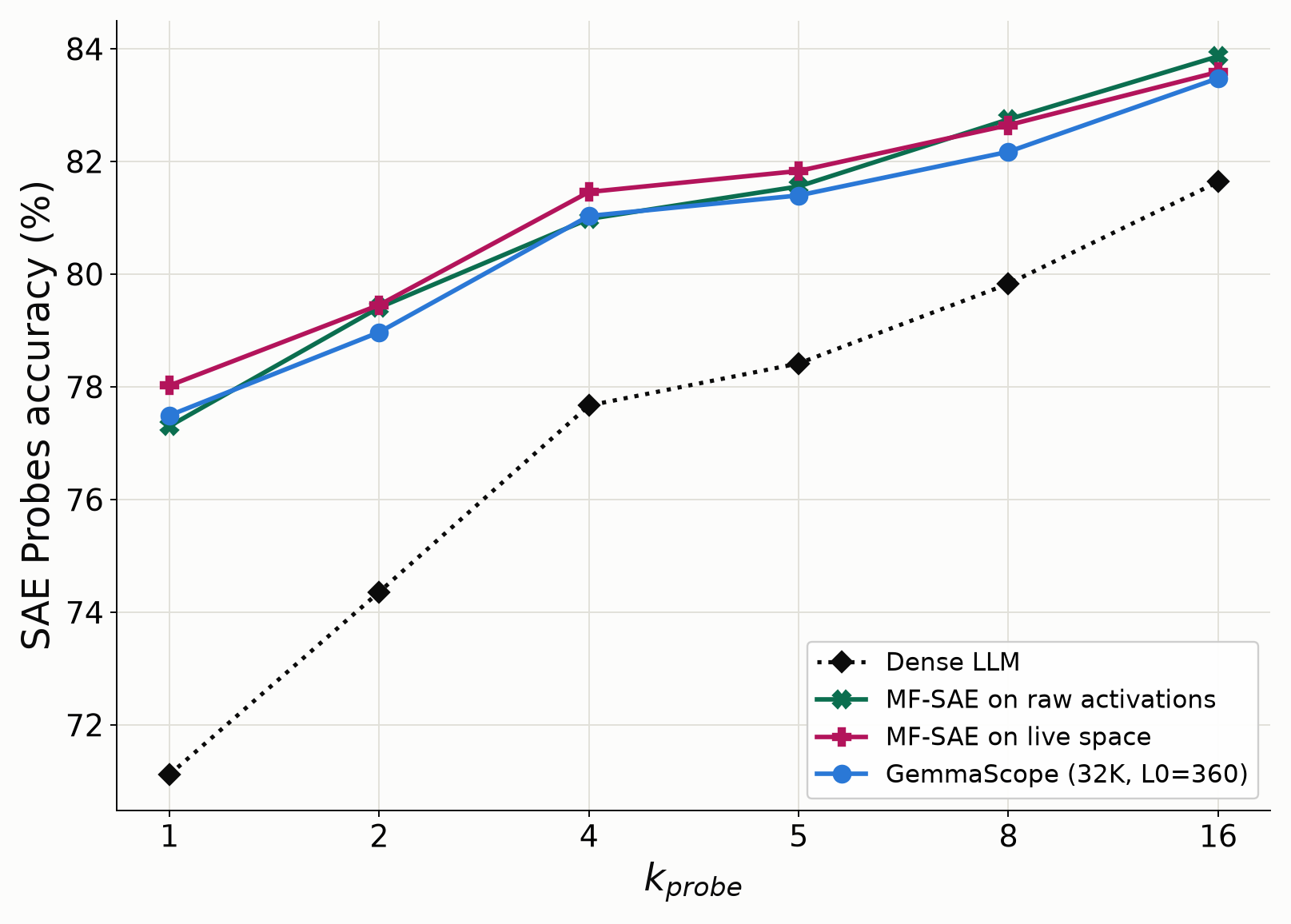}
        \caption{$k=320$}
        \label{fig:a32-l12-sp-d}
    \end{subfigure}
    \caption{\textbf{SAE Probes, layer 12, 32K dictionary.} Mean test accuracy against $k_{\text{probe}}$, with the same methods and layout as Figure~\ref{fig:a32-l12-sb}. In (a), the Matryoshka SAE is comparable to the MF-SAEs.}
    \label{fig:a32-l12-sp}
\end{figure*}

\begin{figure*}[t]
    \centering
    \begin{subfigure}[b]{0.48\textwidth}
        \centering
        \includegraphics[width=\linewidth, height=0.5\linewidth]{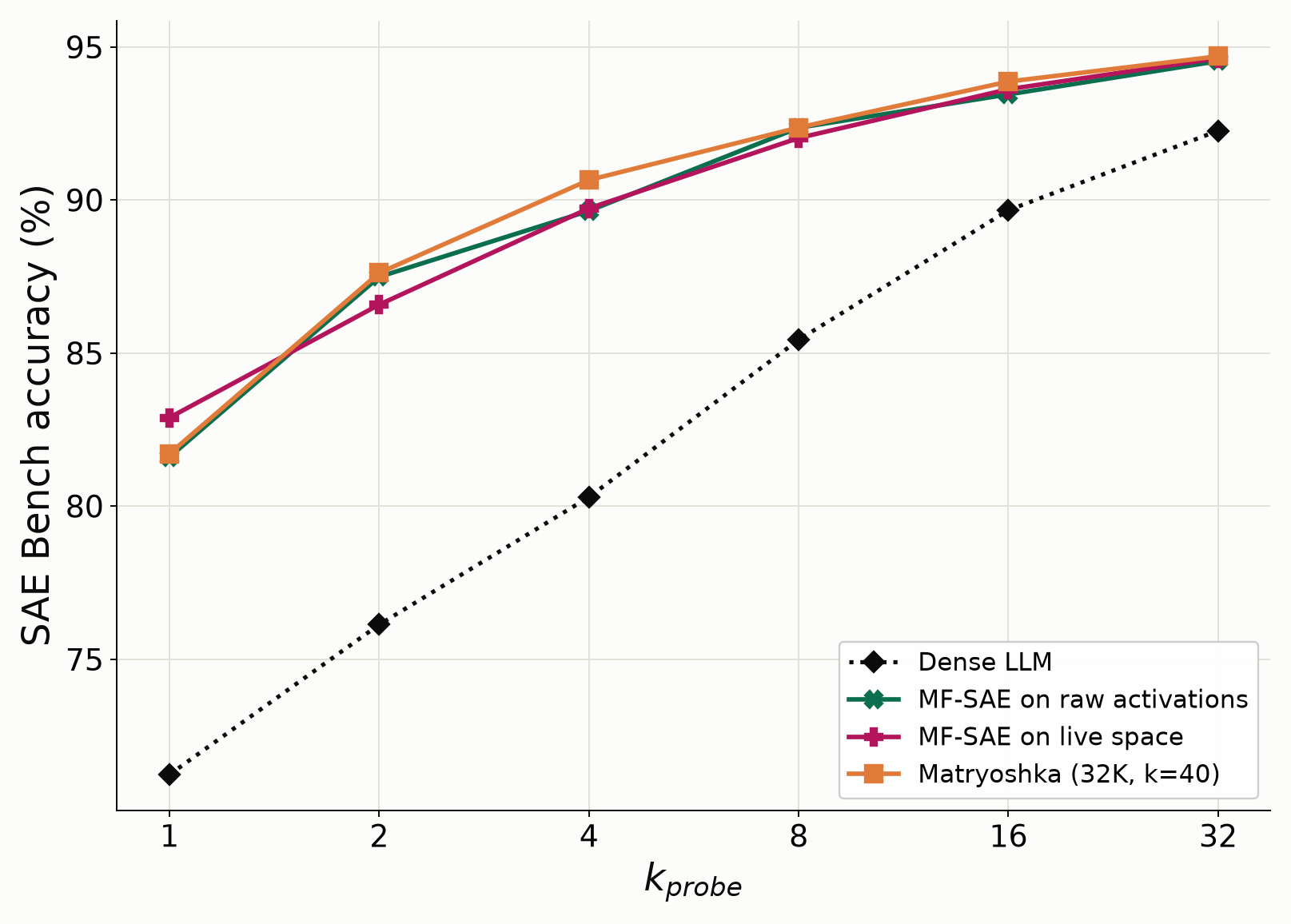}
        \caption{$k=40$}
        \label{fig:a32-l20-sb-a}
    \end{subfigure}
    \hfill
    \begin{subfigure}[b]{0.48\textwidth}
        \centering
        \includegraphics[width=\linewidth, height=0.5\linewidth]{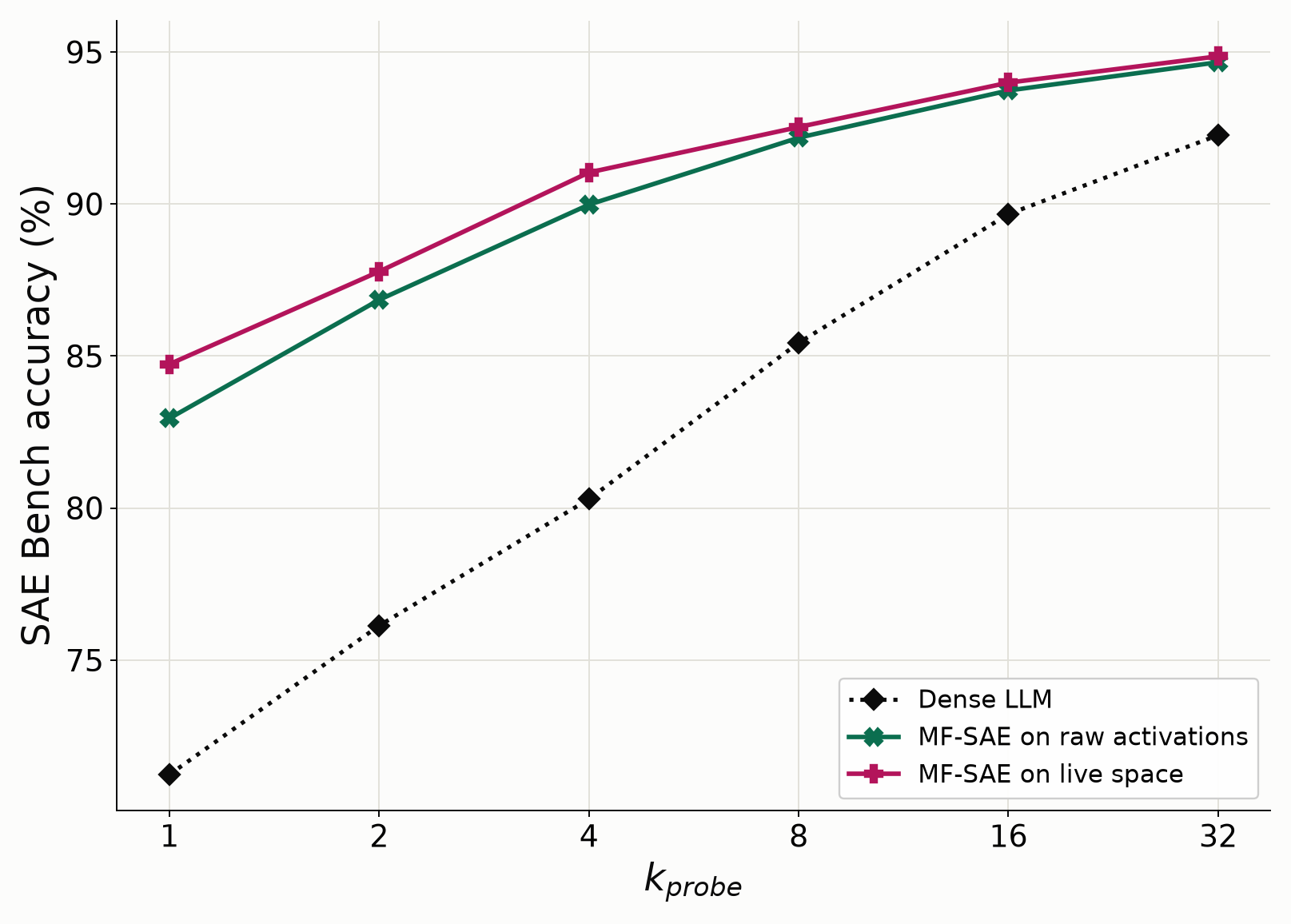}
        \caption{$k=80$}
        \label{fig:a32-l20-sb-b}
    \end{subfigure}

    \vspace{6pt}

    \begin{subfigure}[b]{0.48\textwidth}
        \centering
        \includegraphics[width=\linewidth, height=0.5\linewidth]{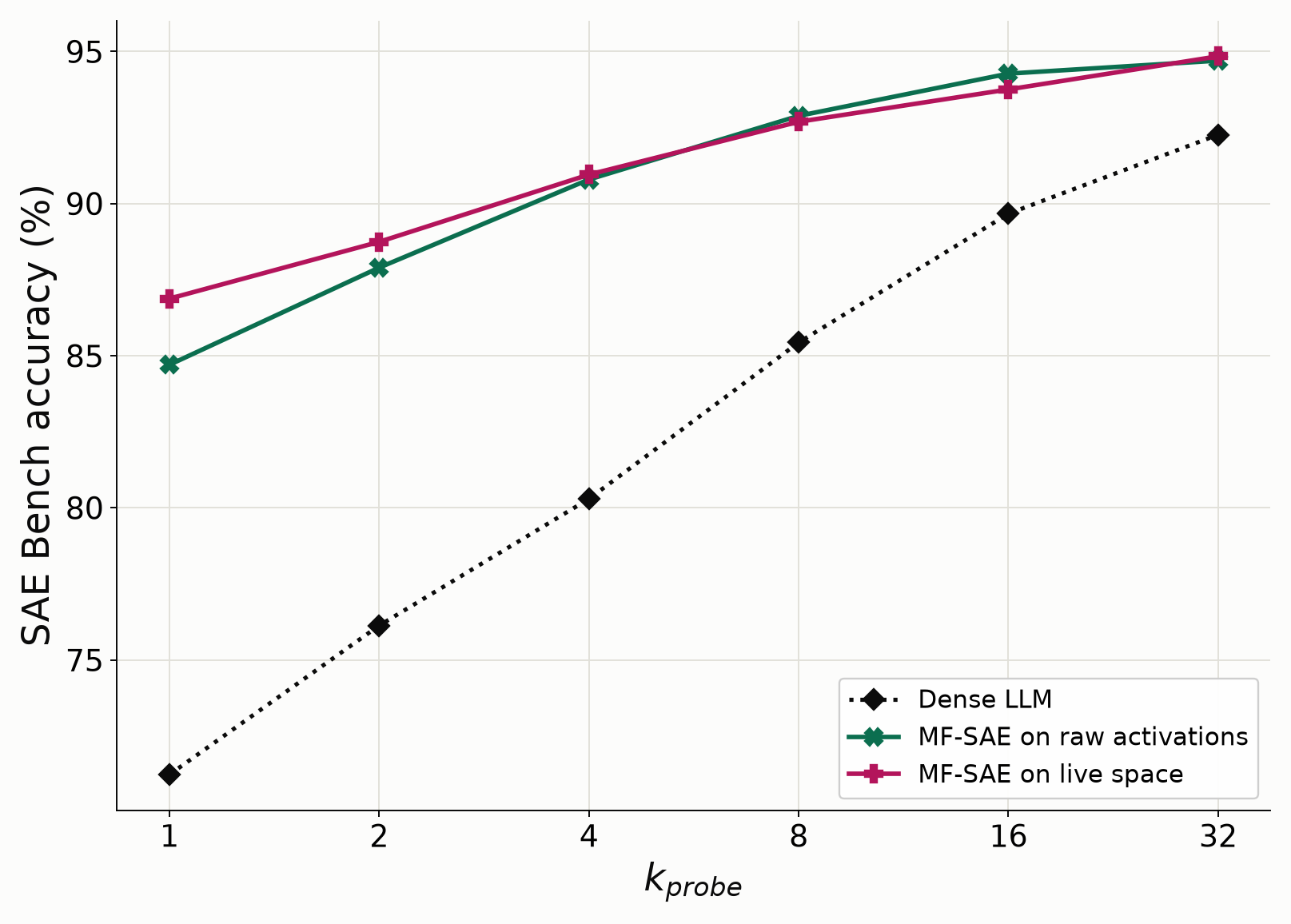}
        \caption{$k=160$}
        \label{fig:a32-l20-sb-c}
    \end{subfigure}
    \hfill
    \begin{subfigure}[b]{0.48\textwidth}
        \centering
        \includegraphics[width=\linewidth, height=0.5\linewidth]{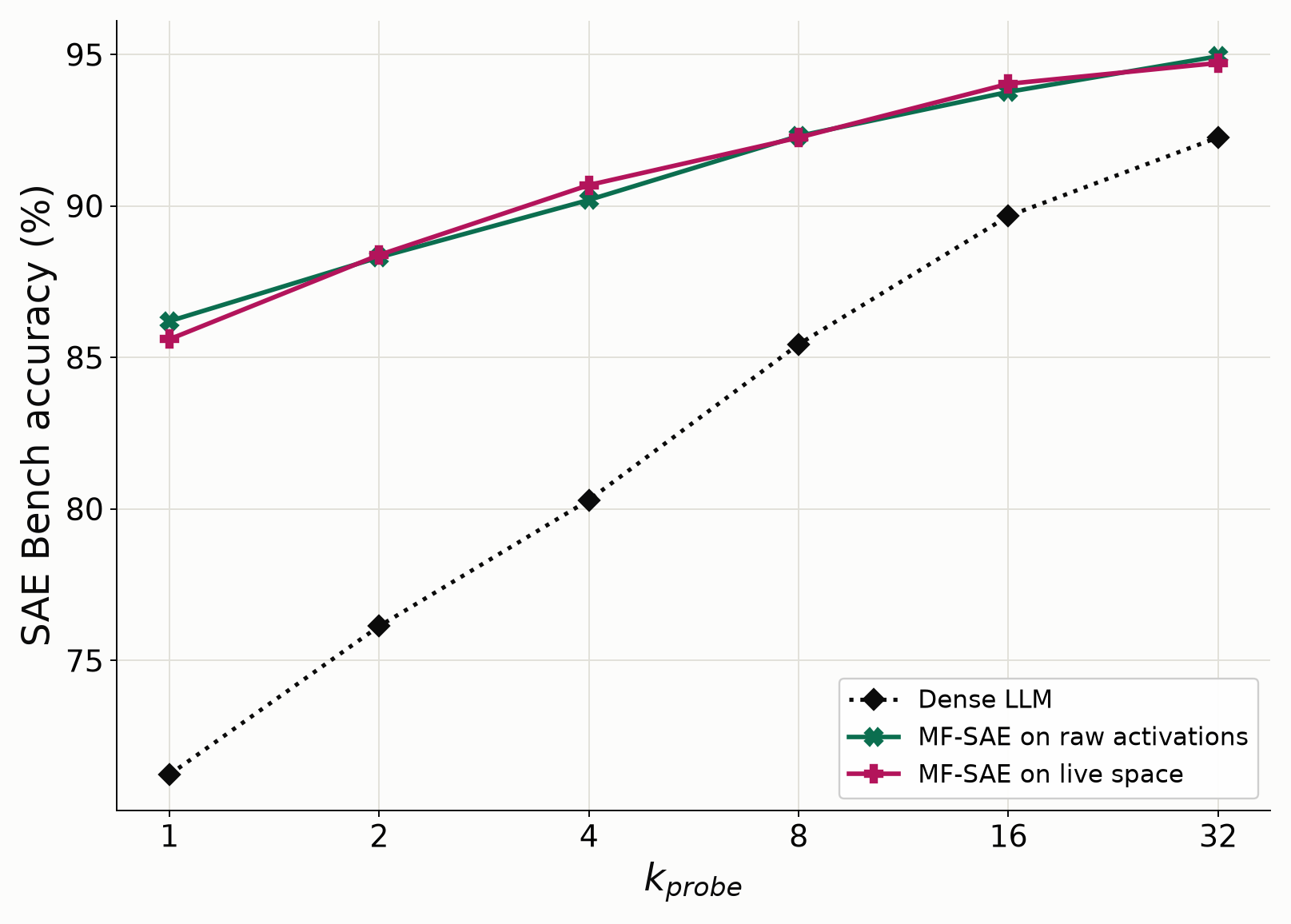}
        \caption{$k=320$}
        \label{fig:a32-l20-sb-d}
    \end{subfigure}
    \caption{\textbf{SAEBench, layer 20, 32K dictionary.} SB-Acc against $k_{\text{probe}}$ for the raw and live MF-SAEs and, in (a), the Matryoshka SAE. No SAEBench or Gemma Scope 32K SAEs are available at this layer.}
    \label{fig:a32-l20-sb}
\end{figure*}

\begin{figure*}[t]
    \centering
    \begin{subfigure}[b]{0.48\textwidth}
        \centering
        \includegraphics[width=\linewidth, height=0.5\linewidth]{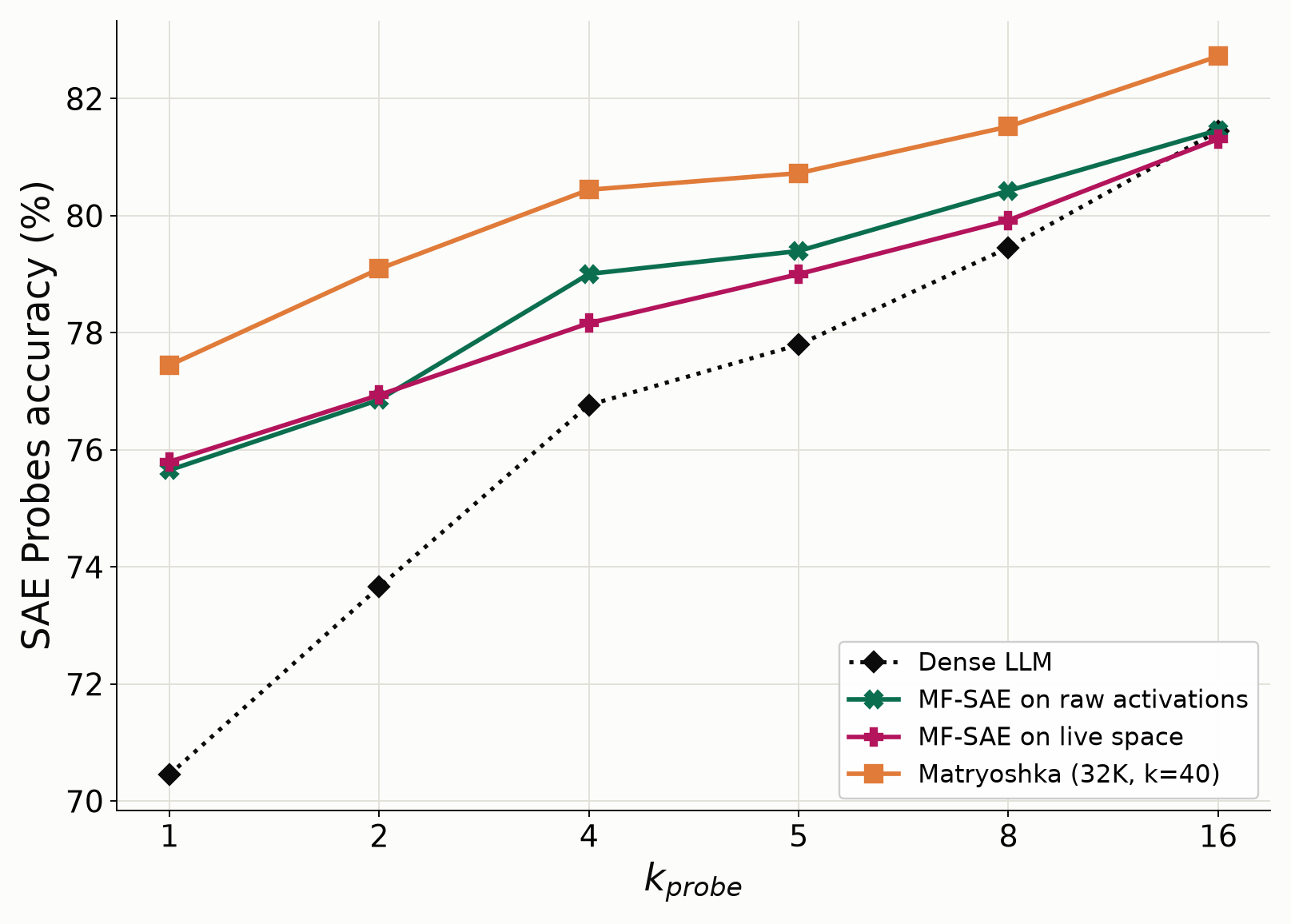}
        \caption{$k=40$}
        \label{fig:a32-l20-sp-a}
    \end{subfigure}
    \hfill
    \begin{subfigure}[b]{0.48\textwidth}
        \centering
        \includegraphics[width=\linewidth, height=0.5\linewidth]{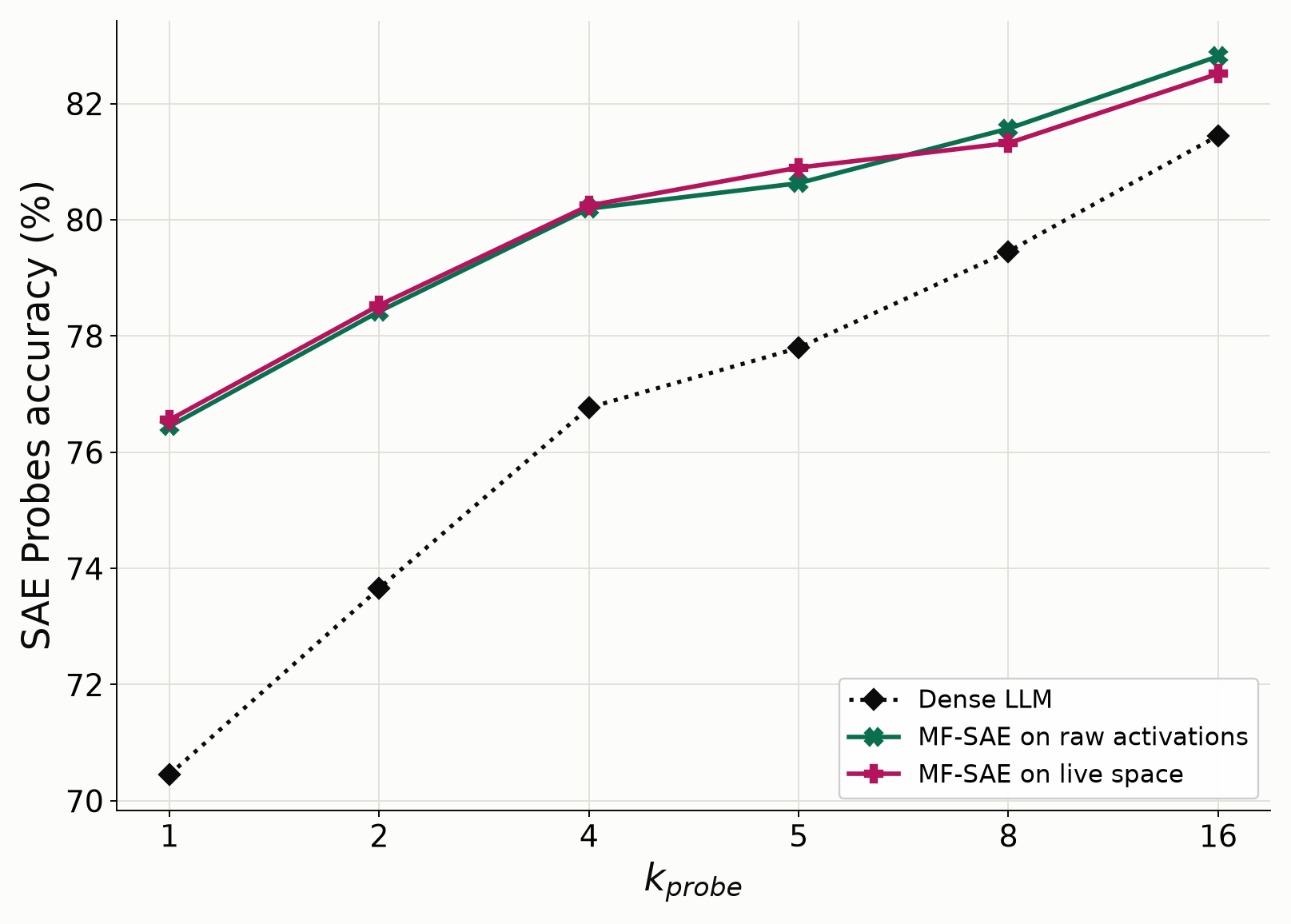}
        \caption{$k=80$}
        \label{fig:a32-l20-sp-b}
    \end{subfigure}

    \vspace{6pt}

    \begin{subfigure}[b]{0.48\textwidth}
        \centering
        \includegraphics[width=\linewidth, height=0.5\linewidth]{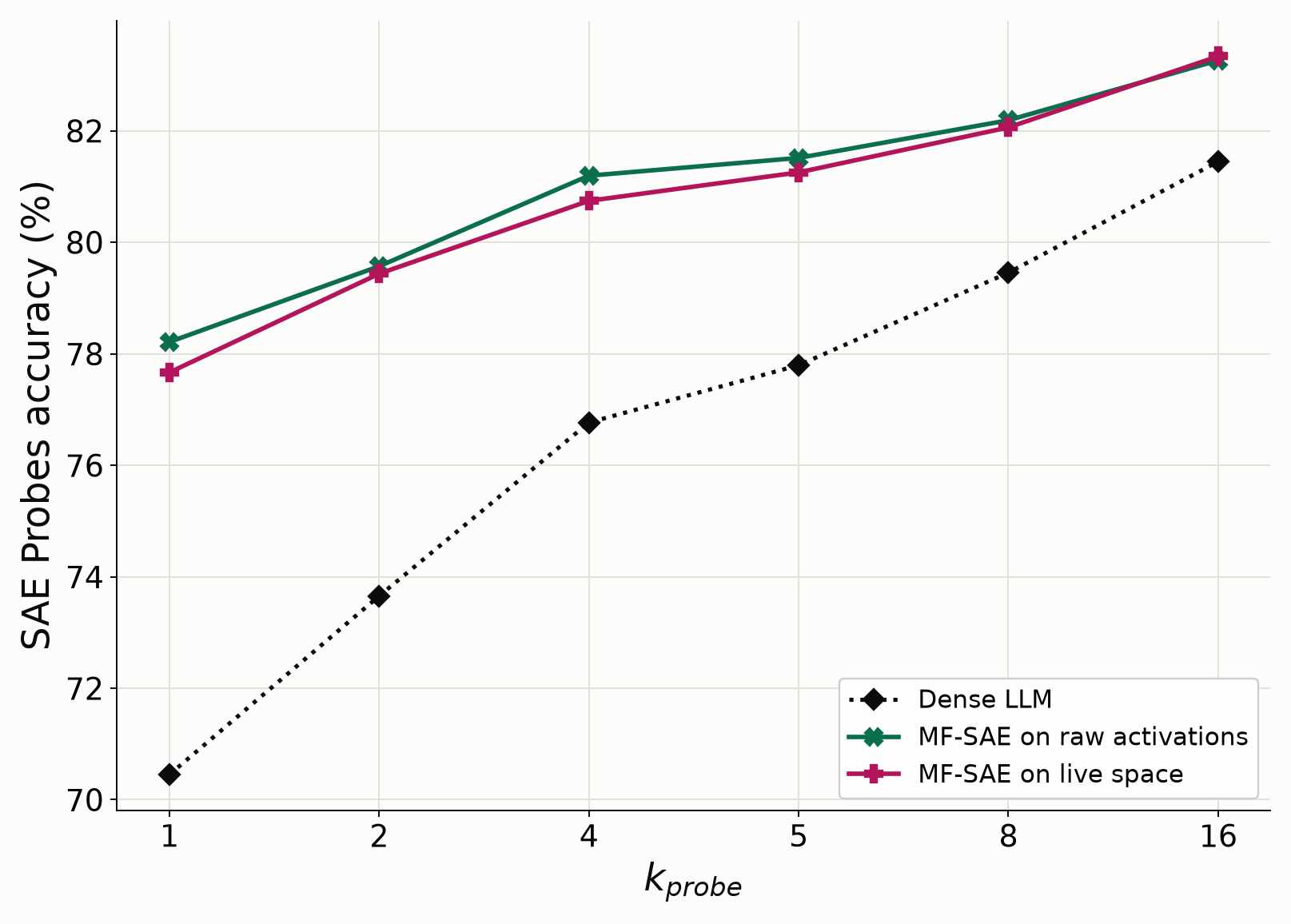}
        \caption{$k=160$}
        \label{fig:a32-l20-sp-c}
    \end{subfigure}
    \hfill
    \begin{subfigure}[b]{0.48\textwidth}
        \centering
        \includegraphics[width=\linewidth, height=0.5\linewidth]{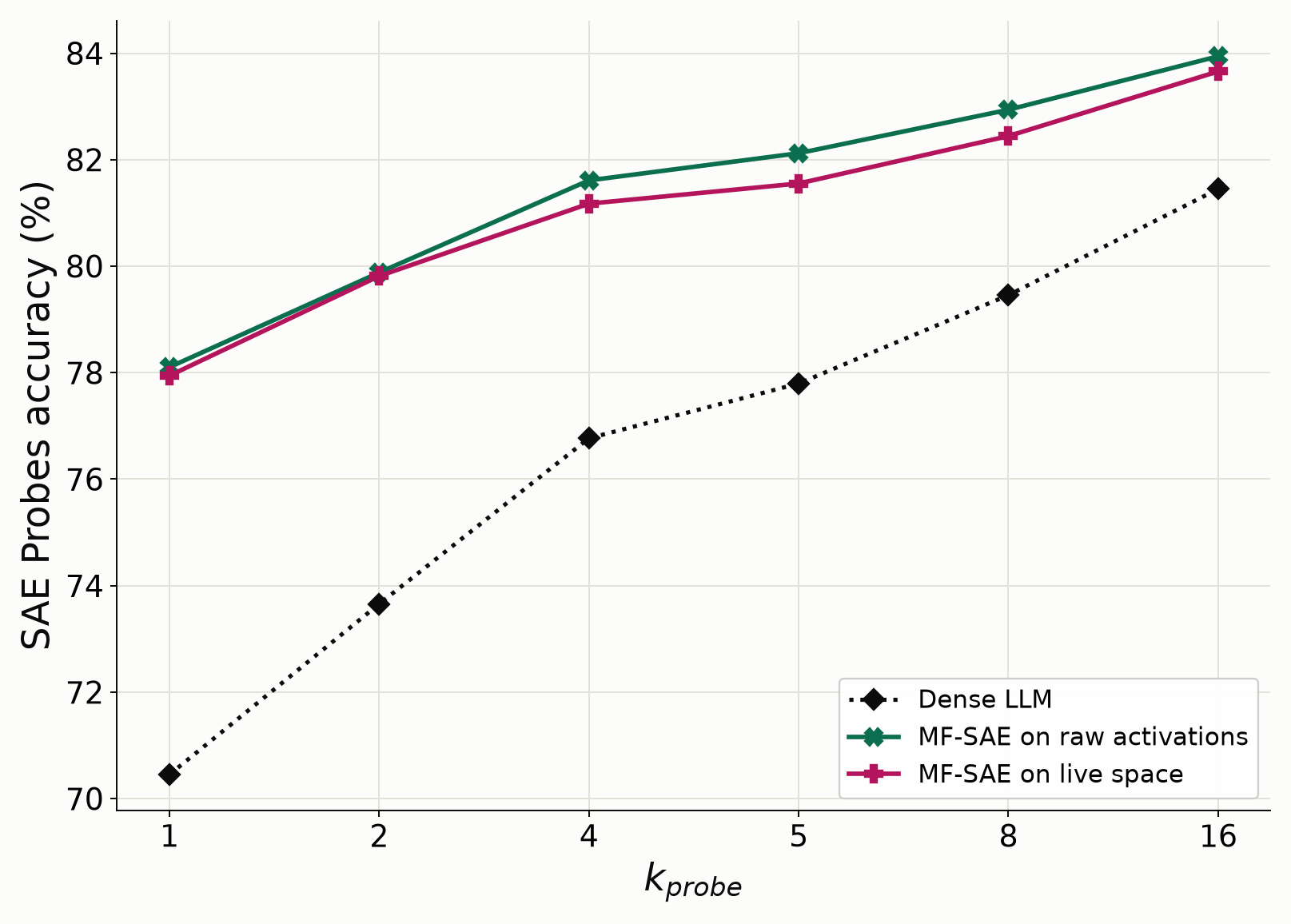}
        \caption{$k=320$}
        \label{fig:a32-l20-sp-d}
    \end{subfigure}
    \caption{\textbf{SAE Probes, layer 20, 32K dictionary.} Mean test accuracy against $k_{\text{probe}}$, with the same methods and layout as Figure~\ref{fig:a32-l20-sb}. In (a), the Matryoshka SAE outperforms the MF-SAEs.}
    \label{fig:a32-l20-sp}
\end{figure*}

This section repeats Appendix~\ref{app:sparsity-16k} with a 32K dictionary. As with 16K, $k=320$ gives the best MF-SAE on SAE Probes, while on SAEBench it is among the best sparsity levels but not always the best (Figures~\ref{fig:a32-mf-sb} and~\ref{fig:a32-mf-sp}). SAEBench does not release 32K SAEs. At layer 12 the baselines are Gemma Scope 32K and the Matryoshka SAE at L0 $=40$; at layer 20, Gemma Scope has no 32K SAEs and only the Matryoshka SAE is available. Since the Matryoshka SAE exists only with a 32K dictionary, it appears only in this section. On SAE Probes at $k=40$, the Matryoshka SAE is comparable to the MF-SAEs at layer 12 and outperforms them at layer 20. The comparisons with these baselines are shown in Figures~\ref{fig:a32-l12-sb} and~\ref{fig:a32-l12-sp} for layer 12 and Figures~\ref{fig:a32-l20-sb} and~\ref{fig:a32-l20-sp} for layer 20.

\section{Linear projection ablation}
\label{app:ablations}

\begin{figure*}[t]
    \centering
    \includegraphics[width=0.6\textwidth]{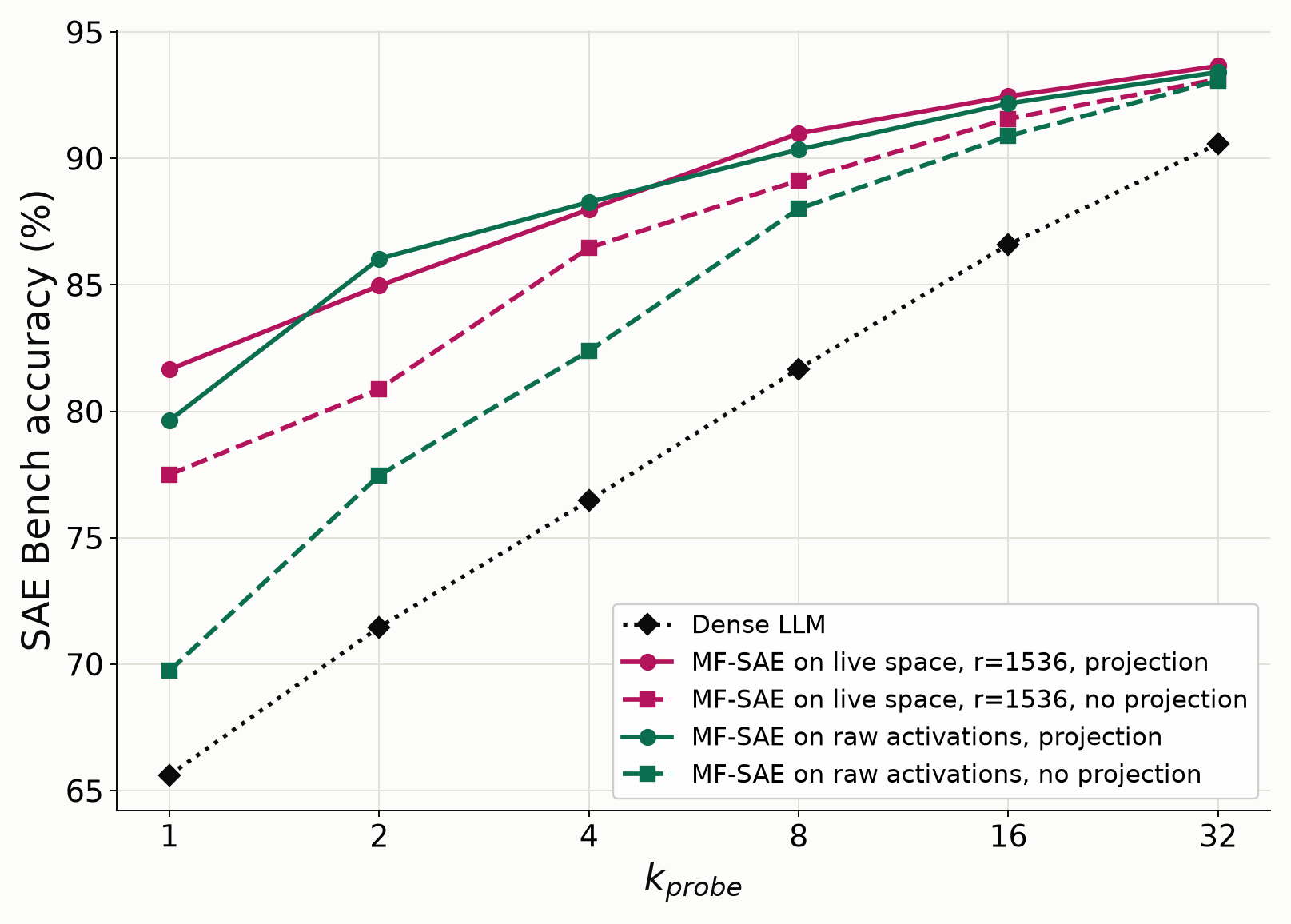}
    \caption{\textbf{Linear projection ablation (SAEBench, Gemma-2-2B, layer 12, 16K dictionary, $k=40$).} SB-Acc against $k_{\text{probe}}$ for the raw and live MF-SAEs with and without the linear projections. The projections significantly improve both.}
    \label{fig:proj-vs-noproj}
\end{figure*}

We compare the MF-SAE with and without the linear projections $W_e$ and $W_d$ (Section~\ref{sec:mfsae}) at layer 12, 16K dictionary, $k=40$. The linear encoder and decoder significantly improve probing accuracy on both the raw and the live space (Figure~\ref{fig:proj-vs-noproj}), consistent with Section~\ref{sec:mfsae}: they are the only mechanism that lets the encoder of the MF-SAE differ from its decoder.

\section{GPT-2 results}
\label{app:gpt2}

We train MF-SAEs on GPT-2 Small ($d=768$) at layers 7 and 8, matching the hook point, dictionary size and sparsity of an available open-source SAE at each layer (Table~\ref{tab:baselines-gpt2}). At layer 7 we train both the raw and the live MF-SAE, with the Jacobian estimated as in Section~\ref{sec:setup-training} and $r=512$ live directions; at layer 8 we train only the raw MF-SAE.

\paragraph{Layer 7.} We compare with the residual-stream SAE of \cite{bloom2024gpt2small} at \texttt{blocks.7.hook\_resid\_pre}, a ReLU SAE trained with an L1 penalty on 300M tokens of OpenWebText, with $24{,}576$ latents and average L0 $\approx 62$. Our MF-SAEs use $n=24{,}576$ and $k=64$.

\paragraph{Layer 8.} We compare with the Top-K SAE of \cite{gao2025scaling} at \texttt{blocks.8.hook\_resid\_post}, with $32{,}768$ latents and $k=32$. Our MF-SAE uses $n=32{,}768$ and $k=32$.

\begin{figure*}[t]
    \centering
    \begin{subfigure}[b]{0.48\textwidth}
        \centering
        \includegraphics[width=\linewidth, height=0.5\linewidth]{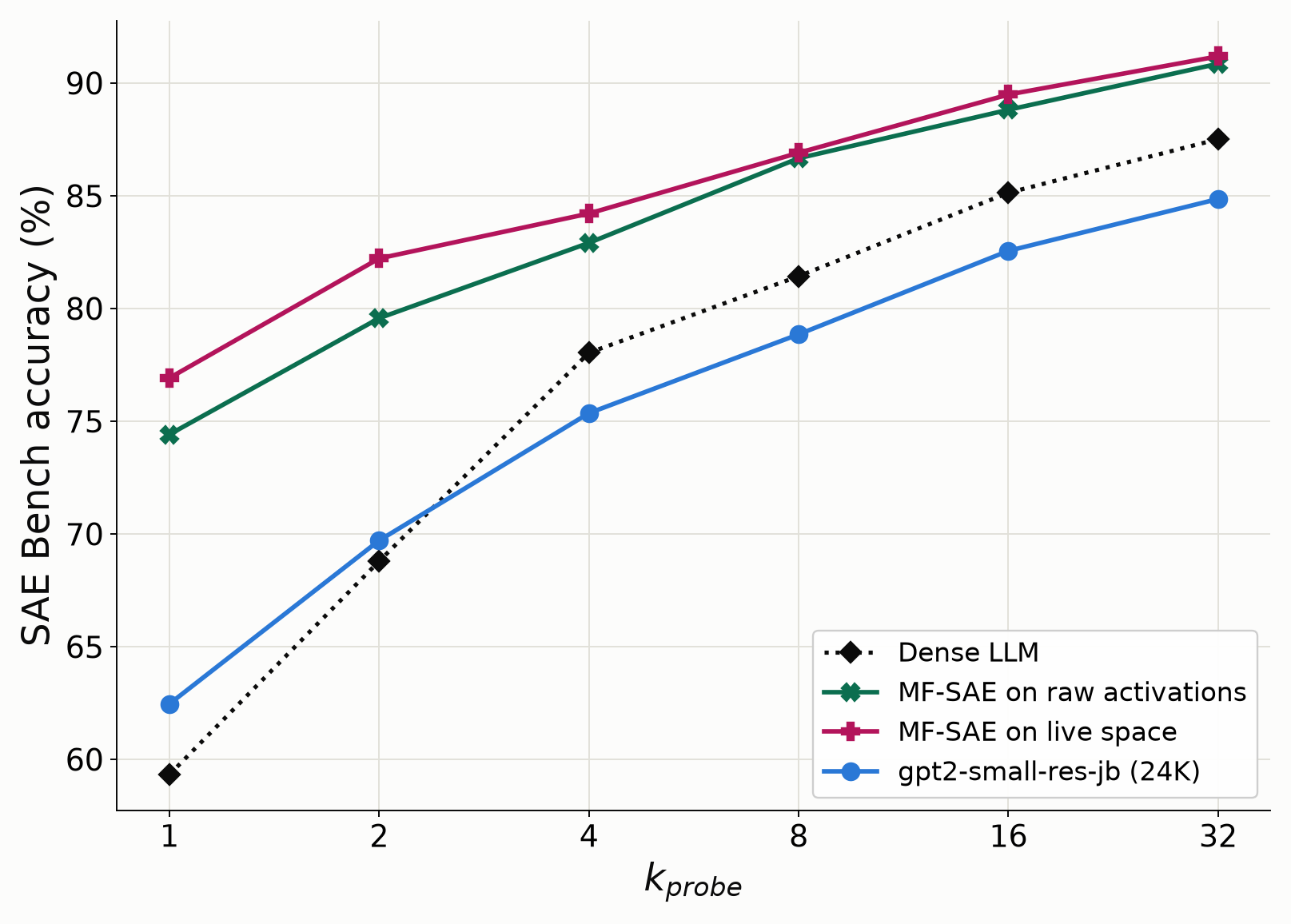}
        \caption{SAEBench, layer 7}
        \label{fig:gpt2s-sb-l7}
    \end{subfigure}
    \hfill
    \begin{subfigure}[b]{0.48\textwidth}
        \centering
        \includegraphics[width=\linewidth, height=0.5\linewidth]{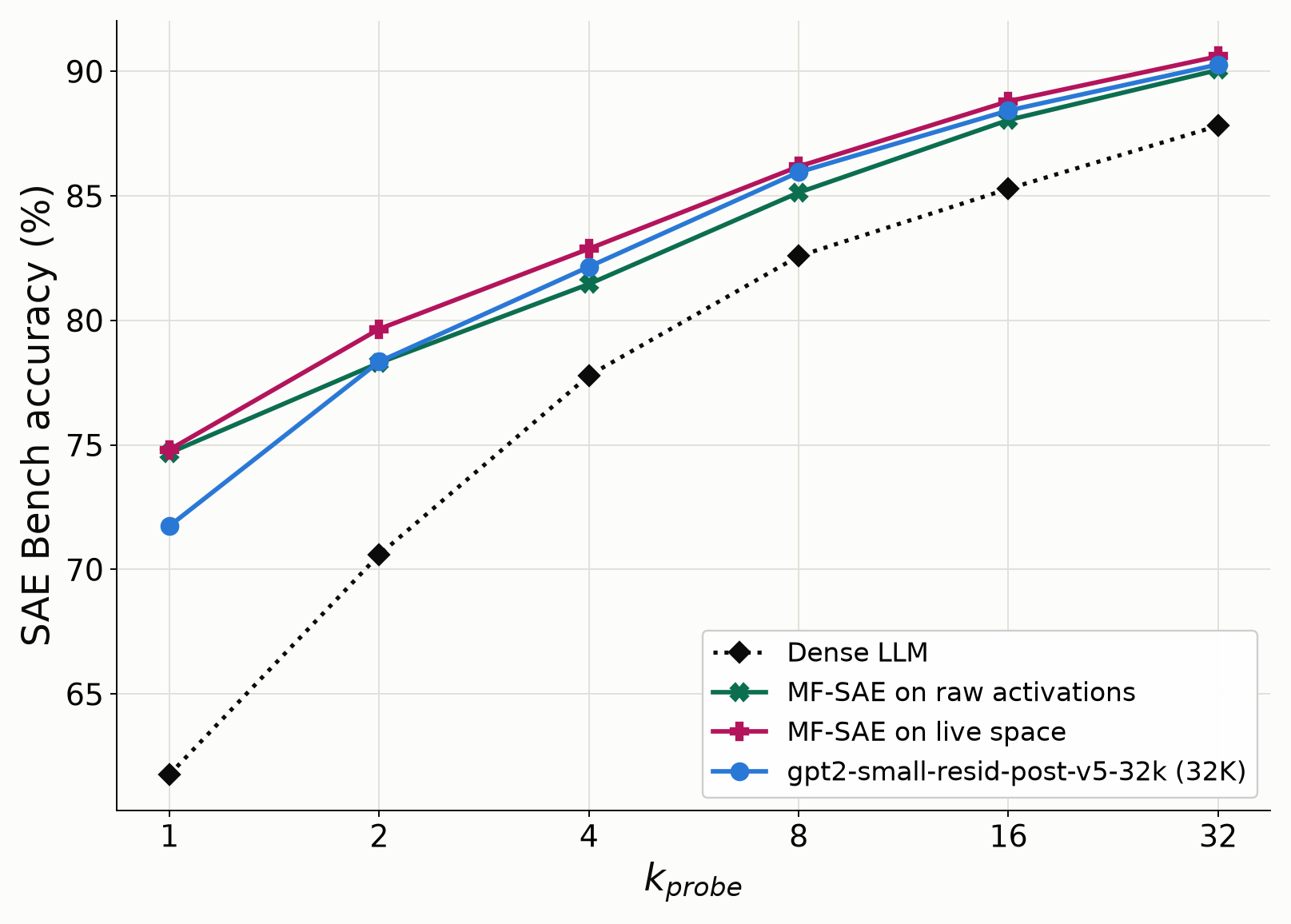}
        \caption{SAEBench, layer 8}
        \label{fig:gpt2s-sb-l8}
    \end{subfigure}

    \vspace{6pt}

    \begin{subfigure}[b]{0.48\textwidth}
        \centering
        \includegraphics[width=\linewidth, height=0.5\linewidth]{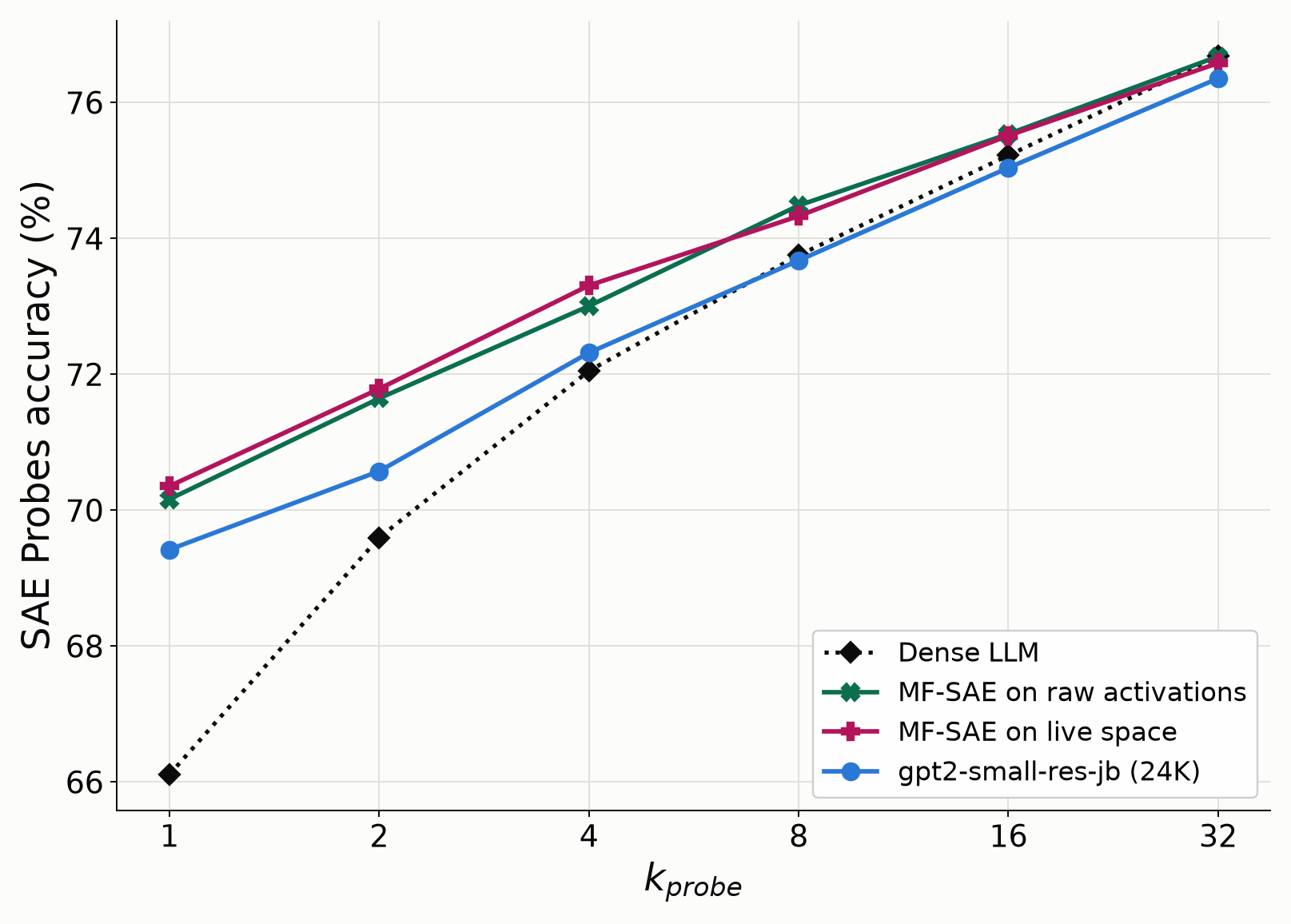}
        \caption{SAE Probes, layer 7}
        \label{fig:gpt2s-sp-l7}
    \end{subfigure}
    \hfill
    \begin{subfigure}[b]{0.48\textwidth}
        \centering
        \includegraphics[width=\linewidth, height=0.5\linewidth]{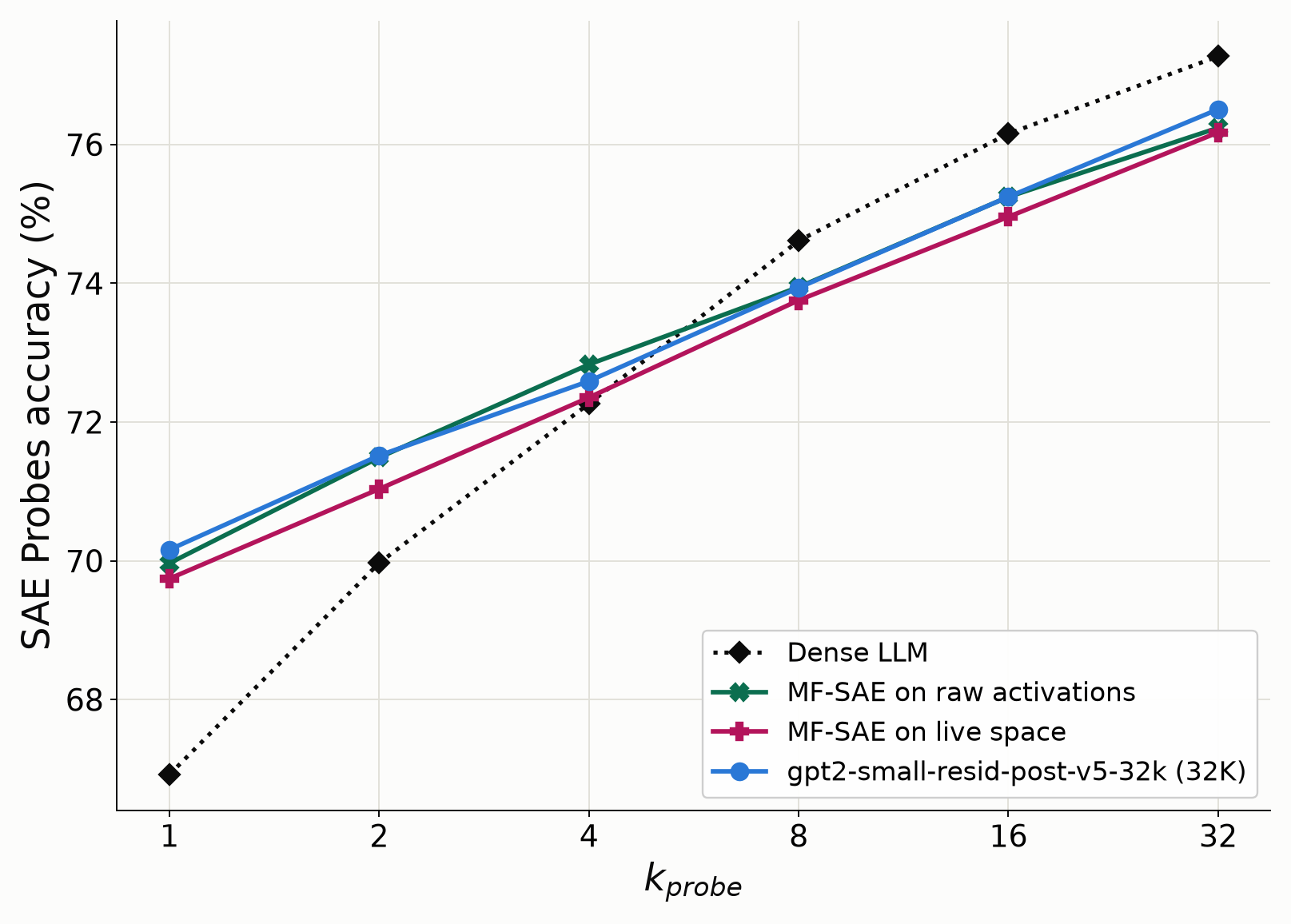}
        \caption{SAE Probes, layer 8}
        \label{fig:gpt2s-sp-l8}
    \end{subfigure}
    \caption{\textbf{GPT-2 Small, matched baselines.} Probe accuracy against $k_{\text{probe}}$. Rows are SAEBench and SAE Probes; columns are layer 7 (raw and live MF-SAEs, $n=24{,}576$, $k=64$, against the SAE of \cite{bloom2024gpt2small}) and layer 8 (raw MF-SAE, $n=32{,}768$, $k=32$, against the Top-K SAE of \cite{gao2025scaling}).}
    \label{fig:gpt2s}
\end{figure*}

\end{document}